\documentclass{informs3}
\OneAndAHalfSpacedXI
\usepackage{shortcuts}
\usepackage{amsmath,amssymb,mathtools,bm}
\usepackage{outlines}
\usepackage{hyperref}
\usepackage[capitalise]{cleveref}
\Crefname{assumption}{Assumption}{Assumptions}
\usepackage{etoolbox}
\AtBeginEnvironment{APPENDICES}{\crefalias{section}{appendix}}
\AtBeginEnvironment{APPENDICES}{\crefalias{subsection}{appendix}}
\usepackage{booktabs}
\usepackage{nicefrac}
\usepackage{url,graphicx}
\usepackage{algorithm,subcaption}
\usepackage[noend]{algpseudocode}
\usepackage{array,multirow}
\usepackage[dvipsnames]{xcolor}
\usepackage{wrapfig}
\usepackage{tikz}
\usetikzlibrary{positioning,arrows,decorations,decorations.markings,shadows,positioning,arrows.meta,matrix,fit}

\newcommand{\X}{\mathcal X}
\newcommand{\Z}{\mathcal Z}
\newcommand{\Y}{\mathcal Y}
\newcommand{\F}{\mathcal F}
\newcommand{\K}{\mathcal K}
\newcommand{\D}{\mathcal D}
\newcommand{\G}{\mathcal G}
\newcommand{\Hil}{\mathcal H}
\newcommand{\Set}{\mathcal S}
\newcommand{\T}{\mathcal T}
\newcommand{\pr}{\mathbb{P}}
\newcommand{\expect}{\mathbb{E}}
\newcommand{\ind}{\mathbb{I}}

\newcommand{\Esig}{\E_{\bm \sigma}}

\newcommand{\myendproof}{\hfill\halmos\endproof}

\newcommand{\tr}{^\top}
\newcommand{\norm}[1]{\left\lVert#1\right\rVert}

\usepackage[sort]{natbib}
\bibpunct[, ]{(}{)}{,}{a}{}{,}%
 \def\bibfont{\small}%

\def\edit{}
\def\blockedit{}

\TheoremsNumberedThrough     %
\ECRepeatTheorems

\EquationsNumberedThrough    %

\newcommand{\myfulltitle}{Fast Rates for Contextual Linear Optimization}%

\begin{document}

\newcommand*\samethanks[1][\value{footnote}]{\footnotemark[#1]}
\RUNTITLE{\myfulltitle}
\TITLE{\myfulltitle}
\RUNAUTHOR{Hu, Kallus, \& Mao}
\ARTICLEAUTHORS{%
\AUTHOR{Yichun Hu,\thanks{Alphabetical order.}~~~~Nathan Kallus,\samethanks~~~~Xiaojie Mao\samethanks}
\AFF{Cornell University, New York, NY 10044, \EMAIL{\{yh767, kallus, xm77\}@cornell.edu}}
}

\ABSTRACT{Incorporating side observations in decision making can reduce uncertainty and boost performance, but it also requires we tackle a potentially complex predictive relationship.
While one may use off-the-shelf machine learning methods to separately learn a predictive model and plug it in, a variety of recent methods instead integrate estimation and optimization by fitting the model to directly optimize downstream decision performance.
Surprisingly, in the case of contextual linear optimization, we show that the na\"ive plug-in approach actually achieves regret convergence rates that are significantly faster than methods that directly optimize downstream decision performance.
We show this by leveraging the fact that specific problem instances do not have arbitrarily bad near-dual-degeneracy.
While there are other pros and cons to consider as we discuss and illustrate numerically, our results highlight a nuanced landscape for the enterprise to integrate estimation and optimization. Our results are overall positive for practice: predictive models are easy and fast to train using existing tools, simple to interpret, and, as we show, lead to decisions that perform very well.}
\HISTORY{First posted version: November, 2020. This version: August, 2021.}

\maketitle

\vspace{-1\baselineskip}

\section{Introduction}

A central tenet of machine learning is the use of rich feature data to reduce uncertainty in an unknown variable of interest, whether it is the content of an image, medical outcomes, or future stock price. Recent work in data-driven optimization has highlighted the potential for rich features to similarly reduce uncertainty in decision-making problems with uncertain objectives and thus improve resulting decisions' performance \citep{kallus2020stochastic,bertsimas2014predictive,elmachtoub2017smart,el2019generalization,ho2019data,notz2019prescriptive,donti2017task,estes2019objective,Loke2020,Nam2019,diao2020distribution,chen2021statistical,rudin2014big}.
For decision-making problems modeled by linear optimization with uncertain coefficients, this is captured by the contextual linear optimization (CLO) problem, defined as follows:
\begin{equation}\label{eq:clo}\ts
\pi^*(x)\in\Z^*(x)=\argmin_{z\in \Z}f^*(x)\tr z,\quad f^*(x)=\Eb{Y\mid X=x},\quad \Z=\braces{z\in\R d:Az\leq b}.
\end{equation}
Here, $X\in\R p$ represents the contextual features, $z\in\Z\subseteq\R d$ linearly-constrained decisions, and $Y\in\R d$ the random coefficients.
Examples of CLO are vehicle routing with uncertain travel times, portfolio optimization with uncertain security returns, and supply chain management with uncertain shipment costs.
In each case, $X$ represents anything that we can observe before making a decision $z$ that can help reduce uncertainty in the random coefficients $Y$, such as recent traffic or market trends.
The decision policy $\pi^*(x)$ optimizes the conditional expected costs, given the observation $X=x$.
(We reserve $X,Y$ for random variables and $x,y$ for their values.)
{We assume throughout that $\Z$ is a polytope ($\sup_{z\in\Z}\magd{z}\leq B$) and $Y$ is bounded (without loss of generality, \edit{$Y\in\Y=\{y:\magd{y}\leq 1\}$}), and we let $\Z^\angle$ denote the set of extreme points of $\Z$.}
Nominally, we only do better by taking features $X$ into consideration when making decisions:
$$\ts
\min_{z\in \Z}\Eb{Y\tr z}\geq \Eb{\min_{z\in \Z}\Eb{Y\tr z\mid X}}=\Eb{f^*(X)\tr \pi^*(X)},
$$
and the more $Y$-uncertainty explained by $X$ the larger the gap. That is, at least if we knew the true conditional expectation function $f^*$. In practice, we do not, and we only have data \edit{$\D=\{(X_1,Y_1),\dots,(X_n,Y_n)\}$, which we assume consist of $n$ independent draws of $(X,Y)$}. The task is then to use these data to come up with a well-performing data-driven policy $\hat \pi(x)$ for the decision we will make when observing $X=x$, namely one having low average regret:
\begin{equation}\label{eq:regret}\ts
\op{Regret}(\hat \pi)=\E_\D\E_X\bracks{f^*(X)\tr(\hat \pi(X)-\pi^*(X))},
\end{equation}
where we marginalize \emph{both} over new features $X$ and over the sampling of the data $\D$ (\ie, over $\hat\pi$).

One approach is the na\"ive plug-in method, also known as \textbf{``estimate and then optimize'' (ETO)}.
Since $f^*$ is the regression of $Y$ on $X$, we can estimate it using a variety of off-the-shelf methods, whether parametric regression such as ordinary least squares or generalized linear models, nonparametric regression such as $k$-nearest neighbors or local polynomial regression, or machine learning methods such as random forests or neural networks. Given an estimate $\hat f$ of $f^*$, we can construct the induced policy $\pi_{\hat f}$, where for any generic $f:\R p\to\R d$ we define the plug-$f$-in policy
\begin{equation}\label{eq:plugin}\pi_f(x)\in\argmin_{z\in \Z}f(x)\tr z.\end{equation}
Notice that given $f$, $\pi_f$ need not be unique; we restrict to choices $\pi_f(x)\in\Z^\angle$ that break ties arbitrarily but consistently (\ie, by some ordering over $\Z^\angle$).
Notice also that $\pi_{f^*}(x)\in\Z^*(x)$.
Given a hypothesis class $\F\subseteq[\R p\to\edit{\Y}]$ for $f^*$, we can for example choose $\hat f$ by least-squares regression:
\begin{equation}\label{eq:MSEERM}
\edit{\hat f_\F\in\argmin_{f\in\F}\frac1n\sum_{i=1}^n\magd{Y_i-f(X_i)}^2.}
\end{equation}
We let $\hat \pi^\text{ETO}_{\F}=\pi_{\hat f_\F}$ be the ETO policy corresponding to least-squares regression over $\F$.
ETO has appealing practical benefits.
It is easily implemented using tried-and-true, off-the-shelf, potentially flexible prediction methods.
More crucially, it easily adapts to decision support, which is often the reality for quantitative decision-making tools: rather than a blackbox prescription, it provides a decision maker with a prediction that she may judge and eventually use as she sees fit.

Nonetheless, a criticism of this approach is that \cref{eq:MSEERM} uses the \emph{wrong} loss function as it does not consider the impact of $\hat f$ on the downstream performance of the policy $\pi_{\hat f}$ and in a sense ignores the decision-making problem. %
The alternative empirical risk minimization \textbf{(ERM) method} directly minimizes an empirical estimate of the average costs of a policy: given a policy class $\Pi\subseteq[\R p\to\Z]$,
\begin{equation}\label{eq:erm}\hat \pi^\text{ERM}_{\Pi}\in\argmin_{\pi\in\Pi}\frac1n\sum_{i=1}^nY_i\tr \pi(X_i).\end{equation}
In particular, a hypothesis class $\F$ induces the plug-in policy class $\Pi_\F=\{\pi_f:f\in\F\}$, and ERM over $\Pi_\F$ corresponds to optimizing the empirical risk of $\pi_f$ over choices $f\in\F$, yielding a \emph{different} criterion from \cref{eq:MSEERM} for choosing $f\in\F$. We call this the induced ERM \textbf{(IERM) method}, which thus \emph{integrates} the estimation and optimization aspects of the problem into one, sometimes referred to as \emph{end-to-end estimation}. We let $\hat \pi^\text{IERM}_{\F}=\hat \pi^\text{ERM}_{\Pi_\F}$ denote the IERM policy induced by $\F$.

Although the latter IERM approach appears to much more correctly and directly deal with the decision-making problem of interest, in this paper we demonstrate a surprising fact:
\vspace{0.25\baselineskip}\begin{center}\it
Estimate-and-then-optimize approaches
can have \underline{\textbf{much}} faster regret-convergence rates.
\end{center}\vspace{0.25\baselineskip}
\label{page:newintro}\edit{To theoretically characterize this phenomenon, we develop regret bounds for ETO and IERM when $f^*\in\F$. Without further assumptions beyond such well-specification (which is necessary for any hope of vanishing regret), we show that the regret convergence rate $1/\sqrt{n}$ reigns. However, appropriately limiting how degenerate an instance can be uncovers faster rates and a divergence between ETO and IERM favoring ETO.
This can be attributed to ETO leveraging structure in $\F$ compared to IERM using only what is implied about $\Pi_\F$.
Numerical examples corroborate our theory's predictions and demonstrate the conclusions extend to flexible/nonparametric specifications, while highlighting the benefits of IERM for simple/interpretable models that are bound to be misspecified.}
We provide a detailed discussion 
on how this fits into the larger practical considerations of choosing between ETO and end-to-end methods such as IERM for developing decision-making and decision-support systems.

\subsection{Background and Relevant Literature}

\paragraph*{Contextual linear and stochastic optimization.}
The IERM problem is generally nonconvex in $f\in\F$. For this reason \citet{elmachtoub2017smart} develop a convex surrogate loss they call SPO+, which they show is Fisher consistent under certain regularity conditions in that if $f^*\in\F$ then the solution to the convex surrogate problem solves the nonconvex IERM problem. \citet{el2019generalization} prove an $O(\log(\abs{\Z^\angle}n)/\sqrt{n})$ regret bound for IERM when $\F$ is linear functions. Both \citet{elmachtoub2017smart,el2019generalization} advocate for the integrated IERM approach to CLO, referring to it as \emph{smart} in comparison to the na\"ive ETO method.

CLO is a special case of the more general contextual stochastic optimization (CSO) problem, $\pi^*(x)\in\argmin_{z\in\Z}\Eb{c(z;Y)\mid X=x}$.
\citet{bertsimas2014predictive} study ETO approaches to CSO where the distribution of $Y\mid X=x$ is estimated by a re-weighted empirical distribution of $Y_i$,
for which they establish asymptotic optimality. 
\citet{diao2020distribution} study stochastic gradient descent approaches to solving the resulting problems.
\citet{ho2019data} propose to add variance regularization to this ETO rule to account for errors in this estimate. \citet{bertsimas2014predictive} additionally study ERM approaches to CSO and \edit{provide} generic regret bounds (see their Appendix EC.1).
\citet{notz2019prescriptive} apply these bounds to reproducing kernel Hilbert spaces (RKHS) in a capacity planning application.
\citet{rudin2014big} study ERM with a sparse linear model for the newsvendor problem.
\citet{kallus2020stochastic} construct forest policies for CSO by using optimization perturbation analysis to approximate the generally intractable problem of ERM for CSO over trees; they also prove asymptotic optimality. Many other works that study CSO generally advocate for \emph{end-to-end} solutions that \emph{integrate} or \emph{harmonize} estimation and optimization \citep{Nam2019,estes2019objective,Loke2020,donti2017task}.

\paragraph*{Classification.} Classification is a specific case of CLO with $Y\in\{-1,1\}$ and $\Z=[-1,1]$. Then \edit{$\frac12\op{Regret}(\hat \pi)=\Prb{Y\neq \hat\pi(X)}-\Prb{Y\neq \pi^*(X)}$} is the excess error rate.
\citet{bartlett2005local,koltchinskii2006local,massart2006risk,vapnik1974theory,tsybakov2004optimal} among others study regret and generalization bounds for ERM approaches, convexifications, and related approaches.
\edit{Our work is partly inspired by}\label{ATdiscussion} 
\citet{audibert2007fast}, who compared \edit{such ERM classification approaches} to methods that estimate $\Prb{Y=1\mid X}$ and then classify by thresholding at $1/2$ \edit{and showed that these can enjoy fast regret convergence rates} under a noise condition (also known as margin) that quantifies the concentration of $\Prb{Y=1\mid X}$ near $1/2$.
\edit{In contrast to \citet{audibert2007fast}, we study fast rates for the more general CLO problem as our aim is to shed light on data-driven optimization, we use complexity notions that allow direct comparison of ETO and IERM (rather than ERM) using the same hypothesis class (while entropy conditions for ERM and plug-in used by \citealt{audibert2007fast} are incomparable), and we provide lower bounds that rigorously show the gap between IERM and ETO for any given polytope (the lower bounds of \citealt{audibert2007fast} only apply to H\"older-smooth functions and classification and they show the optimality of plug-in methods rather than the \emph{sub}optimality of ERM).}
Similar noise \edit{or margin} conditions have also been used in contextual bandits \citep{hu2020smooth,perchet2013multi,rigollet2010nonparametric,goldenshluger2013linear,bastani2020online}. \edit{Our condition is similar to these but adapted to CLO.}

\break
\subsection{A Simple Example}\label{sec:simple}
\begin{wrapfigure}{r}{0.475\textwidth}%
\vspace{-2.5\baselineskip}%
\hfill%
\includegraphics[width=0.42\textwidth]{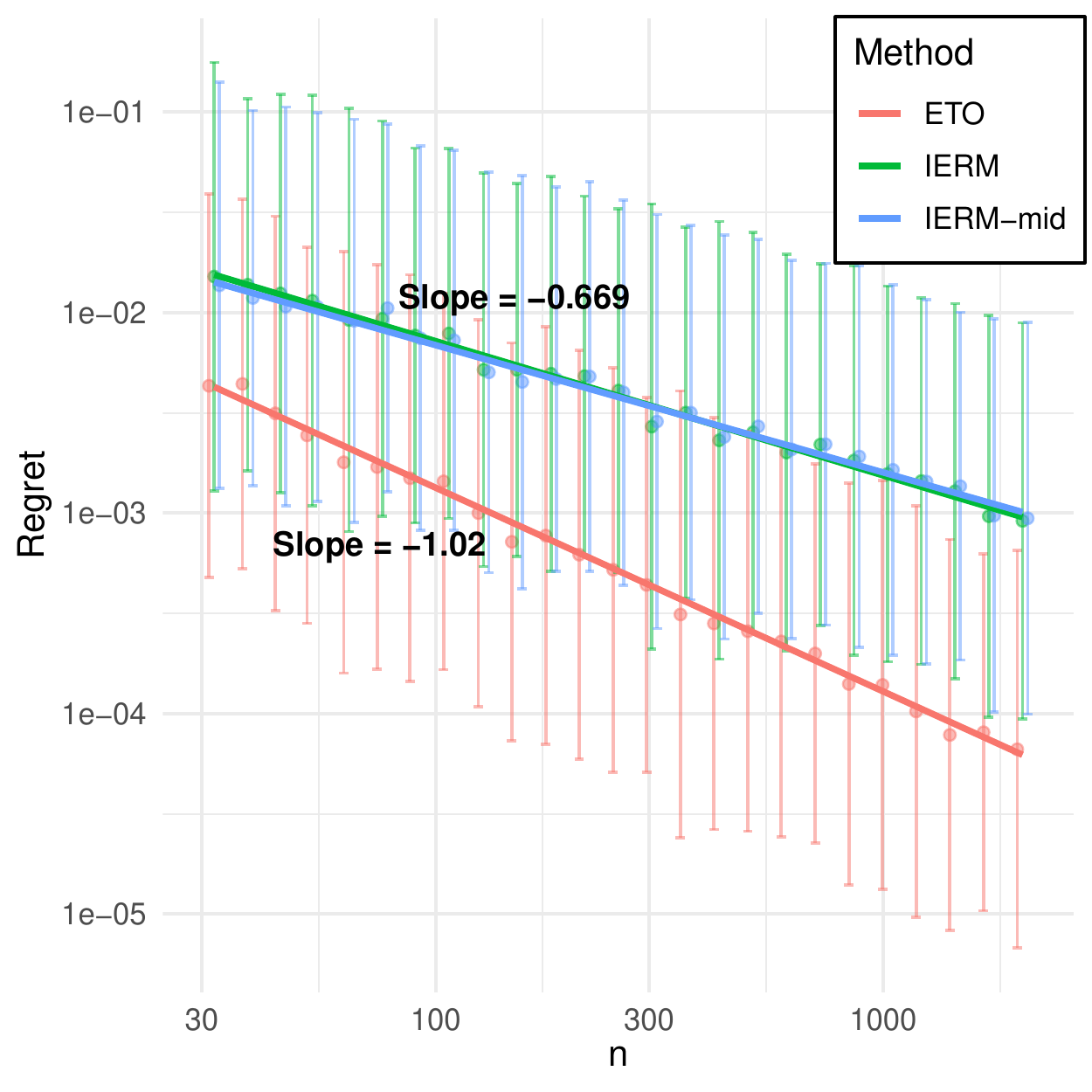}%
\bgroup\hfill\captionof{figure}{Regret convergence rates. Shown is average regret by $n$ plus/minus one standard deviation. Solid lines are log-log linear fits.}\label{fig:simple}\egroup%
\vspace{-1.5\baselineskip}%
\end{wrapfigure}%
We start with a simple illustrative example.
Consider
univariate decisions, $\Z=[-1,1]$,
univariate features, $X\sim\op{Unif}[-1,1]$, and a simple linear relationship, $f^*(X)=X$, $Y-f^*(X)\sim\mathcal N(0,\sigma^2)$. Let us default to $z=-1$ under ties. Then, given a hypothesis set $\F=\{f_{\theta}(x)= x-\theta:\theta\in[-1,1]\}$, we have $\pi_{f_{\theta}}(x)=2\findic{x\leq\theta}-1$. Let us default to smaller $\theta$ under ties.
We can compute $\E_X[f^*(X)\tr(\pi_{f_{\theta}}(X)-\pi^*(X))]=
\frac12\theta^2$.
We can also immediately see that $\hat\pi_\F^\text{ETO}(x)=2\findic{x\leq\hat\theta_\text{OLS}}-1$ where $\hat\theta_\text{OLS}=\frac1n\sum_{i=1}^nX_i-\frac1n\sum_{i=1}^nY_i\sim\mathcal N(0,\frac{\sigma^2}n)$. Thus, $\op{Regret}(\hat\pi_\F^\text{ETO})=\frac{\sigma^2}{2n}$.

Unfortunately, $\hat\pi_\F^\text{IERM}$ and its regret is harder to compute. We can instead study it empirically. 
\Cref{fig:simple} displays results for 500 replications for each of $n=32,38,45,\dots,2048$ with $\sigma^2=1$. The plot is shown on a log-log scale with linear trend fits. The slope for ETO is $-1.02$ and for IERM is $-0.669$. (We also plot IERM where we choose the midpoint of the argmin set for $\theta$ rather than left endpoint to show not much changes. In the special case of $\sigma^2=0$, we can actually analytically derive $\op{Regret}(\hat\pi_\F^\text{IERM})=\Theta(1/n^2)$, infinitely slower than $\op{Regret}(\hat\pi_\F^\text{ETO})=0$; see \cref{sec:simpleierm}.)

The first thing to note is that \emph{both} slopes are \emph{steeper} than the usual $1/\sqrt{n}$ convergence rate (\ie, $-0.5$ slope), such as is derived in \citet{el2019generalization}. This suggests the usual theory does not correctly predict the behavior in practice. %
The second thing to note is that the slope for ETO is steeper than for IERM, 
with an apparent rate of convergence of
$n^{-1}$ as compared to $n^{-2/3}$.
While ETO is leveraging all the information about $\F$, IERM is only leveraging what is implied about $\Pi_\F$, so it cannot, for example, distinguish between $\theta$ values lying between two consecutive observations of $X$.
Our fast (noise-dependent) rates will \emph{exactly} predict this divergent regret behavior.
Note this very simple example is only aimed to illustrate this convergence phenomenon and need not be representative of real problems, which we explore further in \cref{sec:discussion,sec:exp}.

\section{Slow (Noise-Independent) Rates}\label{sec:slow}

Our aim is to obtain regret bounds in terms of \emph{primitive} quantities that are \emph{common} to \emph{both} the ETO and IERM approaches.
To compare them, we will consider implications of our general results for the case of a correctly specified hypothesis class $\F$ with bounded \emph{complexity}.
One standard notion of the complexity for scalar-valued functions $\F\subseteq[\R p\to\Rl]$ is the VC-subgraph dimension \citep{dudley1987universal}. No commonly accepted notions appear to exist for vector-valued classes of functions. Here we define and use an apparently new, natural extension of VC-subgraph dimension.

\begin{definition}\label{def:vc-lin-major}
The VC-linear-subgraph dimension of a class of functions $\F\subseteq[\R p\to\R d]$ is the
VC dimension of the sets $\F^\circ=\{\{(x,\beta,t):\beta\tr f(x)\leq t\}:f\in\F\}$ in $\R{p+d+1}$, that is, the largest integer $\nu$ for which there exist $x_1,\dots,x_\nu\in\R p,\,\beta_1,\dots,\,\beta_\nu\in\R d,\,t_1\in\Rl,\,\dots,\,t_\nu\in\Rl$ such that
$$
{\{(\findic{\beta_1\tr f(x_1)\leq t_1},\dots, \findic{\beta_\nu\tr f(x_\nu)\leq t_\nu}):f\in\F\}}=\{0,1\}^\nu.
$$
\end{definition}
{\blockedit Our standing assumption will be that $f^*\in\F$ where $\F$ has bounded VC-linear-subgraph dimension.} (In \cref{appendix:rkhs,appendix: local poly} we study other functions classes, including RKHS and \edit{H\"older functions}.)
\begin{assumption}[Hypothesis class]\label{asm:vc}
$f^*\in\F$, 
$\F$ has VC-linear-subgraph dimension at most $\nu$.
\end{assumption}
\begin{example}[Vector-valued linear functions]\label{ex: linear predictors}
Suppose $\F\subseteq\{Wx:W\in\R{d\times p}\}$.
(Note we can always pre-transform $x$.)
Since $\beta\tr f(x)={\op{vec}(W)}\tr{\op{vec}(\beta x\tr)}$, the VC-linear-subgraph dimension of $\F$ is at most the usual VC-subgraph dimension of $\{v\mapsto w\tr v:w\in \R{dp}\}$, which is $dp$.
\end{example}
\begin{example}[Trees]\label{ex: trees}
Suppose $\F$ consists of all binary trees of depth at most $D$, where each internal node queries ``$w^\top x\leq\theta$?'' for a choice of $w\in\R p,\,\beta_0\in\Rl$ for {each} internal node, splitting left if true and right otherwise, and each leaf node assigns the output $v$ to $x$ that reach it, for any choice of $v\in\R d$ for {each} leaf node. (In particular, this is a superset of restricting $w$ to be a vector of all zeros except for a single one so that the splits are axis-aligned.) Then, $\F^\circ$ is contained in the disjunction over leaf nodes of the classes of sets representable by a leaf, which is the conjunction over internal nodes' half-spaces on the path to the leaf and over the final query of $\beta\tr v\leq t$. Since there are at most $2^D$ leaf nodes and at most $D$ internal nodes on the path to each, applying \citet[Theorem 1.1]{van2009note} twice, the VC dimension of $\F^\circ$ is at most $22(D^2p+Dd)2^{D}\log(8D)$.
\end{example}

\subsection{Slow Rates for ERM and IERM}\label{sec: slow erm}

We first establish a generalization result for generic ERM for CLO and then apply it to IERM.
\begin{definition}\label{def: natarajan}
The Natarajan dimension of a class of functions $\G\subseteq[\R p\to\mathcal S]$ with co-domain $\mathcal S$ is the largest integer $\eta$ for which there exist $x_1,\dots,x_\eta\in\R p,\,s_1\neq s_1',\dots,s_\eta\neq s'_\eta\in\mathcal S$ such that
$$
\{(\findic{g(x_1)=s_1},\dots,\findic{g(x_\eta)=s_\eta}):g\in\G,\,g(x_1)\in\{s_1,s'_1\},\,\cdots,\,g(x_\eta)\in\{s_\eta,s'_\eta\}\}=\{0,1\}^\eta.
$$
\end{definition}
\begin{theorem}\label{thm: ERM slow}
Suppose $\Pi\subseteq[\R p\to\Z^\angle]$ has Natarajan dimension at most $\eta$.
Then, for a universal constant $C$, with probability at least $1-\delta$,
\begin{equation}\label{eq: slow generalization}
\sup_{\pi\in\Pi}\abs{
\frac1n\sum_{i=1}^nY_i\tr \pi(X_i)-\E_X\bracks{f^*(X)\tr \pi(X)}
}\leq CB\sqrt{\frac{\eta\log(\abs{\Z^\angle}+1)\log(5/\delta)}{n}}.
\end{equation}
\end{theorem}
\Cref{eq: slow generalization} also implies that the in-class excess loss, $\inf_{\pi\in\Pi}\E_X\bracks{f^*(X)\tr \prns{\hat\pi^\text{ERM}_\Pi(X)-\pi(X)}}$, is bounded by twice the right-hand side of \cref{eq: slow generalization}.
Note \citet{el2019generalization} prove a similar result to \cref{thm: ERM slow} but with an additional suboptimal dependence on $\sqrt{\log(n)}$.

To study IERM, we next relate VC-linear-subgraph dimension to Natarajan dimension.
\begin{theorem}\label{prop: n-dim}
The VC-linear-subgraph dimension of $\F$ bounds the Natarajan dimension of $\Pi_\F$.
\end{theorem}
\begin{corollary}\label{thm: IERM slow}
Suppose \cref{asm:vc} holds. Then, for a universal constant $C$,
$$
\op{Regret}(\hat \pi^\text{IERM}_{\F})\leq CB\sqrt{\frac{\nu\log\prns{\abs{\Z^\angle}+1}}{n}}.
$$
\end{corollary}

\edit{We can in fact show that the rate in \cref{thm: ERM slow} is optimal in $n$ and $\eta$ by showing any algorithm must suffer at least this rate on some example. 
When $\Z=[-1,1]$ our result reduces to that of 
\citet{devroye1995lower} for binary classification, but we tackle CLO with \emph{any} polytope $\Z$.
\begin{theorem}\label{thm:lowerboundslow}
Fix any polytope $\Z$.
Fix any $\Pi\subseteq[\R p\to\Z^\angle]$ with Natarajan dimension at least $\eta$.
Fix any algorithm mapping $\D\mapsto\hat\pi\in\Pi$. Then there exists a distribution $\mathbb P$ on $(X,Y)\in\R{p}\times\Y$ satisfying $\pi^*\in\Pi$ such that for any $n\ge 4\eta$, when $\D\sim\mathbb P^n$, we have
$$
\op{Regret}(\hat \pi)\geq \frac{\rho(\Z)}{2e^4}\sqrt{\frac{\eta}{n}},
$$
where $\rho(\Z)=\inf_{z\in\Z^\angle,\,z'\in\op{conv}(\Z^\angle\backslash\{z\})}\magd{z-z'}$ (which is positive by definition).
\end{theorem}
In general \cref{thm:lowerboundslow} also shows that the rate in \cref{thm: IERM slow} is optimal in $n$ when we only assume $\pi^*\in\Pi_\F$, but not necessarily in $\nu$, since \cref{prop: n-dim} is only an upper bound. 
In many cases, however, it can be very tight. 
In \cref{ex: linear predictors} we upper bounded the VC-linear-subgraph dimension of $\F$ by $dp$, while Corollary 29.8 of \citet{shalev2014understanding} shows the Natarajan dimension of $\Pi_\F$ is at least $(d-1)(p-1)$ when $\Z$ is the simplex, so the gap is very small.
}

\subsection{Slow Rates for ETO}\label{sec: slow eto}

We next establish comparable rates for ETO. 
The following is immediate from Cauchy-Schwartz.

\begin{theorem}\label{thm: slowrate generic}
Let $\hat f$ be given. Then,
\begin{align*}
\op{Regret}(\pi_{\hat{f}}) \le 2B\E_\D\E_X{\|f^*(X)- \hat{f}(X) \|}.
\end{align*}
\end{theorem}
To study ETO under \cref{asm:vc}, we next establish a convergence rate for $\hat f_\F$ to plug in above. 
\begin{theorem}\label{thm: MSE ERM}
Suppose \cref{asm:vc} holds and that $\F$ is star shaped at $f^*$, \ie, $(1-\lambda)f+\lambda f^*\in\F$ for any $f\in\F,\lambda\in[0,1]$. Then, there exist positive universal constants $C_0,C_1,C_2>0$ such that, for any $\delta\leq (nd+1)^{-C_0}$,
with probability at least $1-C_1\delta^\nu$,
\begin{align*}
\E_X\|\hat{f}_\F(X) - f^*(X)\|\leq C_2\sqrt{\frac{\nu\log(1/\delta)}{n}}.
\end{align*}
\end{theorem}
In \cref{sec: MSE ERM proof}, we prove a novel finite-sample guarantee for least squares with vector-valued response over a general function class $\F$, which is of independent interest (relying on existing results for scalar-valued response leads to suboptimal dependence on $d$). 
\Cref{thm: MSE ERM} is its application to the VC-linear-subgraph case.
The star shape assumption is purely technical but, while it holds for \cref{ex: linear predictors}, it does not for \cref{ex: trees}. We can avoid it by replacing $\F$ with $\overline{\F}=\{(1-\lambda)f+\lambda f':f,f'\in\F,\lambda\in[0,1]\}$ in \cref{eq:MSEERM} (for \cref{ex: trees}, we even have $\overline{\F}=\F+\F$), which does not affect the result, only the universal constants. We omit this because least squares over $\overline{\F}$ is not so standard.

\begin{corollary}\label{thm:slowrate}
Suppose the assumptions of \cref{thm: MSE ERM} hold.
Then, for a universal constant $C$,
$$
\op{Regret}(\hat \pi^\text{ETO}_{\F})\leq CB\sqrt{\frac{\nu \log(nd+1)}{n}}.
$$
\end{corollary}
We can remove the term $\log(nd+1)$ in the specific case of \cref{ex: linear predictors} (see \cref{lemma: linear class rates} in \cref{appendix:rkhs}). Since $\log\prns{\abs{\Z^\angle}+1}$ is generally of order $d$ \citep{barvinok2013bound,henk200416}, the $d$-dependence above may be better than in \cref{thm: IERM slow} even for general VC-linear-subgraph classes.

\label{page:complexityremark}\edit{Note \cref{thm: IERM slow,thm:slowrate} uniquely enable us to compare ETO and IERM using the same primitive complexity measure. 
In contrast, complexity measures like bounded metric entropy or Rademacher complexity on $\F$  may not provide similar control on the complexity of $\Pi_\F$.}
\edit{The slow rates for IERM and ETO are nonetheless the same (up to polylogs), suggesting no differentiation between the two.
Studying finer instance characteristics beyond specification reveals the differentiation.}

\section{Fast (Noise-Dependent) Rates}\label{sec:fast}

We next show that much faster rates actually occur in any one instance. To establish this, we characterize the \emph{noise} in an instance as the level of near-dual-degeneracy (multiplicity of solutions).

\begin{assumption}[Noise condition]\label{asm:margin}
Let $\Delta(x)=\inf_{z\in\Z^\angle\backslash\Z^*(x)}f^*(x)\tr z-\inf_{z\in\Z}f^*(x)\tr z$ if $\Z^*(x)\neq \Z$ and otherwise $\Delta(x)=0$.
Assume for some $\alpha,\gamma\geq0$,
\begin{equation}\label{def:margin}
    \mathbb P_X\prns{0<\Delta(X)\leq \delta}\leq (\gamma\delta\edit{/B})^\alpha\quad\forall \delta>0.
\end{equation}
\end{assumption}
\Cref{asm:margin} controls the mass of $\Delta(X)$ near (but not at) zero.
It always holds for $\alpha=0$ (with $\gamma=1$). If $\Delta(X)\geq B/\gamma$ is bounded away from zero (akin to strict separation assumptions in \citealp{foster2020instance,massart2006risk}) then \cref{asm:margin} holds for $\alpha\to\infty$.
Generically, for any one instance, \cref{asm:margin} holds for \emph{some} $\alpha\in(0,\infty)$.
\Eg, if $X$ has a bounded density and $f^*(x)$ has a Jacobian that is uniformly nonsingular (or, if $f^*(x)$ is linear) then \cref{asm:margin} holds with $\alpha=1$.
In particular, the example in \cref{sec:simple} has $\Delta(X)=\abs{X}\sim\op{Unif}[0,1]$ 
and hence $\alpha=1$.

\subsection{Fast Rates for ERM and IERM}\label{sec: fast erm}

Under \cref{asm:margin}, we can obtain a faster rate both for generic ERM and specifically for IERM.
\begin{theorem}\label{thm:fastrateerm}
Suppose \cref{asm:margin} holds, $\Prb{\abs{\Z^*(X)}>1}=0$, $\Pi\subseteq[\R p\to\Z^\angle]$ has Natarajan dimension at most $\eta$, and $\pi^* \in \Pi$. Then, for a constant \edit{$C(\alpha,\gamma)$ depending only on $\alpha,\gamma$},
$$\op{Regret}(\hat \pi^\text{ERM}_{\Pi})\leq \edit{C(\alpha, \gamma) B} \prns{\frac{\eta\log(\abs{\Z^\angle}+1)\log(n+1)}{n}}^{\frac{1+\alpha}{2+\alpha}}.$$\end{theorem}
Whenever $\alpha>0$, this is faster than the noise-independent rate (\cref{thm: ERM slow}).
$\Prb{\abs{\Z^*(X)}>1}=0$ requires that, in addition to nice near-dual-degeneracy, we almost never have exact dual degeneracy.

\begin{corollary}\label{thm:fastrate IERM}
Suppose \cref{asm:vc,asm:margin} hold and $\Prb{\abs{\Z^*(X)}>1}=0$. Then,
$$\op{Regret}(\hat \pi^\text{IERM}_{\F})\leq \edit{C(\alpha, \gamma) B}\prns{\frac{\nu\log(\abs{\Z^\angle}+1)\log(n+1)}{n}}^{\frac{1+\alpha}{2+\alpha}}.$$\end{corollary}

Notice that with $\alpha=1$, this \emph{exactly} recovers the rate behavior observed empirically in \cref{sec:simple}.
\edit{We next show that the rate in $n$ in \cref{thm:fastrateerm,thm:fastrate IERM} (and in $\eta$ in the former) is in fact optimal (up to polylogs) under \cref{asm:margin} when we only rely on well-specification of the policy.
\begin{theorem}\label{thm:lowerboundfast}
Fix any $\alpha\geq0$.
Fix any polytope $\Z$.
Fix any $\Pi\subseteq[\R p\to\Z^\angle]$ with Natarajan dimension at least $\eta$.
Fix any algorithm mapping $\D\mapsto\hat\pi\in\Pi$. Then there exists
a distribution $\mathbb P$ on $(X,Y)\in\R{p}\times\Y$ satisfying $\pi^*\in\Pi$ and \cref{asm:margin} with the given $\alpha$ and $\gamma = B/\rho(\Z)$ such that for any $n\ge 2^{2+\alpha} (\eta-1)$, when $\D\sim\mathbb P^n$, we have
$$
\op{Regret}(\hat \pi)\geq \frac{\rho(\Z)}{2e^4}\prns{\frac{\eta-1}{n}}^{\frac{1+\alpha}{2+\alpha}}.
$$
\end{theorem}}

\subsection{Fast Rates for ETO}\label{sec: fast eto}

We next show the noise-level-specific rate for ETO is \emph{even} faster, sometimes \emph{much} faster. While \edit{\cref{thm:fastrateerm,thm:lowerboundfast}} are tight if we only leverage information about the policy class, leveraging the information on $\F$ itself, as ETO does, can break that barrier and lead to better performance.

\begin{theorem}\label{thm:fastrate}
Suppose \cref{asm:margin} holds and, for universal constants $C_1,C_2$ and a sequence $a_n$, $\hat f$ satisfies that, for any $\delta>0$ and almost all $x$,
$\pr(\|\hat{f}(x) - f^*(x)\|\ge \delta)\le C_1 \exp(- C_2 a_n \delta^2)$.
Then, for a constant \edit{$C(\alpha,\gamma)$ depending only on $\alpha,\gamma$},
\begin{align*}\ts
\op{Regret}(\pi_{\hat{f}}) \le
\edit{C(\alpha, \gamma) B}\;
a_n^{-\frac{1+\alpha}{2}}.
\end{align*}
\end{theorem}

While \cref{thm: slowrate generic} requires $\hat f$ to have good average error, \cref{thm:fastrate} requires $\hat f$ to have a point-wise tail bound on error with rate $a_n$. This is generally stronger but holds for a variety of estimators. For example, 
if $\hat f$ is given by, \eg, a generalized linear model then we can obtain $a_n=n$ \citep{mccullagh1989generalized}, which together with \cref{thm:fastrate} leads to an \emph{even} better regret rate of $n^{-\frac{1+\alpha}{2}}$.

While such point-wise rates generally hold when $\hat f$ is parametric, \edit{VC-linear-subgraph dimension only characterizes average error so a comparison based on it requires we also make a recoverability assumption to study pointwise error \citep[see also][]{foster2020instance,hanneke2011rates}.}
In \cref{sec: veryify compat}, we show
\cref{asm:compatibility} generally holds for \cref{ex: linear predictors,ex: trees} (\cref{prop: linear compat,prop: tree compat}). 
\begin{assumption}[Recovery]\label{asm:compatibility}
There exists $\kappa$ such that for all $f\in\F$ and almost all $x$,
$$
\|{f}(x) - f^*(x)\|^2\leq \kappa\Efb{\|{f}(X) - f^*(X)\|^2}
$$
\end{assumption}
\begin{corollary} \label{thm:fastrateeto}
Suppose \cref{asm:vc,asm:margin,asm:compatibility} hold and $\F$ is star shaped at $f^*$. Then,%
$$
\op{Regret}(\hat \pi^\text{ETO}_{\F})\leq \edit{C(\alpha, \gamma) B}\edit{\kappa^{1+\alpha}} \prns{\frac{\nu \log(nd+1)}{n}}^{\frac{1+\alpha}{2}}.
$$
\end{corollary}
With $\alpha=1$, this \emph{exactly} recovers the rate behavior observed in \cref{sec:simple}. We can also remove the $\log(nd+1)$ term in the case of \cref{ex: linear predictors} (see \cref{lemma: linear class rates} in \cref{appendix:rkhs}). 
Compared to \cref{thm:fastrate IERM}, we see the regret rate's exponent in $n$ is faster by a factor of $1+\frac{\alpha}{2}$. This can be attributed to using all the information on $\F$ rather than just what is implied about $\Pi_\F$.

\subsection{Fast Rates for Nonparametric ETO}\label{sec:fastnonparam}

\Cref{asm:vc} is akin to a parametric restriction, but ETO can easily be applied using any flexible nonparametric or machine learning regression. For some such methods we can also establish theoretical results (with correct $d$-dependence, compared to relying on existing results for regression).
If, instead of \cref{asm:vc}, we assume that $f^*$ is $\beta$-smooth (roughly meaning it has $\beta$ derivatives), then we show in \cref{appendix: local poly} how to construct an estimator $\hat f$ satisfying the point-wise condition in \cref{thm: slowrate generic} with $a_n=n^{\frac{2\beta}{2\beta+p}}/d$ and without a recovery assumption. This leads to a regret rate of $n^{-\frac{\beta(1+\alpha)}{2\beta+p}}$ for ETO. While slower than the rate in \cref{thm:fastrateeto}, the restriction on $f^*$ is nonparametric, and the rate can still be arbitrarily fast as either $\alpha$ or $\beta$ grow.
In \cref{appendix:rkhs} we also analyze estimates $\hat f$ based on kernel ridge regression, which we also deploy in experiments in \cref{sec:exp}.

\section{Considerations for Choosing Separated vs Integrated Approaches}\label{sec:discussion}

We next provide some perspective on our results and on their implications. We frame this discussion as a comparison between IERM and ETO approaches to CLO along several aspects.

\subsubsection*{Regret rates.} \Cref{sec:slow} shows that the noise-level-agnostic regret rates for IERM and ETO have the same $n^{-1/2}$-rate (albeit, the ETO rate may also have better $d$-dependence). But this hides the fact that specific problem instances do not actually have \emph{arbitrarily} bad near-degeneracy, \ie, they satisfy \cref{asm:margin} for \emph{some} $\alpha>0$. When we restrict how bad the near-degeneracy can be, we obtained fast rates in \cref{sec:fast}. In this regime, we showed that ETO can actually have \emph{much} better regret rates than IERM. It is important to emphasize that, since specific instances \emph{do} satisfy \cref{asm:margin}, this regime truly captures how these methods actually behave in practice in specific problems. Therefore, in terms of regret rates, this shows a clear preference for ETO approaches.

\subsubsection*{Specification.} Our theory focused on the well-specified setting, that is, $f^*\in\F$. When this fails, convergence of the regret of $\hat\pi$ to $\pi^*$ to zero is essentially hopeless for any method that focuses only on $\F$. ERM, nonetheless, can still provide best-in-class guarantees: regret to the best policy in $\Pi$ still converges to zero. For induced policies, $\pi_f$, this means IERM gets best-in-class guarantees over $\Pi_\F$, while ETO may not. Given the fragility of correct specification if $\F$ is too simple, the ability to achieve best-in-class performance is important and may be the primary reason one might prefer (I)ERM to ETO. Nonetheless, if $\F$ is not well-specified, it begs the question why use IERM rather than ERM directly over some policy class $\Pi$. The benefit of using $\Pi_\F$ may be that it provides an easy way to construct a reasonable policy class that respects the decision constraints, $\Z$.

\subsubsection*{\edit{BYOB (Bring Your Own Blackbox).}}\label{paranonparamext} \edit{While IERM is necessarily given by optimizing over $\F$ and is therefore specified by $\F$, ETO accommodates any regression method as a blackbox, not just least squares. This is perhaps most important in view of specification: many flexible regression methods, including local polynomial or gradient boosting regression, do not take the form of minimization over $\F$. (See \cref{sec:fastnonparam} regarding guarantees for the former.)}

\subsubsection*{Interpretability.} 
ETO has the benefit of an \emph{interpretable output}: rather than just having a black box spitting out a decision with no explanation, our output has a clear interpretation as a prediction of $Y$. We can therefore probe this prediction and understand more broadly what other implications it has, such as what happens if we changed our constraints $\Z$ and other counterfactuals.
This is absolutely crucial in decision-support applications, which are the most common in practice.

\edit{If we care about \emph{model explainability} -- understanding \emph{how} inputs lead to outputs -- it may be preferable to focus on simple models like shallow trees. For these, which are likely not well-specified, IERM has the benefit of at least ensuring best-in-class performance \citep{elmachtoub2020decision}.}

\subsubsection*{Computational tractability.} Another important consideration is tractability. For ETO, this reduces to learning $\hat f$, and both classic and modern prediction methods are often tractable and built to scale.
On the other hand, IERM is nonconvex and may be hard to optimize. This is exactly the motivation of \citet{elmachtoub2017smart}, which develop a convex relaxation. However, it is only consistent if $\F$ is well-specified, in which case we expect ETO has better performance.

\subsubsection*{\edit{Contextual} stochastic optimization.} While we focused on CLO, a question is what do our results suggest for CSO generally. CSO with a finite feasible set (or, set of possibly-optimal solutions), $\Z=\{z^{(1)},\dots,z^{(K)}\}$, is immediately reducible to CLO by replacing $\mathcal Z$ with the $K$-simplex and $Y$ with $(c(z^{(1)};Y),\dots,c(z^{(K)};Y))$. Then, our results still apply. Continuous CSO may require a different analysis to account for a non-discrete notion of a noise condition. In either the continuous or finite setting, however, ETO would entail learning a high-dimensional object, being the conditional distribution of $Y\mid X=x$ (or, rather, the conditional expectations $\E[c(z;Y)\mid X=x]$ for \emph{every} $z\in\Z$, whether infinite or finite and big). While certainly methods for this exist, if $\Z$ has reasonable dimensions, 
a purely-policy-based approach, such as ERM or IERM, might be more practical.
For example, \citet{kallus2020stochastic} show that directly targeting the downstream optimization problem when training random forests significantly improves forest-based approaches to CSO.
This is contrast to the CLO case, where both the decision policy and relevant prediction function have the same dimension, both being functions $\R p\to\R d$.

\section{Experiments}\label{sec:exp}

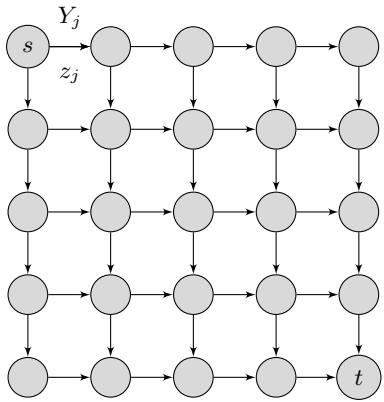
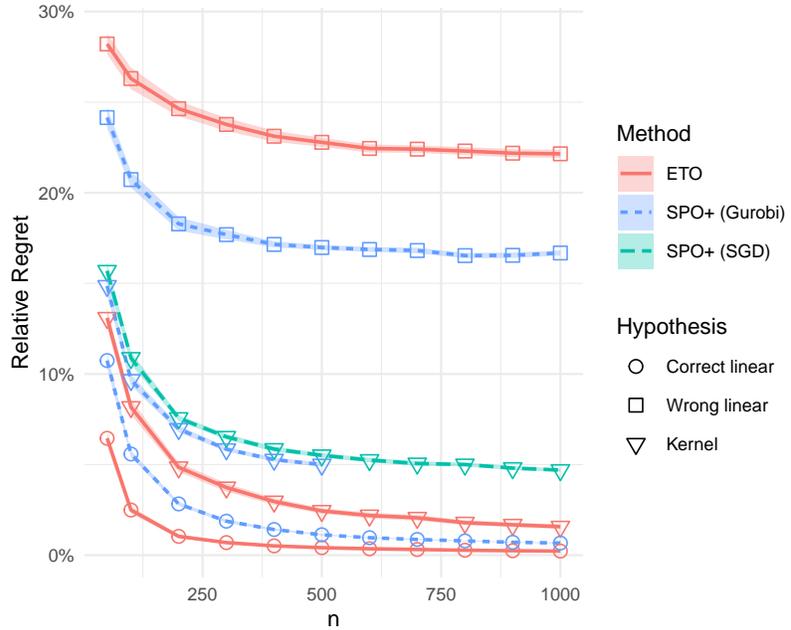
\begin{figure}[t!]%
\begin{minipage}[b]{0.3\textwidth}\centering%
\begin{subfigure}[b]{\textwidth}\centering%
\begin{tikzpicture}[>=latex',font=\footnotesize\selectfont,node distance=1cm, minimum height=0.75cm, minimum width=0.75cm,
state/.style={circle, draw, fill=gray!30, minimum size=15}]
  \foreach \x in {0,...,4}
    \foreach \y in {0,...,4} 
       {
       \node [state]  (\x\y) at (1.1*\x,1.1*\y) {\ifthenelse{\x=0 \AND \y=4}{$s$}{\ifthenelse{\x=4 \AND \y=0}{$t$}{}}};} 
   \foreach \x in {0,...,4}
    \foreach \y [count=\yi] in {0,...,3}  
      {\draw[->] (\x\yi)--(\x\y);
      \draw[->] (\y\x)--(\yi\x);}
    \draw[->] (04)--(14) node [midway, above] {$Y_j$};
    \draw[->] (04)--(14) node [midway, below] {$z_j$};
\end{tikzpicture}\vspace{1.25\baselineskip}
\caption{The CLO instance is a stochastic shortest path problem. We need to go from $s$ to $t$. The random cost of an edge $j$ is $Y_j\in\Rl$. Whether we choose to proceed along an edge $j$ is $z_j\in\{0,1\}$.}
\label{figb}
\end{subfigure}
\end{minipage}\hspace{0.05\textwidth}\begin{minipage}[b]{0.65\textwidth}\centering%
\begin{subfigure}[b]{\textwidth}\centering%
\includegraphics[width=\textwidth]{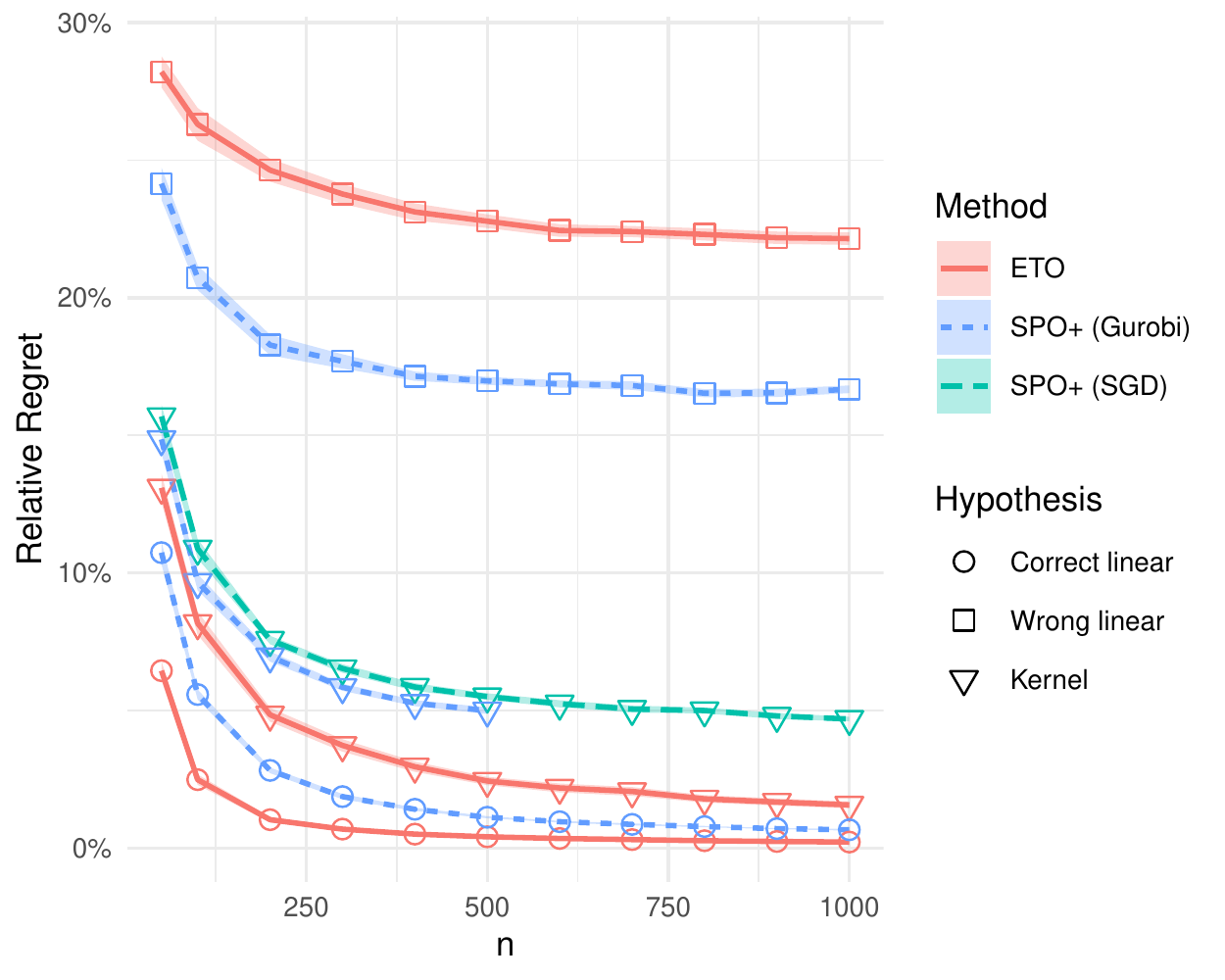}
\caption{\edit{The regret of different methods, relative to average minimal cost. Shaded regions represent 95\% confidence intervals.}}
\label{figa}
\end{subfigure}%
\end{minipage}
\caption{Comparing ETO and SPO+ with well-specified, misspecified, and nonparametric hypotheses.}\label{fig: empirics}
\end{figure}

We next demonstrate these considerations in an experiment. We consider the stochastic shortest path problem shown in \cref{figb}. We aim to go from $s$ to $t$ on a $5\times 5$ grid, and the cost of traveling on edge $j$ is $Y_j$. There are $d=40$ edges, and $\Z$ is given by standard flow preservations constraints, with a source of $+1$ at $s$ and a sink of $-1$ at $t$. We consider covariates with $p=5$ dimensions and $f^*(x)$ being a degree-5 polynomial in $x$, as we detail in \cref{sec:expdetail}.

Ideally we would like to compare ETO to IERM. However, IERM involves a difficult optimization problem that cannot feasibly be solved in practice. We therefore employ the SPO+ loss proposed by \citet{elmachtoub2017smart}, which is a convex surrogate for IERM's objective function. Like IERM, this is still an end-to-end method that integrates estimation and optimization, standing in stark contrast to ETO, which completely separates the two steps.
We consider three different hypothesis classes $\F$ for each of ETO (using least-squares regression, $\hat f_\F$) and SPO+:
\begin{itemize}
\item Correct linear: \edit{$\F$ is as in \cref{ex: linear predictors} with $\phi(x)$ a 31-dimensional basis of monomials spanning $f^*$.}
This represents the unrealistic ideal where we have a perfectly-specified parametric model.
\item Wrong linear: \edit{$\F$ is as in \cref{ex: linear predictors} with $\phi(x)=x \in \R{5}$}. This represents the realistic setting where parametric models are misspecified.
\item Kernel: $\F$ is the RKHS with Gaussian kernel, $\mathcal K(x,x')=\exp(-\rho\magd{x-x'}^2)$. This represents the realistic setting of using flexible, nonparametric models.
\end{itemize}

We employ a ridge penalty in each of the above and choose $\rho$ and this penalty by validation. 
\edit{We use Gurobi to solve the SPO+ optimization problem, except for the RKHS case where due to the heavy computational burden of this we must instead use stochastic gradient descent (SGD) for $n$ larger than $500$. See details in \cref{sec:expdetail}.}
By averaging over $50$ replications of $\D$,
we estimate relative regret, $\E_\D\E_X\bracks{f^*(X)\tr(\hat\pi(X)-\pi^*(X))}/\E_\D\E_X\bracks{f^*(X)\pi^*(X)}$, for each method and each $n=50,100,\dots,1000$, shown in \cref{figa} with shaded bands for plus/minus one standard error.

Although the theoretical results in \cref{sec:slow,sec:fast} do not directly apply to SPO+, our experimental results support our overall insights. With correctly specified models, the ETO method can achieve better performance than end-to-end methods that integrate estimation with optimization (see circle markers for ``Correct linear''). However, from a practical lens, perfectly specified linear models are not realistic. For misspecified linear models, our experiments illustrate how end-to-end methods can account for misspecification to obtain best-in-class performance, beating the corresponding misspecified ETO method (see square markers for ``Wrong linear''). At the same time, we see that such best-in-class performance may sometimes still be bad in an absolute sense. Using more flexible models can sometimes close this gap. The kernel model (triangle markers) is still misspecified in the sense that the RKHS does not contain the true regression function and can only approximate it using functions of growing RKHS norm.
 When using such a flexible model, we observe that ETO achieves regret converging to zero with performance just slightly worse than the correctly-specified case, while end-to-end methods have higher regret. 
Therefore, even though end-to-end methods handle decision-making problems more directly, our experiments demonstrate that the more straightforward ETO approach can be better even in decision-problem performance. 

\section{Concluding Remarks}\label{sec:conclusion}

In this paper we studied the regret convergence rates for two approaches to CLO: the na\"ive, optimization-ignorant ETO and the end-to-end, optimization-aware IERM. We arrived at a surprising fact: 
the convergence rate for ETO is \emph{orders faster} than for IERM, despite \edit{its ignoring the downstream effects of estimation}. We reviewed various reasons for preferring either approach. This highlights a nuanced landscape for the enterprise to integrate estimation and optimization. The practical implications, nonetheless, are positive: relying on regression as a plug-in is easy and fast to run using existing tools, simple to interpret as predictions of uncertain variables, and as our results show it provides downstream decisions with very good performance.
\edit{Beyond providing new insights with practical implications, we hope our work inspires closer investigation of the statistical behavior of data-driven and end-to-end optimization in other settings. \Cref{sec:discussion} points out \emph{nonlinear} CSO as one interesting setting; other settings requiring attention include partial feedback (observe $Y\tr Z$, not $Y$, for historical $Z$), sequential/dynamic optimization problems, and online learning. The unique structure of constrained optimization problems brings up new algorithmic and statistical questions, and the right approach is not always immediately clear, as we showed here for CLO.}

\bibliographystyle{informs2014}
\bibliography{literature}

\newpage
\ECHead{
\begin{center}
$ $\\
Supplemental Material for\\[8pt]
\myfulltitle
\end{center}
}\vspace{8pt}

\begin{APPENDICES}

\section{Proofs}

\subsection{Preliminaries and Definitions}

For any integer $q$, we let $[q]=\{1,\dots,q\}$.

Throughout the following we define $$\Psi(t)=\frac15\exp(t^2).$$
Notice that whenever $\E\Psi(\abs W/w)\leq 1$ for some random variable $W$, we have by Markov's inequality that
\begin{align}\label{eq: prb orlicz}
\Prb{\abs W>t}&\leq 5\exp(-t^2/w^2),\\
\label{eq: l1 orlicz}
\E\abs W&=\int_0^\infty\Prb{\abs W>t}dt\leq 5 w.
\end{align}

We use the shorthand
$$\Esig f(\bm \sigma)=\frac1{2^q}\sum_{\bm \sigma\in\{-1,1\}^q}f(\bm \sigma),$$ where the dimension $q$ is understood from context (will either be $n$ or $nd$, depending on the case). That is, $\Esig$ denotes an expectation over $q$ independent and identically distributed Rademacher random variables independent of all else, in which we marginalize over nothing else (\eg, the data is treated as fixed).

Given a set $\Set\subseteq \R q$ we let $D(\epsilon,\Set)$ be the $\epsilon$-packing number, or the maximal number of elements in $\Set$ that can be taken so that no two are $\epsilon$ close to one another in Euclidean distance, and $N(\epsilon,\Set)$ be the $\epsilon$-covering number, or the minimal number of $\R q$ elements (not necessarily in $\Set$) needed so that every element of $\Set$ is at least $\epsilon$ close to one of these in Euclidean distance. It is immediate \citep[see][Lemma 5.5]{wainwright2019high} that
\begin{equation}\label{eq: packing covering relationship}
N(\epsilon,\Set)\leq D(\epsilon,\Set)\leq N(\epsilon/2,\Set).
\end{equation}

The Natarajan dimension of a \emph{set} $\T\subseteq \Set^q$ (in contrast to a class of functions as in \cref{def: natarajan}) is the largest integer $\eta$ for which there exists $i_1,\dots,i_\eta\in\{1,\dots, q\}$ and $s_1\neq s_1',\dots,s_\eta\neq s_\eta'\in\Set$ such that
$$\edit{
\{(\findic{t_{i_1}=s_1},\dots,\findic{t_{i_\eta}=s_\eta})~:~t\in\T,\,t_{i_1}\in\{s_1,s'_1\},\,\cdots,\, t_{i_\eta}\in\{s_\eta,s'_\eta\}\}\}=\{0,1\}^\eta.
}$$
Thus, the Natarajan dimension of a function class $\G\subseteq[\R p\to\Set]$ is exactly the largest possible Natarajan dimension of $\{(g(x_1),\cdots,g(x_n)):g\in\G\}$ for $x_1,\dots,x_n\in\R p$.

When $\Set\subseteq\Rl$, the pseudo-dimension of $\T$ (also known as its VC-index or VC-subgraph dimension) is the largest integer $\nu$ for which there exists $i_1,\dots,i_\nu\in\{1,\dots, q\}$ and $s_1,\dots,s_\nu\in\Set$ such that
$$
\{(\findic{t_{i_1}\leq s_1},\dots,\findic{t_{i_\nu}\leq s_\nu}):t\in\T\}\}=\{0,1\}^\nu.
$$
The pseudo-dimension (or VC-index or VC-subgraph dimension) of a function class $\G\subseteq[\R p\to\Set]$ is the largest possible pseudo-dimension of $\{(g(x_1),\cdots,g(x_n)):g\in\G\}$ for $x_1,\dots,x_n\in\R p$. Notice that our \cref{def:vc-lin-major} is equivalent to the pseudo-dimension of $\{(\beta,x)\mapsto \beta\tr f(x):f\in\F\}$.

\subsection{Slow Rates for ERM and IERM (\cref{sec: slow erm})}

\proof{Proof of \cref{thm: ERM slow}}
Let
\begin{align*}
h_i(\pi)&=Y_i\tr\pi(X_i),\quad \bm h(\pi)=(h_1(\pi),\dots,h_n(\pi)),\quad\bm H=\{\bm h(\pi):\pi\in\Pi\},\\
L_n(\pi)&=\frac1n\sum_{i=1}^nh_i(\pi),\quad L(\pi)=\E h_1(\pi)=\E[Y\tr \pi(X)]=\E[f^*(X)\tr \pi(X)].
\end{align*}
Notice that all of these but $L(\pi)$ are \emph{random} objects as they depend on the data.

By \citet[Theorem 2.2]{pollard1990empirical},
for any convex, increasing $\Phi$,
\begin{equation}\label{eq: slow erm symmetrize}
\E\Phi\prns{\sup_{\pi\in\Pi}\abs{L_n(\pi)-L(\pi)}}
\leq\E\Esig\Phi\prns{\frac2n\sup_{\bm h\in\bm H}\abs{\ip{\bm\sigma}{\bm h}}}.
\end{equation}

Notice that $\sup_{\pi\in\Pi}\abs{h_i(\pi)}\leq B$ for each $i$.
By \citet[Theorem 3.5]{pollard1990empirical},
\begin{equation}\label{eq: slow erm chain}
\E\Esig\Psi\prns{\frac{1}{nJ}\sup_{\bm h\in\bm H}\abs{\ip{\bm\sigma}{\bm h}}}
\leq1,
\quad\text{where}\quad
J=\frac{9}{n} \int_{0}^{2B\sqrt{n}} \sqrt{\log D(\epsilon, {\bm H})} d\epsilon.
\end{equation}

Let $V$ denote the pseudo-dimension of $\bm H$.
By \cref{eq: packing covering relationship} and \citet[Theorem 2.6.7]{van1996weak}, there exists a universal constant $K_0$ such that
\begin{align*}
D(2B\sqrt{n}\epsilon, {\bm H}) & \le  N(B\sqrt{n}\epsilon, {\bm H})\\
& \le  K_0 (V+1) (16 e)^{V+1} \prns{\frac{2}{\epsilon}}^{2V}.
\end{align*}
Therefore,
\begin{align*}
 J& =  \frac{18B}{\sqrt{n}} \int_{0}^{1} \sqrt{\log D(2B\sqrt{n} \epsilon, {\bm H})} d\epsilon \\
 &\le  \frac{18B}{\sqrt{n}} \int_{0}^{1} \sqrt{\log K_0+ \log(V+1) + (V+1)\log(16e) + 2V\log 2 - 2V\log \epsilon} d\epsilon \\
 &\le    18B\int_{0}^{1} \sqrt{\log K_0+ 3\log 2 + 2\log(16e) - 2\log \epsilon} d\epsilon \sqrt{\frac{V}{n}}\\
 &=  C'B\sqrt{\frac{V}{n}} ,
\end{align*}
where $C' =  18\int_{0}^{1} \sqrt{\log K_0+ 3\log 2 + 2\log(16e) - 2\log \epsilon} d\epsilon<\infty$ is some universal constant.

Combining \cref{eq: slow erm symmetrize,eq: slow erm chain,eq: prb orlicz},
$$
\Prb{\sup_{\pi\in\Pi}\abs{L_n(\pi)-L(\pi)}>t}\leq 5\exp(-nt^2/(VB^2C'^2)).
$$

We now proceed to bound $V$.
Note that $h_i(\pi)$ can only take values in the multiset $\Set_i=\{Y_i^Tz : z \in \Z^\angle\}$, which has cardinality $\abs{\Z^\angle}$.
Let $R_i(\pi)\in\bracks{\abs{\Z^\angle}}$ be the rank of
$h_i(\pi)$ in $\Set_i$, \edit{where we give ties equal rank}, $\bm R(\pi)=(R_1(\pi),\dots,R_n(\pi))$, and
$\tilde{\bm H}=\{\bm R(\pi):\pi\in\Pi\}\subseteq\bracks{\abs{\Z^\angle}}^n$. Then, the pseudo-dimension $\tilde{\bm H}$ is the same as that of $\bm H$, \ie, $V$, and the Natarajan dimension of $\tilde{\bm H}$ is \edit{the same as} the Natarajan dimension of $\bm H$, which is at most $\eta$ by assumption.
By Theorem 10 and Corollary 6 of \citet{bendavid1995characterizations},
\begin{align*}
V\leq 5 \eta \log(\abs{\Z^\angle} +1),
\end{align*}
completing the proof.
\myendproof

\proof{Proof of \cref{prop: n-dim}}
Suppose there exist $x_1, \dots, x_m\in\R p,z_1\neq z'_1,\dots,z_m\neq z'_m\in\Z^\angle$ such that for any $I\subset\{1,\dots,m\}$, some $\pi \in \Pi_{\mathcal F}$ satisfies that
\[
    \pi(x_i) = z_i~~ \forall i \in I, \quad \pi(x_i) = z'_i~~ \forall i \notin I.
\]
For each pair $z_i,z_i'$, let $z_i$ be the one first in the tie-breaking preference ordering.
This must then necessarily mean that there exists some $f \in \mathcal F$ such that
\[
f(x_i)^\top z_i \leq f(x_i)^\top z_i'~~ \forall i \in I, \quad f(x_i)^\top z_i > f(x_i)^\top z_i'~~ \forall i \notin I.
\]
Equivalently, letting $\beta_i=z_i-z_i'$ and $t_i=0$,
\[
{\braces{\prns{
\indic{\beta_i^\top f(x_i)\leq t_i}
}_{i = 1}^m: f\in\mathcal F}} = \{0,1\}^m,
\]
which must mean that $m\leq \nu$.

\myendproof

\proof{Proof of \cref{thm: IERM slow}}
Using the definitions of $L_n,L$ from the proof of \cref{thm: ERM slow} and by optimality of $\hat\pi^\text{IERM}_\F$ for $L_n$ and of $\pi^*$ for $L$, we have
\begin{align*}
L\prns{\hat\pi^\text{IERM}_\F}&\leq L_n\prns{\hat\pi^\text{ERM}_\Pi}+\sup_{\pi\in\Pi_\F}\abs{L(\pi)-L_n(\pi)}\\
&\leq  L_n\prns{\pi^*}+\sup_{\pi\in\Pi_\F}\abs{L(\pi)-L_n(\pi)}\\
&\leq  L\prns{\pi^*}+2\sup_{\pi\in\Pi_\F}\abs{L(\pi)-L_n(\pi)}.
\end{align*}
Applying \cref{thm: ERM slow,prop: n-dim},
we have
$$
\Prb{L\prns{\hat\pi^\text{IERM}_\F}-L\prns{\pi^*}>t}\leq
5\exp(-n(t/2)^2/(CB\nu\log\prns{\abs{\Z^\angle}+1})).
$$
Integrating over $t$ from $0$ to $\infty$, we obtain for another universal constant $C'$ that
$$
\Eb{L\prns{\hat\pi^\text{IERM}_\F}-L\prns{\pi^*}}\leq C'B\sqrt{\frac{\nu\log\prns{\abs{\Z^\angle}+1}}{n}}.
$$
Iterated expectations reveal that the left-hand side is equal to $\op{Regret}(\hat \pi^\text{IERM}_{\F})$, completing the proof.
\myendproof

{\blockedit
\proof{Proof of \cref{thm:lowerboundslow}}

To prove the theorem, we will construct a collection of distributions $\mathbb P$ such that the average regret among them satisfies the lower bound; thus, at least one will satisfy the lower bound.

First we will make some preliminary constructions.
For any $z\in \Z^\angle$, let $\bar z$ be the projection of $z$ onto $\op{conv}(\Z^\angle\backslash\{z\})$, and define
\begin{align*}%
  v(z) = \frac{\bar z-z}{\norm{\bar z-z}}.
\end{align*}
We have
\begin{align} \label{eq: lower bound rho Z}
 \min_{z'\in \Z^\angle, z'\ne z} v(z)\tr  \prns{z' - z}
  =  \min_{z'\in \Z^\angle, z'\ne z} v(z)\tr  \prns{\bar z - z} +  v(z)\tr  \prns{z' - \bar z}
  \ge \rho(\Z),
\end{align}
where the last inequality comes from the definition of $\rho(\Z)$ and the fact that $v(z) \perp \prns{\bar z-z'}$ due to projection.

Since $\Pi$ has Natarajan dimension at least $\eta$, there exist $x_1,\dots,x_\eta\in\R p,\,z\s0_1\neq z\s1_1,\dots,z\s0_\eta\neq z\s1_\eta\in\Z^\angle$ such that, for every $\mathbf{b} = \prns{b_1, \dots, b_{\eta}} \in \braces{0,1}^{\eta}$, there is a $\pi_{\mathbf{b}}\in\Pi$ such that $\pi_{\mathbf{b}}(x_i)=z\s{b_i}_i$ for $i=1,\dots,\eta$.

We now construct a distribution $\mathbb P_{\mathbf b}$ for each $\mathbf{b} \in \braces{0,1}^{\eta}$.
For the marginal distribution of $X$,
we always put equal mass $1/\eta$ at each $x_i$, $i=1,\dots, \eta$.
We next construct the conditional distribution of $Y\mid X=x_i$.
For each $i\in [\eta]$, let
\begin{align*}
  u_{0i} = \frac{\zeta \fprns{v(z\s0_i) + v(z\s1_i)} + v(z\s0_i) - v(z\s1_i) }{2}, \\
  u_{1i} = \frac{\zeta\fprns{v(z\s0_i) + v(z\s1_i)} + v(z\s1_i) - v(z\s0_i) }{2},
\end{align*}
We now construct the conditional distribution of $Y\mid X= x_i$.
Set $\zeta = \sqrt{\eta/n}$ and note $\zeta\in[0,1/2]$ by assumption.
If $b_i = 0$, we let
\begin{align*} %
  Y=
\begin{cases}
  u_{0i},~~ \text{ with probability } (1+\zeta)/2,\\
  u_{1i},~~ \text{ with probability } (1-\zeta)/2,
\end{cases}
\end{align*}
and if $b_i = 1$, we let
\begin{align*} %
  Y=
\begin{cases}
  u_{1i},~~ \text{ with probability } (1+\zeta)/2,\\
  u_{0i},~~ \text{ with probability } (1-\zeta)/2.
\end{cases}
\end{align*}
Since $\norm{v(z\s0_i)}=\norm{v(z\s1_i)}= 1$ by definition, triangle inequality yields $\norm{u_{0i}}\le 1,\,\norm{u_{1i}} \le 1$, and hence the above distribution is on $\Y$.
We then have that,
$$
\E_{\mathbb P_{\mathbf b}}[Y\mid X=x_i]=f_{\mathbf b}(x_i)=\zeta v(z\s{b_i}_i).
$$
By \cref{eq: lower bound rho Z}, the optimal decision at $x_i$ is uniquely $z\s{b_i}_i$. That is, the optimal policy is $\pi_{\mathbf b}$, which is in $\Pi$, as was desired.

For $\hat{\pi} \in \Pi$, define $\hat{\mathbf{b}} \in \{0,1\}^{\eta}$ to be a binary vector whose $i$th element is $\hat{b}_i = \ind\braces{\hat{\pi}(x_i) = z\s1_i}$. Consider a prior on $\mathbf{b}$ such that $b_1, \dots, b_{\eta}$ are i.i.d. and $b_1\sim\op{Ber}(1/2)$.
Let $\op{Regret}_{\mathbf b}(\hat{\pi})$ denote the regret when the data is drawn from $\mathbb P_{\mathbf b}$, the regret satisfies the following inequalities:
\begin{align*}
  \sup_{\mathbf{b}\in\{0,1\}^{\eta} } \op{Regret}_b(\hat{\pi}) 
   &=
  \sup_{\mathbf{b}\in\{0,1\}^{\eta} }\expect_{\mathbb{P}^n_{\mathbf{b}}} \expect_{X} \bracks{f_{\mathbf b}(X)\tr  \prns{\hat{\pi}(X) - \pi_{\mathbf b}(X)}}\\
  &\ge  \expect_{\mathbf{b}} \expect_{\mathbb{P}^n_{\mathbf{b}}} \expect_X \bracks{f_{\mathbf b}(X)\tr  \prns{\hat{\pi}(X) - \pi_{\mathbf b}(X)}} \\
 & =  \expect_{\mathbf{b}} \expect_{\mathbb{P}^n_{\mathbf{b}}} \expect_X \bracks{f_{\mathbf b}(X)\tr  \prns{\hat{\pi}(X) - \pi_{\mathbf b}(X)}\ind\bracks{\hat{\pi}(X) \ne \pi_{\mathbf b}(X)}} \\
 & \ge  \zeta \rho(\Z) \expect_{\mathbf{b}} \expect_{\mathbb{P}^n_{\mathbf{b}}} \expect_X \bracks{\ind\bracks{\hat{\pi}(X) \ne \pi_{\mathbf b}(X)}} \\
 & =    \frac{\zeta \rho(\Z)}{\eta} \sum_{i=1}^{\eta}\expect_{\mathbf{b}} \expect_{\mathbb{P}^n_{\mathbf{b}}} \bracks{\ind\bracks{\hat{\pi}(x_i) \ne \pi_{\mathbf b}(x_i)}}   \\
 & \ge  \frac{\zeta \rho(\Z)}{\eta} \sum_{i=1}^{\eta}\expect_{\mathbf{b}} \expect_{\mathbb{P}^n_{\mathbf{b}}}\bracks{\ind\bracks{b_i \ne \hat{b}_i}}   ,
\end{align*}
where the second inequality comes from \cref{eq: lower bound rho Z}, and the third inequality comes from the fact that $b_i \ne \hat{b}_i$ implies $\hat{\pi}(x_i) \ne \pi_{\mathbf b}(x_i)$.

The term $\expect_{\mathbf{b}} \expect_{\mathbb{P}^n_{\mathbf{b}}}\bracks{\ind\bracks{b_i \ne \hat{b}_i}}$ above is the Bayes risk of the algorithm $\mathcal D\mapsto \hat b_i$ with respect to the loss function being the misclassification error of the random bit $b_i$. 
We will now lower bound this by computing the minimum Bayes risk.
Letting $\tilde{\mathbb P}$ denote the distribution of $(\mathbf b,\mathcal D)$ when we draw $b_i$ as iid Bernoulli as above and then data as $(\mathcal{D}\mid \mathbf b)\sim\mathbb P_{\mathbf b}^n$. 
Then $\tilde{\mathbb P}\fprns{b_i = 1\mid {D}}$ is the posterior probability that $b_i = 1$ and the minimum Bayes risk is simply $\min\braces{\tilde{\mathbb P}\fprns{b_i = 1\mid {\mathcal{D}}}, 1- \tilde{\mathbb P}\fprns{b_i = 1\mid {\mathcal{D}}}}$. We conclude that
\begin{align} \label{eq: min bayes risk}
   \frac{\zeta \rho(\Z)}{\eta} \sum_{i=1}^{\eta}\expect_{\mathbf{b}} \expect_{\mathbb{P}^n_{\mathbf{b}}}\bracks{\ind\bracks{b_i \ne \hat{b}_i}} \geq
   \frac{\zeta \rho(\Z)}{\eta} \sum_{i=1}^{\eta} \expect_{\tilde{\mathbb P}} \bracks{\min\braces{\tilde{\mathbb P}\fprns{b_i = 1\mid {\mathcal{D}}}, 1- \tilde{\mathbb P}\fprns{b_i = 1\mid {\mathcal{D}}}}}.
\end{align}

We proceed to calculate the latter.
For $i \in [\eta]$, let
\begin{align*}
  N_i^0 = \sum_{j=1}^n \ind\bracks{X_j = x_i, Y_j = u_{0i}},\quad
  N_i^1 = \sum_{j=1}^n \ind\bracks{X_j = x_i, Y_j = u_{1i}}.
\end{align*}
We can then write the posterior distribution as
\begin{align*}
  \tilde{\mathbb P}\prns{b_i = 1\mid \mathcal{D}} = \frac{\prns{\frac{1+\zeta}{2}}^{N_i^1} \prns{\frac{1-\zeta}{2}}^{N_i^0}}{ \prns{\frac{1+\zeta}{2}}^{N_i^1} \prns{\frac{1-\zeta}{2}}^{N_i^0} + \prns{\frac{1+\zeta}{2}}^{N_i^0} \prns{\frac{1-\zeta}{2}}^{N_i^1}}.
\end{align*}
Hence,
\begin{align*}
  \min\braces{\tilde{\mathbb P}\prns{b_i = 1\mid {\mathcal{D}}}, 1- \tilde{\mathbb P}\prns{b_i = 1\mid {\mathcal{D}}}} 
  &= \frac{\min \braces{\prns{\frac{1+\zeta}{2}}^{N_i^1} \prns{\frac{1-\zeta}{2}}^{N_i^0}, \prns{\frac{1+\zeta}{2}}^{N_i^0} \prns{\frac{1-\zeta}{2}}^{N_i^1}} }{ \prns{\frac{1+\zeta}{2}}^{N_i^1} \prns{\frac{1-\zeta}{2}}^{N_i^0} + \prns{\frac{1+\zeta}{2}}^{N_i^0} \prns{\frac{1-\zeta}{2}}^{N_i^1}} \\
&=  \frac{1}{ 1 + \prns{\frac{1+\zeta}{1-\zeta}}^{\abs{N_i^1 - N_i^0}}}.
\end{align*}
Therefore,
\begin{align*}
  \frac{\zeta \rho(\Z)}{\eta} \sum_{i=1}^{\eta} \expect_{\tilde{\mathbb P}} \bracks{\min\braces{\tilde{\mathbb P}\fprns{b_i = 1\mid {\mathcal{D}}}, 1- \tilde{\mathbb P}\fprns{b_i = 1\mid {\mathcal{D}}}}}
  &=  \frac{\zeta \rho(\Z)}{\eta} \sum_{i=1}^{\eta} \expect_{\tilde{\mathbb{P}}} \bracks{\prns{ 1 + \prns{\frac{1+\zeta}{1-\zeta}}^{\abs{N_i^1 - N_i^0}}}^{-1}} \\
  &\ge  \frac{\zeta \rho(\Z)}{2\eta} \sum_{i=1}^{\eta} \expect_{\tilde{\mathbb{P}}} \bracks{\prns{\frac{1+\zeta}{1-\zeta}}^{-\abs{N_i^1 - N_i^0}}} \\
  &\ge  \frac{\zeta \rho(\Z)}{2\eta} \sum_{i=1}^{\eta}  \prns{\frac{1+\zeta}{1-\zeta}}^{-\expect_{\tilde{\mathbb{P}}}\abs{N_i^1 - N_i^0}},
\end{align*}
where the first inequality is due to the fact that $\frac{1+\zeta}{1-\zeta} \ge 1$ and the second inequality follows from Jensen's inequality.
Given our symmetric prior distribution on $\mathbf{b}$, the marginal distribution of $Y_j$ given $X_j = x_i$ is $\tilde{\mathbb{P}}\prns{Y_j = u_{0i} \mid X_j = x_i} = \tilde{\mathbb{P}}\prns{Y_j = u_{1i} \mid X_j = x_i} = 1/2$.
Thus, letting $\op{Bin}\prns{k, \frac{1}{2}}$ represent a binomial random variable with parameters $k$ and $1/2$, we have
\begin{align*}
  \expect_{\tilde{\mathbb{P}}}\abs{N_i^1 - N_i^0} 
  &=  \sum_{k = 0}^{n} {n \choose k} \prns{\frac{1}{\eta}}^k \prns{1-\frac{1}{\eta}}^{n-k} \expect \abs{2\op{Bin}\prns{k, \frac{1}{2}} - k}\\
  &\le  \sum_{k = 0}^{n} {n \choose k} \prns{\frac{1}{\eta}}^k \prns{1-\frac{1}{\eta}}^{n-k} \sqrt{\expect \prns{2\op{Bin}\prns{k, \frac{1}{2}} - k}^2}\\
  &=  \sum_{k = 0}^{n} {n \choose k} \prns{\frac{1}{\eta}}^k \prns{1-\frac{1}{\eta}}^{n-k} \sqrt{k}\\
  &=  \expect \sqrt{\op{Bin}\prns{n, \frac{1}{\eta}}}\\
  &\le  \sqrt{\frac{n}{\eta}},
\end{align*} 
where the two inequalities follow from the Cauchy-Schwarz inequality.

Putting our calculations together, we get
\begin{align} \label{eq: lower bound with zeta}
  \sup_{\mathbf{b}\in\{0,1\}^n } \op{Regret}_b(\hat{\pi})
  \ge \frac{\zeta \rho(\Z)}{2}  \prns{\frac{1+\zeta}{1-\zeta}}^{-\sqrt{n/\eta}}
  \ge \frac{\zeta \rho(\Z)}{2}  \exp\prns{-\frac{2\zeta}{1-\zeta} \sqrt{\frac{n}{\eta}}},
\end{align}
where the second inequality follows from the fact that $1+x\le e^x$ for any $x\in \mathbb{R}$.
Finally, plugging in $\zeta = \sqrt{\eta/n}$ and noting that $\zeta\leq1/2$ since $n\ge 4\eta$, we have
\begin{align*}
  \frac{\zeta \rho(\Z)}{2}  \exp\prns{-\frac{2\zeta}{1-\zeta} \sqrt{\frac{n}{\eta}}} 
  = \frac{\rho(\Z)}{2}\sqrt{\frac{\eta}{n}}\exp\prns{-\frac{2}{1-\zeta}}
  \ge \frac{\rho(\Z)}{2e^4}\sqrt{\frac{\eta}{n}},
\end{align*}
as desired.
\myendproof}

\subsection{Slow Rates for ETO (\cref{sec: slow eto})}

The proof of \cref{thm: MSE ERM} is very involved and is therefore relegated to its own \cref{sec: MSE ERM proof}.

\proof{Proof of \cref{thm: slowrate generic}}
By optimality of $\pi_{\hat f}$ with respect to $\hat f$, we have that
\begin{align*}
\op{Regret}(\pi_{\hat{f}})=&~\Eb{f^*(X)\tr (\pi_{\hat{f}}(X) - \pi^*(X))}\\
  \le &~ \Eb{f^*(X)\tr \pi_{\hat{f}}(X) -\hat{f}(X)\tr\pi_{\hat{f}}(X) + \hat{f}(X)\tr\pi^*(X) - f^*(X)\tr\pi^*(X)}\\
  \le &~ 2B\Eb{\|f^*(X)- \hat{f}(X) \|}.
\end{align*}
\myendproof

\proof{Proof of \cref{thm:slowrate}}
The result follows by integrating the tail bound from \cref{thm: MSE ERM} to bound the expected error and invoking \cref{thm: slowrate generic}.
\myendproof

\subsection{Fast Rates for ERM and IERM (\cref{sec: fast erm})}

\subsubsection{Preliminaries and Definitions.}

For any policy $\pi, \pi' \in [\R p\to\Z^\angle]$, define
\begin{align*}
    & d(\pi, \pi')  = \edit{\frac{1}{B}}\expect_X[f^*(X)^T (\pi'(X) - \pi(X))], \\
    & d_{\Delta}(\pi, \pi')  = \pr_X(\pi(X) \neq \pi'(X)).
\end{align*}

In this section, we let $\expect_P$ be the expectation with respect to ${\mathbb P}_{X,Y}$, $\expect_{\mathcal D}$ the expectation with respect to the sampling of data $\mathcal D$, and $\expect_n$ the expectation with respect to the empirical distribution.
Moreover, for any function $h(x,y)$, we define $||h||_{L_2(P)} = \sqrt{\expect_P [h^2(X,Y)]}$.

\subsubsection{Supporting Lemmas.}

We first show that $d$ and $d_{\Delta}$ have the following relationship:
\begin{lemma}\label{lemma:ddelta}
Suppose \cref{asm:margin} holds and $\Prb{\abs{\Z^*(X)}>1}=0$. Then
\begin{align*}
    & d(\pi^*, \pi) \le \edit{2} d_{\Delta}(\pi^*, \pi), \\
    & d_{\Delta}(\pi^*, \pi) \le c_1 d(\pi^*, \pi)^{\frac{\alpha}{\alpha+1}},
\end{align*}
where $c_1 = (\alpha\gamma^\alpha)^{-\frac{\alpha}{\alpha+1}} (\alpha+1)\gamma^\alpha$.
\end{lemma}

\proof{Proof of \cref{lemma:ddelta}}
First of all,
\begin{align*}
    d(\pi^*, \pi) = & \edit{\frac{1}{B}}\expect_X[f^*(X)^T (\pi(X) - \pi^*(X)) \ind\{\pi(X) \neq \pi^*(X)\}]\\
    \le & \edit{2} \pr_X(\pi(X) \neq \pi^*(X)) \\
    = & \edit{2} d_{\Delta}(\pi^*, \pi).
\end{align*}

Now we prove the second statement.
For any $t>0$, we have
\begin{align*}
    d(\pi^*, \pi) = & \edit{\frac{1}{B}}\expect_X[f^*(X)^T (\pi(X) - \pi^*(X)) \ind\{\pi(X) \neq \pi^*(X)\}]\\
    \ge & \edit{\frac{1}{B}}\expect_X[f^*(X)^T (\pi(X) - \pi^*(X)) \ind\{\pi(X) \neq \pi^*(X), \Delta(X)>t\edit{B}\}]\\
    \ge & t \pr_X(\pi(X) \neq \pi^*(X), \Delta(X)>t\edit{B})\\
    = & t [d_{\Delta}(\pi^*, \pi) - \pr_X(\pi(X) \neq \pi^*(X), \Delta(X)\le t\edit{B})]\\
    \ge & t [d_{\Delta}(\pi^*, \pi) - \pr_X(\Delta(X)\le t\edit{B})]\\
    \ge & t [d_{\Delta}(\pi^*, \pi) - \gamma^\alpha t ^{\alpha}].
\end{align*}
If we take $t = ((\alpha+1)\gamma^\alpha)^{-1/\alpha} [d_{\Delta}(\pi^*, \pi)]^{1/\alpha}$, we have
\begin{align*}
    d(\pi^*, \pi) \ge \alpha\gamma^\alpha ((\alpha+1) \gamma^\alpha)^{-\frac{\alpha+1}{\alpha}} d_{\Delta}(\pi^*, \pi)^{\frac{\alpha+1}{\alpha}}.
\end{align*}
Therefore,
\begin{align*}
    d_{\Delta}(\pi^*, \pi) \le (\alpha\gamma^\alpha)^{-\frac{\alpha}{\alpha+1}} (\alpha+1)\gamma^\alpha d(\pi^*, \pi)^{\frac{\alpha}{\alpha+1}}.
\end{align*}
\myendproof

We will also need the following concentration inequality due to \citet{bousquet2002bennett}.
\begin{lemma}\label{lemma:concentration}
  Let $\mathcal H$ be a countable family of measurable functions such that $\sup_{h\in \mathcal H} E_P(h^2) \le \delta^2$ and $\sup_{h\in \mathcal H} ||h||_{\infty} \le \bar{H}$ for some constants $\delta$ and $\bar{H}$. Let $S = \sup_{h\in \mathcal H} (\expect_n(h) - \expect_P(h))$. Then for every $t>0$,
  \begin{align*}
    \pr \prns{S - \expect(S) \ge \sqrt{\frac{2(\delta^2 + 4\bar{H}\expect(S)) t}{n}} + \frac{2\bar{H}t}{3n}} \le \exp(-t).
  \end{align*}
\end{lemma}
Note the restriction on \emph{countable} $\mathcal H$. If an uncountable $\mathcal H$, however, satisfies $\sup_{h\in \mathcal H} E_P(h^2) \le \delta^2$ and $\sup_{h\in \mathcal H} ||h||_{\infty} \le \bar{H}$ and is separable with respect with respect to $\max\{L_2(P),L_2(P_n)\}$ then we can just take a dense countable subset, apply \cref{lemma:concentration}, and obtain the result for the uncountable $\mathcal H$, since the random variable $S$ would be unchanged. In particular, we have the above separability if $\mathcal H$ has finite packing numbers with respect to $L_2(Q)$ for any $Q$ because we can simply take the union of $(1/k)$-maximal-packings with respect to $L_2((P_n+P)/2)$ for $k=1,2,\dots$ (note we take the packings and not coverings to ensure the points are inside the set). In the below, our set $\mathcal H$ has a finite pseudo-dimension and therefore finite packing numbers with respect to $L_2(Q)$ for any $Q$ \citep[Theorem 2.6.7]{van1996weak}.

Finally, the following lemma bounds the mean of a supremum of a centered empirical process indexed by functions with bounded $L_2(P)$ norm.
\begin{lemma} \label{lemma:uniformbound}
  Suppose $\Pi\subseteq[\R p\to\Z^\angle]$ has Natarajan dimension at most $\eta$.
  Define a class of functions indexed by $\pi\in \Pi$:
  \begin{align*}
    \mathcal H_{\delta} = \{h(X, Y; \pi) = \edit{\frac{1}{B}\prns{Y^T \pi^*(X) - Y^T \pi(X)}} : \pi\in \Pi, ||h||_{L_2(P)} \le \delta\}.
  \end{align*}
There exists a universal constant $C_0$ such that for any \edit{$n\ge \frac{20 C_0^2 \eta \log(\abs{\Z^\angle}^2 +1)  \log (n+1)}{\delta^2}$},
  \begin{align*}
    \expect_{\mathcal D}[\sup_{h\in \mathcal H_{\delta}} (\expect_n(h) - \expect_P(h))] \le (1+\sqrt{2}) C_0 \sqrt{\frac{5 \eta \log(\abs{\Z^\angle}^2 +1)\log (n+1)}{n}}\delta.
  \end{align*}
\end{lemma}

\proof{Proof of \cref{lemma:uniformbound}}
Fix $(X_1, Y_1), \dots, (X_n, Y_n)$.
Define ${\bm h}(\pi) = (h(X_1, Y_1;\pi), \dots, h(X_n, Y_n; \pi))$ and ${\bm H}_{\delta} = \{{\bm h}(\pi): h(\cdot; \pi)\in \mathcal H_{\delta}\} \subseteq \mathbb{R}^n$.
Let $V$ denote the pseudo-dimension of ${\bm H}_{\delta}$.
Let $\delta_n = \frac{1}{\sqrt{n}}\sup_{{\bm h}^{\pi} \in {\bm H}_{\delta}} \magd{{\bm h}^{\pi}}$ and $H_{\delta}$ be the envelope of ${\bm H}_{\delta}$. We have $||H_{\delta}|| \le n \delta_n$.
By \citet[Theorem 2.2]{pollard1990empirical},
\begin{align}
  \expect_{\mathcal D}[\sup_{h\in \mathcal H_{\delta}} (\expect_n(h) - \expect_P(h))] \le   \expect_{\mathcal D} \expect_{ \sigma} \bigg[\frac{2}{n}\sup_{\bm h\in{\bm H}_{\delta}}\abs{\ip{\bm\sigma}{\bm h}}\bigg]. \label{eq: fast erm symmetrize}
\end{align}
By \citet[Theorem 3.5]{pollard1990empirical},
\begin{equation}\label{eq: fast erm chain}
\Esig\Psi\prns{\frac{1}{nJ}\sup_{\bm h\in{\bm H}_{\delta}}\abs{\ip{\bm\sigma}{\bm h}}}
\leq1,
\quad\text{where}\quad
J=\frac{9}{n} \int_{0}^{\sqrt{n}\delta_n} \sqrt{\log D(\epsilon, {\bm H}_{\delta})} d\epsilon.
\end{equation}
By \cref{eq: packing covering relationship} and \citet[Theorem 2.6.7]{van1996weak},
there exists a universal constant $K_0$ such that
\begin{align*}
D(\sqrt{n}\delta_n x, {\bm H}_{\delta}) \le & N \prns{\frac{1}{2}\sqrt{n}\delta_n x, {\bm H}_{\delta}}\\
\le & N\prns{\frac{x}{2\sqrt{n}}||H_{\delta}||, {\bm H}_{\delta}} \\
\le & K_0 (V+1) (16 e)^{V+1} \prns{\frac{2\sqrt{n}}{x}}^{2V}.
\end{align*}
Therefore,
\begin{align*}
J = & \frac{9}{\sqrt{n}} \int_{0}^{1} \delta_n\sqrt{\log D(\sqrt{n}\delta_n x, {\bm H}_{\delta})} dx \\
 \le & \frac{9}{\sqrt{n}} \int_{0}^{1} \delta_n\sqrt{\log K_0 + \log(V+1) + (V+1)\log(16e) + V\log n + 2V\log 2 - 2V\log x} dx \\
 \le &   9\int_{0}^{1} \sqrt{2\log K_0+ 4 + 4\log(16e)  - 4\log x} dx \sqrt{\frac{V\log (n+1)}{n}} \delta_n\\
 = & \frac{C_0}{10}\sqrt{\frac{V\log (n+1)}{n}} \delta_n,
\end{align*}
where $C_0 =  90\int_{0}^{1} \sqrt{2\log K_0+ 4 + 4\log(16e) - 4\log x} dx<\infty$.
By \cref{eq: l1 orlicz},
\begin{align*}
  \expect_{ \sigma} \bigg[\frac{1}{n}\sup_{\bm h\in{\bm H}_{\delta}}\abs{\ip{\bm\sigma}{\bm h}}\bigg] \le \frac{C_0}{2}\sqrt{\frac{V\log (n+1)}{n}} \delta_n.
\end{align*}
and combining \cref{eq: fast erm symmetrize,eq: fast erm chain} we get
\begin{align}
  \expect_{\mathcal D}[\sup_{h\in \mathcal H_{\delta}} (\expect_n(h) - \expect_P(h))] \le & C_0 \sqrt{\frac{V\log (n+1)}{n}}  \expect_{\mathcal D}(\delta_n)\notag \\
  = & C_0 \sqrt{\frac{V\log (n+1)}{n}} \expect_{\mathcal D}\bigg(\bigg[\sup_{h\in \mathcal H_{\delta}} \expect_n (h^2)\bigg]^{1/2}\bigg)\notag \\
  \le & C_0 \sqrt{\frac{V\log (n+1)}{n}} \bigg(\expect_{\mathcal D}\bigg[\sup_{h\in \mathcal H_{\delta}} \expect_n (h^2)\bigg]\bigg)^{1/2}.  \label{eq: expect bound}
\end{align}
Note that $\expect_n (h^2)$ can be bounded by
\begin{align*}
  \expect_n (h^2) =&  \expect_n (h^2 - \expect_P(h^2)) + \expect_P(h^2) \\
  = & \expect_n ((h - ||h||_{L_2(P)})(h+||h||_{L_2(P)})) + ||h||_{L_2(P)}^2 \\
  \le & \edit{4}\expect_n (h - ||h||_{L_2(P)}) + \delta^2 \\
  \le & \edit{4}\expect_n (h - \expect_P(h)) + \delta^2.
\end{align*}
Combining with \cref{eq: expect bound} we get
\begin{align*}
\expect_{\mathcal D}[\sup_{h\in \mathcal H_{\delta}} (\expect_n(h) - \expect_P(h))] \le C_0 \sqrt{\frac{V\log (n+1)}{n}} \sqrt{\edit{4}\expect_{\mathcal D}[\sup_{h\in \mathcal H_{\delta}} (\expect_n(h) - \expect_P(h))] + \delta^2}.
\end{align*}
Solving this inequality for $\expect_{\mathcal D}[\sup_{h\in \mathcal H_{\delta}} (\expect_n(h) - \expect_P(h))]$ we get
\begin{align*}
\expect_{\mathcal D}[\sup_{h\in \mathcal H_{\delta}} (\expect_n(h) - \expect_P(h))] \le  \edit{2} C_0^2 \sqrt{\frac{V\log (n+1)}{n}} \prns{\sqrt{\frac{V\log (n+1)}{n}} + \sqrt{\frac{V\log (n+1)}{n} + \frac{\delta^2}{\edit{4}C_0^2}}} .
\end{align*}
When $\frac{V\log (n+1)}{n} \le \frac{\delta^2}{\edit{4} C_0^2}$, i.e., when $n \ge \frac{\edit{4} C_0^2 V\log (n+1)}{\delta^2}$, we have
\begin{align*}
\expect_{\mathcal D}[\sup_{h\in \mathcal H_{\delta}} (\expect_n(h) - \expect_P(h))] \le   (1+\sqrt{2})C_0  \sqrt{\frac{V\log (n+1)}{n}}\delta .
\end{align*}

Finally, by similar arguments as in the proof of \cref{thm: ERM slow},
\begin{align*}
V \le 5 \eta \log \prns{\abs{\Z^\angle}^2 +1},
\end{align*}
completing the proof.
\myendproof

\subsubsection{Proof of \cref{thm:fastrateerm} and \cref{thm:fastrate IERM}.}
\proof{Proof of \cref{thm:fastrateerm}}
Note that
\begin{align*}
 d(\pi, \pi') = \edit{\frac{1}{B}}\expect_P[Y\tr (\pi'(X)-\pi(X)],
\end{align*}
and define
\begin{align*}
  \mathcal H_{\Pi} = \{h(X, Y; \pi) = \edit{\frac{1}{B}\prns{Y\tr \pi^*(X) - Y\tr \pi(X)}} : \pi\in \Pi\}.
\end{align*}
Because $\magd Y\leq 1,\,\sup_{z\in\Z}\magd z\leq B$, $\mathcal H_{\Pi}$ has envelope \edit{$2$}. Besides, we can write $d(\pi^*, \pi) = -\expect_P(h(X, Y; \pi))$, and we know that $-\expect_P(h) \ge 0$ for all $h \in \mathcal H_{\Pi}$.
Moreover, we can define the sample analogue of $d(\pi, \pi')$ as
\begin{align*}
 \edit{d_n(\pi, \pi') =\frac{1}{nB} \sum_{i=1}^n Y_i \tr \prns{\pi'(X_i) -\pi(X_i)}.}  
\end{align*}

Let $a = \sqrt{\kappa t}\epsilon_n$ with $\kappa\ge 1, t\ge 1$, and $\epsilon_n>0$, where $t\ge 1$ is arbitrary, $\kappa$ is a constant that we choose later, and $\epsilon_n$ is a sequence indexed by sample size $n$ whose proper choice will be discussed in a later step.
Define
\begin{align*}
V_a = \sup_{h\in \mathcal H_{\Pi}} \prns{\frac{\expect_n(h) - \expect_P(h)}{-\expect_P(h) + a^2}} =  \sup_{h\in \mathcal H_{\Pi}} \prns{\expect_n \prns{\frac{h}{-\expect_P(h) + a^2}} - \expect_P \prns{\frac{h}{-\expect_P(h) + a^2}}}.
\end{align*}
By definition $\hat{\pi}_{\Pi}^{ERM} = \arg\min_{\pi\in \Pi} \frac{1}{n}\sum_{i=1}^n Y_i\tr\pi(X_i)$.
Since $\pi^* \in \Pi$, $d_n(\pi^*, \hat{\pi}_{\Pi}^{ERM}) \le 0$ and
\begin{align*}
d(\pi^*, \hat{\pi}_{\Pi}^{ERM}) \le & d(\pi^*, \hat{\pi}_{\Pi}^{ERM}) - d_n(\pi^*, \hat{\pi}_{\Pi}^{ERM})\\
= & \expect_n(h(X, Y; \hat{\pi}_{\Pi}^{ERM}))-\expect_P(h(X, Y; \hat{\pi}_{\Pi}^{ERM})) \\
\le & V_a[d(\pi^*, \hat{\pi}_{\Pi}^{ERM})  + a^2].
\end{align*}
On the event $V_a<1/2$, we have $d(\pi^*, \hat{\pi}_{\Pi}^{ERM})< a^2$ holds, which implies
\begin{align}
  \pr(d(\pi^*, \hat{\pi}_{\Pi}^{ERM})\ge a^2) \le \pr(V_a\ge 1/2). \label{eq: d and va}
\end{align}
In what follows, we aim to prove that $\pr(V_a\ge 1/2) \le \exp(-t)$.

First of all, note that for all $h \in \mathcal H_{\Pi}$,
\begin{align*}
  \expect_P \prns{\prns{\frac{h}{-\expect_P(h) + a^2}}^2} \le & \frac{ \edit{4}d_{\Delta}(\pi^*, \pi)}{(-\expect_P(h) + a^2)^2} \\
  \le & \edit{4c_1}\frac{ (-\expect_P(h))^{\frac{\alpha}{1+ \alpha}}}{(-\expect_P(h) + a^2)^2} \\
  \le & \edit{4c_1} \sup_{\epsilon\ge 0} \frac{\epsilon^{\frac{2\alpha}{1+ \alpha}}}{(\epsilon^2 + a^2)^2} \\
  \le & \edit{4c_1} \frac{1}{a^2}\sup_{\epsilon\ge 0} \frac{\epsilon^{\frac{2\alpha}{1+ \alpha}}}{\epsilon^2 + a^2}\\
  \le & \edit{4c_1} \frac{1}{a^2}\sup_{\epsilon\ge 0} \prns{\frac{\epsilon^{\frac{\alpha}{1+ \alpha}}}{\epsilon \vee a}}^2\\
  = & \edit{4c_1} a^{\frac{2\alpha}{1+ \alpha} -4}.
\end{align*}
where $c_1$ is the constant in \cref{lemma:ddelta}. Moreover,
\begin{align*}
\sup_{h\in \mathcal H_{\Pi}}\magd{\frac{h}{-\expect_P(h) + a^2}}_{\infty} \le \frac{\edit{2}}{a^2}.
\end{align*}
By \cref{lemma:concentration},
\begin{align}
\pr \prns{V_a \le \expect(V_a) + \edit{\sqrt{\frac{8(c_1 a^{\frac{2\alpha}{1+ \alpha} -2} + 2\expect(V_a)) t}{a^2 n}} + \frac{4t}{3a^2n}} }  \ge 1-\exp(-t). \label{eq: high prob Va}
\end{align}

We now aim to prove an upper bound on $\expect(V_a)$.
Let $r>1$ be arbitrary and partition $\mathcal H_{\Pi}$ by $\mathcal H_0, \mathcal H_1, \dots$ where $\mathcal H_0 = \{h\in \mathcal H_{\Pi}: -\expect_P(h) \le a^2\}$ and $\mathcal H_j = \{h\in \mathcal H_{\Pi}:  r^{2(j-1)}a^2< -\expect_P(h) \le r^{2j}a^2\}$ for $j\ge 1$. Then,
\begin{align}
V_a \le & \sup_{h\in\mathcal H_0} \prns{\frac{\expect_n(h) - \expect_P(h)}{-\expect_P(h) + a^2}}+ \sum_{j\ge 1}\sup_{h\in\mathcal H_j} \prns{\frac{\expect_n(h) - \expect_P(h)}{-\expect_P(h) + a^2}} \notag \\
\le & \frac{1}{a^2}\bigg[\sup_{h\in\mathcal H_0} \{\expect_n(h) - \expect_P(h)\} + \sum_{j\ge 1} (1+ r^{2(j-1)})^{-1}\sup_{h\in\mathcal H_j} \{\expect_n(h) - \expect_P(h)\} \bigg]\notag \\
\le & \frac{1}{a^2}\bigg[\sup_{- \expect_P(h)\le a^2} \{\expect_n(h) - \expect_P(h)\} + \sum_{j\ge 1} (1+ r^{2(j-1)})^{-1}\sup_{- \expect_P(h)\le r^{2j}a^2} \{\expect_n(h) - \expect_P(h)\} \bigg]. \label{eq: va}
\end{align}
By \cref{lemma:ddelta},
\begin{align*}
  ||h||^2_{L_2(P)} = \expect_P(h^2) \le \edit{4} d_{\Delta} (\pi^*, \pi) \le \edit{4}c_1[-\expect_P(h)]^{\frac{\alpha}{1+\alpha}},
\end{align*}
so we know that $- \expect_P(h)\le r^{2j}a^2$ implies $||h||_{L_2(P)}\le \edit{2}c_1^{1/2}r^{\frac{\alpha}{1+ \alpha}j}a^{\frac{\alpha}{1+ \alpha}}$.
Thus, \edit{\cref{eq: va}} can be further bounded by
\begin{align*}
V_a \le    \frac{1}{a^2}\bigg[\sup_{||h||_{L_2(P)}\le \edit{2}c_1^{1/2}a^{\frac{\alpha}{1+ \alpha}}} \{\expect_n(h) - \expect_P(h)\} + \sum_{j\ge 1} (1+ r^{2(j-1)})^{-1}\sup_{||h||_{L_2(P)}\le\edit{2} c_1^{1/2}r^{\frac{\alpha}{1+ \alpha}j}a^{\frac{\alpha}{1+ \alpha}}} \{\expect_n(h) - \expect_P(h)\} \bigg].
\end{align*}
For the rest of the proof, we let $\bar V = 5 \eta \log(\abs{\Z^\angle}^2 +1)$ for notational simplicity.
By \cref{lemma:uniformbound},
\begin{align*}
\expect_{\mathcal D}[V_a] \le &  \edit{2}(1+\sqrt{2}) C_0 c_1^{1/2}  \sqrt{\frac{\bar V\log (n+1)}{n}} a^{\frac{\alpha}{1+ \alpha}-2}\bigg[1 + \sum_{j\ge 1} (1+ r^{2(j-1)})^{-1}  r^{\frac{\alpha}{1+ \alpha}j}\bigg]\\
\le & \edit{2}(1+\sqrt{2}) C_0 c_1^{1/2}  \sqrt{\frac{\bar V\log (n+1)}{n}} a^{\frac{\alpha}{1+ \alpha}-2} \bigg(\frac{r^2}{1-r^{\frac{2+\alpha}{1+\alpha}}}\bigg) \\
\le & c_2 \sqrt{\frac{\bar V\log (n+1) }{n}} a^{\frac{\alpha}{1+ \alpha}-2}
\end{align*}
for
\begin{align}
  n \ge \frac{\edit{4} C_0^2\bar V\log (n+1)}{c_1 a^{\frac{2\alpha}{1+ \alpha}}} \Longleftrightarrow
  a \ge  \bigg(\frac{\edit{4} C_0^2}{c_1}\bigg)^{\frac{1+\alpha}{2\alpha}} \bigg(\frac{\bar V\log (n+1) }{n}\bigg)^{\frac{1+\alpha}{2\alpha}},\label{eq: a inequality}
\end{align}
where $c_2 = \edit{2}(1+\sqrt{2}) C_0 c_1^{1/2} \bigg(\frac{r^2}{1-r^{\frac{2+\alpha}{1+\alpha}}}\bigg) \vee 1$.
Plugging this back into \cref{eq: high prob Va} we get with probability at least $1-\exp(-t)$,
\begin{align}
V_a \le c_2 \sqrt{\frac{\bar V\log (n+1) }{n}} a^{\frac{\alpha}{1+ \alpha}-2} + \edit{\sqrt{\frac{8 \prns{c_1 a^{\frac{2\alpha}{1+ \alpha} -2} + 2c_2 \sqrt{\frac{\bar V\log (n+1)}{n}} a^{\frac{\alpha}{1+ \alpha}-2}} t}{a^2 n}} + \frac{4t}{3a^2n}}. \label{eq: va inequality}
\end{align}
Choose $\epsilon_n$ to be
\begin{align*}
\epsilon_n = \bigg(c_2 \sqrt{\frac{\bar V\log (n+1)}{n}}\bigg)^{\frac{1+\alpha}{2+\alpha}}.
\end{align*}
Note that the right hand side of \cref{eq: va inequality} is decresing in $a$ and $a\ge \epsilon_n$ by construction.
Thus, if $\epsilon_n$ satisfies
\begin{align*}
  \edit{\epsilon_n \ge \bigg(\frac{4 C_0^2}{c_1}\bigg)^{\frac{1+\alpha}{2\alpha}} \bigg(\frac{\bar V\log (n+1) }{n}\bigg)^{\frac{1+\alpha}{2\alpha}} \Longleftrightarrow n \ge  c_2^{-\alpha} \bigg(\frac{ 4 C_0^2}{c_1}\bigg)^{\frac{2+\alpha}{2}} \bar V\log (n+1),}
\end{align*}
we can substitute $\epsilon_n$ for $a$ to bound the right hand side of \cref{eq: va inequality}.
Note that
\begin{align*}
& c_2 \sqrt{\frac{\bar V\log (n+1)}{n}} a^{\frac{\alpha}{1+ \alpha}-2} \le \frac{\epsilon_n}{a} = \frac{1}{\sqrt{kt}} \le \frac{1}{\sqrt{k}}, \\
& a^{\frac{2\alpha}{1+ \alpha} -2} \le \epsilon_n^{\frac{2\alpha}{1+ \alpha} -2} = \prns{\epsilon_n^{\frac{\alpha}{1+ \alpha} -2}}^2 \epsilon_n^2 \le c_2^{-2} \bar V^{-1} n \epsilon_n^2,\\
& n \epsilon_n^2 =c_2^{\frac{2+2\alpha}{2+\alpha}} (\bar V\log (n+1))^{\frac{1+\alpha}{2+\alpha}}n^{\frac{1}{2+\alpha}} \ge 1.
\end{align*}
Therefore, with probability at least $1-\exp(-t)$ we have
\begin{align}
V_a \le &  \frac{1}{\sqrt{k}} + \sqrt{\frac{8(\edit{c_1} c_2^{-2} \bar V^{-1} n \epsilon_n^2 + \edit{2}) }{nk\epsilon_n^2}} + \frac{\edit{4}}{3nk\epsilon_n^2} \notag\\
\le &  \frac{1}{\sqrt{k}} + \sqrt{\frac{8(\edit{c_1} c_2^{-2}  + \edit{2}) }{k}} + \frac{\edit{4}}{3k}.  \label{eq: last va bound}
\end{align}
By choosing $k$ large enough we can make the right hand side of \cref{eq: last va bound} less than $1/2$, and we can conclude that
\begin{align*}
  \pr\mathbb(V_a < \frac{1}{2}) \ge 1-\exp(-t).
\end{align*}
Combining with \cref{eq: d and va} we get for all $t\ge 1$,
\begin{align*}
  \pr(d(\pi^*, \hat{\pi}_{\Pi}^{ERM})\ge kt\epsilon_n^2)\le \exp(-t).
\end{align*}
\edit{Thus, when $n \ge  c_2^{-\alpha} \bigg(\frac{ 4 C_0^2}{c_1}\bigg)^{\frac{2+\alpha}{2}} \bar V\log (n+1)$,}
\begin{align*}
\expect_{\mathcal D}[d(\pi^*, \hat{\pi}_{\Pi}^{ERM})] = & \int_0^{\infty} \pr(d(\pi^*, \hat{\pi}_{\Pi}^{ERM})\ge t') dt'\\
\le & k\epsilon_n^2 + \int_{k\epsilon_n^2}^{\infty} \pr(d(\pi^*, \hat{\pi}_{\Pi}^{ERM})\ge t') dt'\\
\le & (1+e^{-1})k\epsilon_n^2\\
\le & (1+e^{-1})k c_2^{\frac{2+2\alpha}{2+\alpha}} \bigg( \frac{\bar V\log (n+1)}{n}\bigg)^{\frac{1+\alpha}{2+\alpha}}.
\end{align*}
\edit{
On the other hand, when $n <  c_2^{-\alpha} \bigg(\frac{ 4 C_0^2}{c_1}\bigg)^{\frac{2+\alpha}{2}} \bar V\log (n+1)$, it is trivially true that
\begin{align*}
  \expect_{\mathcal D}[d(\pi^*, \hat{\pi}_{\Pi}^{ERM})] \le 2 \le 2c_2^{-\frac{\alpha(\alpha+1)}{2+\alpha}} \prns{\frac{4C_0^2}{c_1}}^{\frac{1+\alpha}{2}}\prns{\frac{\bar V\log (n+1)}{n}}^{\frac{1+\alpha}{2+\alpha}}.
\end{align*}
}

\edit{Finally, by noting that $ \op{Regret}(\hat \pi^\text{ERM}_{\Pi}) = B\expect_{\mathcal D}[d(\pi^*, \hat{\pi}_{\Pi}^{ERM})] $ we complete the proof.}
\myendproof

\proof{Proof of \cref{thm:fastrate IERM}}
\cref{thm:fastrate IERM} follows directly from \cref{asm:vc,prop: n-dim,thm:fastrateerm}.
\myendproof

{\blockedit
\subsubsection{Proof of \cref{thm:lowerboundfast}}
\proof{Proof of \cref{thm:lowerboundfast}}
Similarly to the proof of \cref{thm:lowerboundslow}, we will construct a collection of distributions and lower bound the average regret among them. This time, our distributions will additionally satisfy \cref{asm:margin}.

Again, we make some preliminary constructions.
For any $z\in \Z^\angle$, let again $\bar z$ denote the projection of $z$ onto $\op{conv}(\Z^\angle\backslash\{z\})$, and define
\begin{align*}
  w(z) = \frac{\rho(\Z)}{\norm{\bar z-z}^2}\prns{\bar z-z}.
\end{align*}
By definition of $\rho(\Z)$, we have $\norm{w(z)}\le 1$.
Moreover, like before, we have
\begin{align}\label{eq: gap}
 \min_{z'\in \Z^\angle\backslash\{z\}} w(z)\tr  \prns{z' - z}
  =  \min_{z'\in \Z^\angle \backslash\{z\}} w(z)\tr  \prns{\bar z - z} +  w(z)\tr  \prns{z' - \bar z}
  = \rho(\Z).
\end{align}

As before, since $\Pi$ has Natarajan dimension at least $\eta$, there exist $x_1,\dots,x_\eta\in\R p,\,z\s0_1\neq z\s1_1,\dots,z\s0_\eta\neq z\s1_\eta\in\Z^\angle$ such that, for every $\mathbf{b} = \prns{b_1, \dots, b_{\eta}} \in \braces{0,1}^{\eta}$, there is a $\pi_{\mathbf{b}}\in\Pi$ such that $\pi_{\mathbf{b}}(x_i)=z\s{b_i}_i$ for $i=1,\dots,\eta$.

We now construct a distribution $\mathbb P_{\mathbf b}$ for each $\mathbf{b} \in \braces{0,1}^{\eta-1}$ (notice there are half as many distributions as in the proof of \cref{thm:lowerboundslow}).
Set $\zeta = \prns{\frac{\eta-1}{n}}^{\frac{1}{2+\alpha}}$ and note that $\zeta\in[0,1/2]$ by assumption.
For the marginal distribution of $X$, we set
\begin{align*}
  & \pr_{\mathbf b}\prns{X = x_i} = \frac{\zeta^{\alpha}}{\eta-1} \text{ for } i\in[\eta-1], \text{ and }\\
  & \pr_{\mathbf b}\prns{X = x_{\eta}} = 1- \zeta^{\alpha}.
\end{align*}
We next construct the conditional distribution of $Y\mid X=x_i$.
For $i\in [\eta-1]$, 
let
\begin{align*}
  u'_{0i} = \frac{\zeta \fprns{w(z\s0_i) + w(z\s1_i)} + w(z\s0_i) - w(z\s1_i) }{2}, \\
  u'_{1i} = \frac{\zeta\fprns{w(z\s0_i) + w(z\s1_i)} + w(z\s1_i) - w(z\s0_i) }{2},
\end{align*}
Since $\norm{w(z\s0_i)}\le 1$ and $\norm{w(z\s1_i)}\le 1$, triangle inequality yields that $\norm{u'_{0i}}\le 1$ and $\norm{u'_{1i}} \le 1$.
We now construct the following conditional distribution of $Y\mid X= x_i$ for $i\in[\eta-1]$: if $b_i = 0$, we let
\begin{align*} %
  Y=
\begin{cases}
  u'_{0i},~~ \text{ with probability } (1+\zeta)/2,\\
  u'_{1i},~~ \text{ with probability } (1-\zeta)/2,
\end{cases}
\end{align*}
and if $b_i = 1$, we let
\begin{align*} %
  Y=
\begin{cases}
  u'_{1i},~~ \text{ with probability } (1+\zeta)/2,\\
  u'_{0i},~~ \text{ with probability } (1-\zeta)/2.
\end{cases}
\end{align*}
We then have that for $i\in[\eta-1]$,
$$
\E_{\mathbb P_{\mathbf b}}[Y\mid X=x_i]=f_{\mathbf b}(x_i)=\zeta w(z\s{b_i}_i).
$$
Finally, for $i=\eta$, we simply set $\pr_{\mathbf{b}}(Y= w(z\s1_\eta)) \mid X=x_{\eta}) = 1$. 
Then $\E_{\mathbb P_{\mathbf b}}[Y\mid X=x_\eta]=f_{\mathbf b}(x_\eta)=w(z\s1_\eta)$.

By \cref{eq: gap}, for $i\in[\eta-1]$, the optimal decision at $x_i$ is $z\s{b_i}_i$, and the optimal decision at $x_\eta$ is $z\s1_i$. In other words, the optimal policy is $\pi_{(b_1,\dots,b_{\eta-1},1)}$, which is in $\Pi$, as desired. For brevity, we will write $\pi_{\mathbf b}=\pi_{(b_1,\dots,b_{\eta-1},1)}$ when $\mathbf b\in\{0,1\}^{\eta-1}$.

Moreover, under every $\pr_{\mathbf b}$, by \cref{eq: gap},
we have that $\Delta(x_i)=\zeta\rho(\Z)$ for $i\in[\eta-1]$ and $\Delta(x_\eta)=\rho(\Z)$. Therefore, we have
\begin{align*}
  \pr_{\mathbf{b}}\prns{0<\Delta(X)\le \delta} = 
  \begin{cases}
    0, & \text{ for } \delta\in [0, \zeta\rho(\Z)),\\
    \zeta^{\alpha}, & \text{ for } \delta\in [\zeta\rho(\Z), \rho(\Z)),\\
    1, & \text{ for } \delta\in [\rho(\Z), \infty).
  \end{cases}
\end{align*}
Thus, \cref{asm:margin} is satisfied with $\alpha$ and $\gamma = B/\rho(\Z)$, as desired.

For $\hat{\pi} \in \Pi$, define $\hat{\mathbf{b}} \in \{0,1\}^{\eta-1}$ to be a binary vector whose $i$th element is $\hat{b}_i = \ind\braces{\hat{\pi}(x_i) = z\s1i}$.
Consider a prior on $\mathbf{b}$ such that $b_1, \dots, b_{\eta-1}$ are i.i.d. and $b_1\sim\op{Ber}(1/2)$, and as before let $\tilde \pr$ denote the joint distribution of $(\mathbf b,\mathcal D)$ under this prior. 
Letting $\op{Regret}_{\mathbf b}(\hat{\pi})$ denote the regret when the data is drawn from $\mathbb P_{\mathbf b}$,
the regret satisfies the following inequalities:
\begin{align*}
  \sup_{\mathbf b\in\{0,1\}^{\eta-1}} \op{Regret}_{\mathbf b}(\hat{\pi}) 
  &=  \sup_{\mathbf b\in\{0,1\}^{\eta-1}}\expect_{\mathbb{P}^n_{\mathbf{b}}} \expect_X \bracks{f_{\mathbf b}(X)\tr  \prns{\hat{\pi}(X) - \pi_{\mathbf b}(X)}}\\
  &\ge  \expect_{\mathbf{b}} \expect_{\mathbb{P}^n_{\mathbf{b}}} \expect_X \bracks{f_{\mathbf b}(X)\tr  \prns{\hat{\pi}(X) - \pi_{\mathbf b}(X)}} \\
  &\ge  \zeta \rho(\Z) \expect_{\mathbf{b}} \expect_{\mathbb{P}^n_{\mathbf{b}}} \pr_X \prns{\hat{\pi}(X) \ne \pi_{\mathbf b}(X)} \\
  & = \zeta \rho(\Z)\prns{\frac{\zeta^{\alpha}}{\eta-1} \sum_{i=1}^{\eta-1}\expect_{\mathbf{b}} \expect_{\mathbb{P}^n_{\mathbf{b}}} \bracks{\ind\braces{\hat{\pi}(x_i) \ne \pi_{\mathbf b}(x_i)}} + \prns{1-\zeta^{\alpha}} \expect_{\mathbf{b}} \expect_{\mathbb{P}^n_{\mathbf{b}}} \bracks{\ind\braces{\hat{\pi}(x_{\eta}) \ne \pi_{\mathbf b}(x_{\eta})}}}\\
  &\ge    \frac{\zeta^{\alpha+1} \rho(\Z)}{\eta-1} \sum_{i=1}^{\eta-1}\expect_{\mathbf{b}} \expect_{\mathbb{P}^n_{\mathbf{b}}} \bracks{\ind\braces{\hat{\pi}(x_i) \ne \pi_{\mathbf b}(x_i)}}   \\
  &\ge  \frac{\zeta^{\alpha+1} \rho(\Z)}{\eta-1} \sum_{i=1}^{\eta-1}\expect_{\mathbf{b}} \expect_{\mathbb{P}^n_{\mathbf{b}}}\bracks{\ind\braces{b_i \ne \hat{b}_i}} \\
  &\ge  \frac{\zeta^{\alpha+1} \rho(\Z)}{\eta-1} \sum_{i=1}^{\eta-1}\expect_{\tilde{\mathbb{P}}} \bracks{\min\braces{\tilde\pr\prns{b_i = 1\mid {\mathcal{D}}}, 1- \tilde\pr\prns{b_i = 1\mid {\mathcal{D}}}}}\\
  &\ge  \frac{\zeta^{\alpha+1} \rho(\Z)}{2} \exp\prns{-\frac{2\zeta}{1-\zeta}\sqrt{\frac{n\zeta^\alpha}{\eta-1}}},
\end{align*}
where the second inequality comes from the fact that $\min_{i\in[\eta]}\Delta(x_i) \ge \zeta \rho(\Z)$, the third inequality follows from $\prns{1-\zeta^{\alpha}}\zeta \rho(\Z) \expect_{\mathbf{b}} \expect_{\mathbb{P}^n_{\mathbf{b}}} \bracks{\ind\braces{\hat{\pi}(x_{\eta}) \ne \pi_{\mathbf b}(x_{\eta})}}$ being non-negative, the fourth inequality comes from the fact that $b_i \ne \hat{b}_i$ implies $\hat{\pi}(x_i) \ne \pi_{\mathbf b}(x_i)$, the fifth inequality follows from the same reasoning as in obtaining \cref{eq: min bayes risk}, and the sixth inequality follows from the same reasoning as in obtaining \cref{eq: lower bound with zeta}.

Finally, plugging in $\zeta = \prns{\frac{\eta-1}{n}}^{\frac{1}{2+\alpha}}$ and recalling $\zeta\leq1/2$ by assumption that $n\ge 2^{2+\alpha} (\eta-1)$, we have 
\begin{align*}
  \frac{\zeta^{\alpha+1} \rho(\Z)}{2} \exp\prns{-\frac{2\zeta}{1-\zeta}\sqrt{\frac{n\zeta^\alpha}{\eta-1}}} = \frac{\rho(\Z)}{2}\prns{\frac{\eta-1}{n}}^{\frac{1+\alpha}{2+\alpha}} \exp\prns{-\frac{2}{1-\zeta}} \ge \frac{\rho(\Z)}{2e^4}\prns{\frac{\eta-1}{n}}^{\frac{1+\alpha}{2+\alpha}},
\end{align*}
which concludes the proof.
\myendproof}

\subsection{Fast Rates for ETO (\cref{sec: fast eto})}
\proof{Proof of \cref{thm:fastrate}.}
By optimality of $\pi_{\hat f}$ with respect to $\hat f$, we have that
\begin{align*}
  f^*(X)\tr (\pi_{\hat{f}}(X) - \pi^*(X))
  \le &~ f^*(X)\tr \pi_{\hat{f}}(X) -\hat{f}(X)\tr\pi_{\hat{f}}(X) + \hat{f}(X)\tr\pi^*(X) - f^*(X)\tr\pi^*(X)\\
  \le &~ 2B\|f^*(X)- \hat{f}(X) \|.
\end{align*}
Thus, fixing $\delta>0$ and peeling on $\|f(X) -\hat{f}(X)\|$, we obtain
\begin{align*}
    \op{Regret}(\pi_{\hat{f}})
    = &~ \expect[f^*(X)\tr (\pi_{\hat{f}}(X) - \pi_{f^*}(X))\ind\{f^*(X)\tr (\pi_{\hat{f}}(X) - \pi_{f^*}(X))>0\}] \\
    \le &~ 2 B\expect[\|f^*(X) -\hat{f}(X)\|\ind\{f^*(X)\tr (\pi_{\hat{f}}(X) - \pi_{f^*}(X))>0\}] \\
    = &~ 2 B\expect[\|f^*(X) -\hat{f}(X)\|\ind\{f^*(X)\tr (\pi_{\hat{f}}(X) - \pi_{f^*}(X))>0,0 < \|f(X) -\hat{f}(X)\|\le  \delta\}] \\
    &~ + 2 B\sum_{r=1}^{\infty} \expect[\|f^*(X) -\hat{f}(X)\| \ind\{f^*(X)\tr (\pi_{\hat{f}}(X) - \pi_{f^*}(X))>0,2^{r-1} \delta < \|f^*(X) -\hat{f}(X)\|\le  2^r\delta\}] \\
    \le &~ 2 B \delta \pr(f^*(X)\tr (\pi_{\hat{f}}(X) - \pi_{f^*}(X))>0,0 < \|f^*(X) -\hat{f}(X)\|\le  \delta) \\
    &~ + B \delta\sum_{r=1}^{\infty} 2^{r+1}\pr(f^*(X)\tr (\pi_{\hat{f}}(X) - \pi_{f^*}(X))>0,2^{r-1} \delta < \|f^*(X) -\hat{f}(X)\|\le  2^r\delta)\\
    \le &~ 2 B \delta \pr(0<\Delta(X) \le 2B\delta)
 + B \delta\sum_{r=1}^{\infty} 2^{r+1}\pr(\|f^*(X) -\hat{f}(X)\|>2^{r-1} \delta , 0<\Delta(X) \le 2^{r+1}B\delta),
\end{align*}
where the very last inequality is due to the implication
\begin{align*}
f^*(X)\tr (\pi_{\hat{f}}(X) - \pi_{f^*}(X))>0,~\|f^*(X) -\hat{f}(X)\|\le  2^r\delta \implies &~ 0<f^*(X)\tr (\pi_{\hat{f}}(X) - \pi_{f^*}(X)) \le 2^{r+1}B\delta \\
\implies &~ 0<\Delta(X) \le 2^{r+1}B\delta,
\end{align*}
since $\pi_f(x)\in\Z^\angle$ is always an extreme point, for any $f$ and $x$.

Therefore, iterating expectations with respect to $X$, we have
\begin{align*}
\op{Regret}(\pi_{\hat{f}}) \le &~ 2 B \delta \pr(0<\Delta(X) \le 2B\delta)
  + B \delta\sum_{r=1}^{\infty} 2^{r+1}\expect\bigg[\pr(\|f^*(X) -\hat{f}(X)\|>2^{r-1} \delta\mid X) \ind\{ 0<\Delta(X) \le 2^{r+1}B\delta\}\bigg]\\
\le &~ 2 B \delta \pr(0<\Delta(X) \le 2B\delta)
   + C_1 B \delta\sum_{r=1}^{\infty} 2^{r+1} \exp(- C_2 a_n (2^{r-1}\delta)^2) \pr( 0<\Delta(X) \le 2^{r+1}B\delta)\\
   \le &~ \edit{B\gamma^\alpha (2 \delta)^{\alpha+1}
   + B\gamma^\alpha C_1 ( \delta)^{\alpha+1}\sum_{r=1}^{\infty} 2^{(r+1)(\alpha+1)} \exp(- C_2 a_n (2^{r-1}\delta)^2).} 
\end{align*}
If we take $\delta = a_n^{-1/2} $, we get
\begin{align*}
  \op{Regret}(\pi_{\hat{f}}) \le \edit{2^{\alpha+1}\gamma^\alpha B} \bigg[1+ C_1\sum_{r=1}^{\infty} 2^{r(\alpha+1)} \exp(- C_2(2^{2(r-1)})\bigg] a_n^{-(\alpha+1)/2} .
\end{align*}
\myendproof

\proof{Proof of \cref{thm:fastrateeto}}
When \cref{asm:vc,asm:compatibility} hold, by \cref{lemma: 14.15,empirical radius,population radius}, there are universal constants $(c_0, c_1, c_2)$ such that for any $\delta\ge c_0\sqrt{\frac{\nu\log (nd+1)}{n}}$ and almost all $x$,
\begin{align*}
  \pr(||\hat{f}_{\mathcal F} (x) - f^*(x)|| \ge \edit{\kappa}\delta) \le c_1e^{-c_2 n \delta^2}.
\end{align*}
Equivalently, there are universal constants $(c_1, c_2)$ such that for any $\delta>0$ and almost all $x$,
\begin{align*}
  \pr(||\hat{f}_{\mathcal F} (x) - f^*(x)|| \ge \delta) \le c_1e^{-c_2 \frac{n}{\nu\edit{\kappa^2} \log(nd+1)} \delta^2}.
\end{align*}
By \cref{thm:fastrate},
$$
\op{Regret}(\hat \pi^\text{ETO}_{\F})\leq C(\alpha, \gamma, B)\edit{\kappa^{1+\alpha}} \prns{\frac{\nu \log(nd+1)}{n}}^{\frac{1+\alpha}{2}}.
$$
\myendproof

\section{Verifying \cref{asm:compatibility} (Recovery)}\label{sec: veryify compat}

\begin{proposition}\label{prop: linear compat}
Suppose $\F$ is as in \cref{ex: linear predictors}, $\phi(X)$ has nonsingular covariance, and $\magd{\phi(X)}\leq B'$. Then \cref{asm:compatibility} is satisfied.
\end{proposition}
\proof{Proof}
Let $\Sigma$ denote the covariance of $\phi(X)$, $\sigma_{\min}>0$ its smallest eigenvalue, and $f^*(x)=W^*\phi(x)$. Then, for any $f(x)=W\phi(x)$,
\begin{align*}
&\E_X{\| f(X) - f^*(X)\|^2} = \sum_{j = 1}^d \E_X{\prns{W_j^\top \phi(X)- W_j^* X}^2 } = \sum_{j = 1}^d \prns{ W_j -  W_j^*}^\top\Sigma\prns{ W_j -  W_j^*},
\end{align*}
while for almost all $x$, $\|\phi(x)\|\leq B'$, and so,
\begin{align*}
\| f(x) - f^*(x)\|^2 &= \sum_{j=1}^d  \prns{\prns{ W_j- W_j^*}^\top \phi(x)}^2 \le \|\phi(x)\|\sum_{j=1}^d \prns{ W_j -  W_j^*}^\top\prns{W_j -  W_j^*}  \\
&\leq \frac{B'}{\sigma_{\min}}\sum_{j = 1}^d \prns{ W_j -  W_j^*}^\top\Sigma\prns{ W_j -  W_j^*},
\end{align*}
showing \cref{asm:compatibility} holds with $\kappa=B'/\sigma_{\min}$.
\myendproof

\begin{proposition}\label{prop: tree compat}
Suppose $\F$ is as in \cref{ex: trees} where interior nodes queries ``$w\tr x\leq\theta$?'' are restricted to $w$ being a canonical basis vectors and $\theta\in\{1/\ell,\dots,1-1/\ell\}$, $X\in[0,1]^d$, and $X$ has a density bounded below by $\mu_{\min}$. Then \cref{asm:compatibility} is satisfied.
\end{proposition}
\proof{Proof}
Fix $f\in\F$ and $x$.
Consider the intersection $\Set$ of the regions defined by leaves $x$ falls into in $f$ and in $f^*$. Note $\Set$ has volume at least $v_{\min}=(1/\ell)^{2D}$. Then,
\begin{align*}
\|{f}(x) - f^*(x)\|^2 &= \Efb{\|{f}(X) - f^*(X)\|^2\mid X\in\Set}\\
& \leq \Efb{\|{f}(X) - f^*(X)\|^2}/\prns{v_{\min}\mu_{\min}},
\end{align*}
completing the proof.
\myendproof

\section{Finite-Sample Guarantees for Nonparametric Least Squares with Vector-Valued Response}\label{sec: MSE ERM proof}

In this section we prove \cref{thm: MSE ERM}. In particular, we prove a generic result for vector-valued nonparametric least squares, which may be of general interest, and then apply it to the VC-linear-subgraph case.

\subsection{Preliminaries and Definitions}

\edit{For any $\F\subseteq[\R p\to\Y]$, let $\mathcal F^* = \mathcal F - f^*$.}
\edit{When $f^*\in \F$, we have $f^*\in\argmin_{f\in\F}\expect[||Y-f(X)||^2]$, where every element of this argmin is in fact equal to $f^*$ at almost all $x$.}

A set $\Set$ is \emph{star shaped} if $\lambda s\in\Set$ for any $\lambda\in[0,1],\,s\in\Set$. Thus, that \edit{$\F$} is star shaped at $f^*$ is equivalent to $\F^*$ being star shaped.

Define
\begin{align*}
w_i = Y_i - f^*(X_i)  \in \mathbb R^d,
\end{align*}
and note we have $||w_i||\le 2$.
Since the samples $\{(X_i, Y_i)\}_{i=1}^n$ are i.i.d, $w_1,\dots,w_n$ are independent.

Given a function $h=(h_1, \dots, h_d): \mathcal X \rightarrow \mathbb R^d$ and a probability distribution $\mathbb P$ on $\mathcal X$, define the $L_2(\mathbb P)$-norm:
\begin{align*}
||h||_2 = \sqrt{\expect ||h(X)||^2} = \sqrt{\expect \sum_{j=1}^d h^2_j(X)} .
\end{align*}
Given samples $\{X_1, \dots, X_n\}$, define the empirical $L_2$ norm:
\begin{align*}
||h||_n = \sqrt{\frac{1}{n}\sum_{i=1}^n ||h(X_i)||^2} = \sqrt{\frac{1}{n}\sum_{i=1}^n \sum_{j=1}^d h_j^2(X_i)}
\end{align*}

Define the localized $w$-complexity
\begin{align*}
\mathcal G_n(\delta; \mathcal H) = \expect_w\bracks{\sup_{h\in\mathcal H, ||h||_n \le \delta} \abs{\frac{1}{n}\sum_{i=1}^n w_i\tr h(X_i)}},
\end{align*}
where the expectation $\expect_w$ is only over $w_1,\dots,w_n$, \ie, over $(Y_1,\dots,Y_n)\mid(X_1,\dots,X_n)$.
Define the localized empirical Rademacher complexity
\begin{align*}
\hat{\mathcal R}_n(\delta;\mathcal H) = \expect_{\sigma}\bracks{\sup_{h\in\mathcal H, ||h||_n \le \delta} \abs{\frac{1}{n}\sum_{i=1}^n\sum_{j=1}^d \sigma_{ij} h_j(X_i)}},
\end{align*}
and the localized population Rademacher complexity
\begin{align*}
\bar{\mathcal R}_n(\delta;\mathcal H) = \expect_{\sigma, X}\bracks{\sup_{\edit{h}\in\mathcal H, ||h||_2 \le \delta} \abs{\frac{1}{n}\sum_{i=1}^n\sum_{j=1}^d \sigma_{ij} h_j(X_i)}},
\end{align*}
where $\{\sigma_{ij}\}_{i\in[n], j\in[d]}$ are i.i.d Rademacher variables (equiprobably $\pm1$).

\subsection{Generic Convergence Result.}
We will next prove the following generic convergence result for nonparametric least-squares with vector-valued response for a general function class $\F\edit{\subseteq[\R p\to\Y]}$.
\begin{theorem}\label{lemma: 14.15}
Suppose $\mathcal F^*$ is star-shaped. Let $\delta_n$ be any positive solution to $\frac{\bar{\mathcal R}_n(\delta; \mathcal F^*)}{\delta} \le \frac{\delta}{32}$, and $\epsilon_n$ be any positive solution to $\frac{\mathcal G_n(\epsilon; \mathcal F^*)}{\epsilon} \le \epsilon$ (note here $\epsilon_n$ is a random variable that depends on $\{X_i\}_{i=1}^n$).
There are universal positive constants $(c_0, c_1, c_2)$ such that
\begin{align*}
  \pr(||\hat{f}_{\mathcal F} - f^*||^2_2 \ge c_0 (\epsilon_n^2 + \delta_n^2)) \le c_1e^{-c_2 n \delta_n^2}.
\end{align*}
\end{theorem}

\subsubsection{Supporting Lemmas.}
We first prove a lemma that shows the functions $\delta \mapsto \frac{\mathcal G_n(\delta; \mathcal H)}{\delta}$ and $\delta \mapsto \frac{\bar{\mathcal R}_n(\delta; \mathcal H)}{\delta}$ are non-increasing, which will be used repeatedly in the rest of the proof.
\begin{lemma} \label{lemma: 13.6}
For any star-shaped function class $\mathcal H\subseteq[\mathcal X\to\R d]$, the functions $\delta \mapsto \frac{\mathcal G_n(\delta; \mathcal H)}{\delta}$ and $\delta \mapsto \frac{\bar{\mathcal R}_n(\delta; \mathcal H)}{\delta}$ are non-increasing on the interval $(0, \infty)$.
Consequently, for any constant $c>0$, the inequalities $\frac{\mathcal G_n(\delta; \mathcal H)}{\delta} \le c\delta$ and $\frac{\bar{\mathcal R}_n(\delta; \mathcal H)}{\delta} \le c\delta$ have a smallest positive solution.
\end{lemma}

\proof{Proof of \cref{lemma: 13.6}}
Given $0<\delta<t$ and any function $h\in \mathcal H$ with $||h||_n\le t$, we can define the rescaled function $\Tilde{h} = \frac{\delta}{t}h$ such that $||\Tilde{h}||_n \le \delta$. Moreover, since $\delta \le t$, the star-shaped assumption guarantees that $\Tilde{h}\in \mathcal H$. Therefore,
\begin{align*}
\frac{\delta}{t} \mathcal G_n(t; \mathcal H) = &\expect_w\big[\sup_{h\in\mathcal H, ||h||_n \le \delta}\abs{\frac{1}{n}\sum_{i=1}^n w_i \tr (\frac{\delta}{t}h(X_i))} \big] \\
= & \expect_w\big[\sup_{\Tilde{h} = \frac{\delta}{t}h: h\in\mathcal H, ||h||_n \le \delta} \abs{\frac{1}{n}\sum_{i=1}^n w_i \tr \Tilde{h}(X_i)}\big] \\
\le & \mathcal G_n(\delta; \mathcal H).
\end{align*}

The proof for $\frac{\bar{\mathcal R}_n(\delta; \mathcal H)}{\delta}$ is symmetric.
\myendproof

We now prove a technical lemma (\cref{lemma: 3.4}) that will be used to establish our main result.
\cref{lemma: 3.7,lemma: 3.8} below are, in turn, supporting lemmas used to prove \cref{lemma: 3.4}.

For any non-negative random variable $Z\ge 0$, define the entropy
\begin{align*}
  \mathbb H(Z) = \expect[Z\log Z] - \expect[Z]\log \expect[Z].
\end{align*}

\begin{lemma}\label{lemma: 3.7}
Let $X\in \mathbb R^d$ be a random variable such that $||X|| \le b$. Then for any convex and Lipschitz function $g: \mathbb R^d \rightarrow \mathbb R$, we have
\begin{align*}
  \mathbb H(e^{\lambda g(X)}) \le 4b^2\lambda^2 \expect[||\nabla g(X)||^2 e^{\lambda g(X)}] \quad \text{for all } \lambda >0,
\end{align*}
where $\nabla g(x)$ is the gradient (which is defined almost everywhere for convex Lipshitz functions).
\end{lemma}
\proof{Proof of \cref{lemma: 3.7}}
Let $Y$ be an independently copy of $X$.
By definition of entropy,
\begin{align*}
  \mathbb H(e^{\lambda g(X)}) = & \expect_X[\lambda g(X) e^{\lambda g(X)}] - \expect_X[e^{\lambda g(X)}] \log\prns{\expect_Y[e^{\lambda g(Y)}]} \\
  \le & \expect_X[\lambda g(X) e^{\lambda g(X)}] - \expect_{X,Y}[e^{\lambda g(X)}\lambda g(Y)] \\
  = &\frac{1}{2}\lambda \expect[(e^{\lambda g(X)}-e^{\lambda g(Y)})(g(X)-g(Y)) ] \\
  = & \lambda \expect[(e^{\lambda g(X)}-e^{\lambda g(Y)})(g(X)-g(Y)) \ind\{g(X) \ge g(Y)\}],
\end{align*}
where the inequality follows from Jensen's, and the last step follows from symmetry of $X$ and $Y$.
By convexity of the exponential, $e^s - e^t \le e^s (s-t)$ for all $s,t\in \mathbb R$, which implies $(e^s - e^t)(s-t) \ind\{s\ge t\} \le e^s (s-t)^2 \ind\{s\ge t\}$. Therefore,
\begin{align*}
  \mathbb H(e^{\lambda g(X)}) \le \lambda^2 \expect[e^{\lambda g(X)}(g(X)-g(Y))^2 \ind\{g(X) \ge g(Y)\}].
\end{align*}
Since $g$ is convex and Lipschitz, we have $g(x) - g(y) \le \ip{\nabla g(x)}{x-y}$, and hence, for $g(x)\ge g(y)$ and $||x||, ||y|| \le b$,
\begin{align*}
  (g(x) - g(y))^2 \le ||\nabla g(x)||^2||x-y||^2\le 4b^2||\nabla g(x)||^2.
\end{align*}
Combining the pieces yields the claim.
\myendproof

Given a function $f: \mathbb R^{nd} \rightarrow \mathbb R$, an index $k\in [n]$, and a vector $x_{-k} = (x_1,\dots,x_{k-1},x_{k+1},\dots,x_n)\in \mathbb R^{(n-1)d}$ where $x_i \in \mathbb R^d$, we define the conditional entropy in coordinate $k$ via
\begin{align*}
  \mathbb H(e^{\lambda f_k(X_k)}\mid x_{-k}) = \mathbb H(e^{\lambda f(x_1, \dots, x_{k-1}, X_k, x_{k+1}, \dots, x_n)}),
\end{align*}
where $f_k: \mathbb R^{d} \rightarrow \mathbb R$ is the function $x_k \mapsto f(x_1, \dots, x_k, \dots, x_n)$.

\begin{lemma}\label{lemma: 3.8}
Let $f: \mathbb R^{nd} \rightarrow \mathbb R$, and let $\{X_k\}_{k=1}^n$ be independent $d$-dimensional random variables. Then
\begin{align*}
  \mathbb H(e^{\lambda f(X_1, \dots, X_n)}) \le \expect\bigg[\sum_{k=1}^n \mathbb H(e^{\lambda f_k(X_k)}\mid X_{-k})\bigg] \quad \text{for all }\lambda >0.
\end{align*}
\end{lemma}
\proof{Proof of \cref{lemma: 3.8}}
By \citet[Eq. (3.24)]{wainwright2019high},
\begin{align}
    \mathbb H(e^{\lambda f(X)}) = \sup_g \{\expect[g(X)e^{\lambda f(X)}]\mid \expect[e^{g(X)}]\le 1\}.\label{eq: 3.24}
\end{align}

For each $j \in [n]$, define $X_j^n = (X_j, \dots, X_n)$. Let $g$ be any function that satisfies $\expect[e^{g(X)}]\le 1$. We can define a sequence of functions $\{g^1, \dots, g^n\}$ via
\begin{align*}
g^1(X_1, \dots, X_n) = g(X) - \log \expect[e^{g(X)}\mid X_2^n]
\end{align*}
and
\begin{align*}
  g^k(X_k, \dots, X_n) = \log \frac{\expect[e^{g(X)}\mid X_k^n]}{\expect[e^{g(X)}\mid X_{k+1}^n]} \quad \text{for } k = 2, \dots, n.
\end{align*}
By construction,
\begin{align*}
  \sum_{k=1}^n g^k(X_k, \dots, X_n) = g(X) - \log\expect[e^{g(X)}] \ge g(X)
\end{align*}
and $\expect[\exp(g^k(X_k, \dots, X_n))\mid X_{k+1}^n] = 1$. Therefore,
\begin{align*}
  \expect[g(X) e^{\lambda f(X)}] \le &\sum_{k=1}^n \expect[g^k(X_k, \dots, X_n)e^{\lambda f(X)}] \\
  = & \sum_{k=1}^n \expect_{X_{-k}}[\expect_{X_k}[g^k(X_k, \dots, X_n)e^{\lambda f(X)}\mid X_{-k}]] \\
  \le & \sum_{k=1}^n \expect_{X_{-k}}[\mathbb H(e^{\lambda f_k(X_k)}\mid X_{-k})],
\end{align*}
where the last inequality follows from \cref{eq: 3.24}.
Since $g$ is arbitrary, taking the supremum over the left-hand side and combining with \cref{eq: 3.24} yield the claim.
\myendproof

\begin{lemma}\label{lemma: 3.4}
Let $\{X_i\}_{i=1}^n$ be independent $d$-dimensional random vectors satisfying $||X_i|| \le b$ for all $i$, and let $f: \mathbb R^{nd}\rightarrow \mathbb R$ be convex and $L$-Lipshitz with respect to the Euclidean norm. Then, for all $\delta>0$,
\begin{align*}
  \pr \prns{f(X) \ge \expect[f(X)] + \delta} \le \exp\prns{-\frac{\delta^2}{16L^2 b^2}}.
\end{align*}
\end{lemma}
\proof{Proof of \cref{lemma: 3.4}}
For any $k\in [n]$ and fixed vector $x_{-k} \in \mathbb R^{n(d-1)}$, our assumption implies that $f_k$ is convex, and hence \cref{lemma: 3.7} implies that, for all $\lambda>0$,
\begin{align*}
  \mathbb H(e^{\lambda f_k(X_k)}\mid x_{-k}) \le 4b^2\lambda^2 \expect[||\nabla f_k(X_k)||^2 e^{\lambda f_k(X_k)}\mid x_{-k}].
\end{align*}
Combined with \cref{lemma: 3.8}, we find that
\begin{align*}
  \mathbb H(e^{\lambda f(X)}) \le 4b^2\lambda^2  \expect[\sum_{i=1}^n\sum_{j=1}^d\prns{\frac{\partial f(X)}{\partial x_{ij}}}^2 e^{\lambda f(X)}].
\end{align*}
Since $f$ is Lipschitz, we know $\sum_{i=1}^n\sum_{j=1}^d\prns{\frac{\partial f(X)}{\partial x_{ij}}}^2 \le L^2$ almost surely. The conclusion then follows from \citet[Proposition 3.2]{wainwright2019high}.
\myendproof

\subsubsection{Controlling $||\hat{f}_{\mathcal F} - f^*||_n$.}

In this section, we show that for any given samples, $||\hat{f}_{\mathcal F} - f^*||_n$ can be well-bounded with high probability (\cref{lemma: 13.5}). \cref{lemma: 13.12} is a supporting lemma that is used to prove \cref{lemma: 13.5}.

\begin{lemma} \label{lemma: 13.12}
  Fix sample points $\{x_i\}_{i=1}^n$.
  Let $\mathcal H\subseteq[\mathcal X\to\R d]$ be a star-shaped function class, and let $\delta_n>0$ satisfy $\frac{\mathcal G_n(\delta;\mathcal H)}{\delta} \le \delta$. For any $u\ge \delta_n$, define
  \begin{align*}
  \mathcal A(u) = \{\exists h\in \mathcal H \cap \{||h||_n \ge u\} \mid \frac{ 1}{n}\sum_{i=1}^n w_i \tr h(x_i)\ge 2||h||_n u\}.
  \end{align*}
  We have
  \begin{align*}
    \mathbb P_w(\mathcal A(u)) \le e^{-\frac{nu^2}{2}}.
  \end{align*}
\end{lemma}

\proof{Proof of \cref{lemma: 13.12}}
Suppose there exists some $h\in \mathcal H$ with $||h||_n\ge u$ such that
\begin{align*}
\frac{1}{n}\sum_{i=1}^n w_i \tr h_j(x_i)\ge 2||h||_n u.
\end{align*}
Let $\Tilde{h} = \frac{u}{||h||_n}h$, and we have $||\Tilde{h}||_n = u$. Since $h\in \mathcal H$ and $\frac{u}{||h||_n}\in(0,1]$, the star-shaped assumption implies that $\Tilde{h} \in \mathcal H$.
Therefore, $\mathcal A(u)$ implies that there exists a function $\Tilde{h} \in \mathcal H$ with $||\Tilde{h}||_n = u$ such that
\begin{align*}
\frac{1}{n}\sum_{i=1}^n  w_i \tr \Tilde{h}(x_i) = \frac{u}{n||h||_n}\sum_{i=1}^n w_i\tr h(x_i)\ge 2u^2.
\end{align*}
Thus, define $Z_n(u) = \sup_{\Tilde{h} \in \mathcal H, ||\Tilde{h}||_n\le u} \frac{1}{n}\sum_{i=1}^n w_i \tr \Tilde{h}(x_i)$, and we get
\begin{align*}
\mathbb P_w(\mathcal A(u)) \le \mathbb P_w(Z_n(u) \ge 2u^2).
\end{align*}

Let us view $Z_n(u)$ as a function of $(w_1, \dots, w_n)$.
It is convex since it is the maximum of a collection of linear functions.
We now prove that it is Lipschitz.
For another vector $w'\in \mathbb R^{nd}$, define $Z'_n(u) = \sup_{\Tilde{h} \in \mathcal H, ||\Tilde{h}||_n\le u} \frac{1}{n}\sum_{i=1}^n (w'_{i})\tr \Tilde{h}(x_i)$.
For any $\Tilde h \in \mathcal H$ with $||\Tilde h||_n \le u$, we have
\begin{align*}
  \frac{1}{n}\sum_{i=1}^n w_{i}\tr \Tilde{h}(x_i) - Z'_n(u) = & \frac{1}{n}\sum_{i=1}^n w_{i}\tr \Tilde{h}(x_i) - \sup_{\Tilde{h}' \in \mathcal H, ||\Tilde{h}'||_n\le u} \frac{1}{n}\sum_{i=1}^n (w'_{i})\tr \Tilde{h}'(x_i) \\
  \le & \frac{1}{n}\sum_{i=1}^n (w_{i}- w'_{i})\tr \Tilde{h}(x_i)\\
  \le & \frac{1}{\sqrt{n}} ||w-w'|| ||\Tilde h||_n\\
  \le & \frac{u}{\sqrt{n}} ||w-w'||,
\end{align*}
and taking suprema yields that $Z_n(u) - Z'_n(u) \le \frac{u}{\sqrt{n}}||w-w'||$.
Similarly, we can show that $Z'_n(u) - Z_n(u) \le \frac{u}{\sqrt{n}}||w-w'||$,
so $Z_n(u)$ is Lipschitz with constant at most $\frac{u}{\sqrt{n}}$.
By \cref{lemma: 3.4},
\begin{align*}
\mathbb P_w(Z_n(u) \ge \expect_w[Z_n(u)] + u^2)  \le e^{-\frac{nu^2}{64}}.
\end{align*}
Finally,
\begin{align*}
\expect_w[Z_n(u)] \le \mathcal G_n(u; \mathcal H) \le  u \frac{\mathcal G_n(\delta_n; \mathcal H)}{\delta_n} \le u\delta_n \le u^2,
\end{align*}
where the first inequality follows from \cref{lemma: 13.6}, and the second follows from the definition of $\delta_n$.
Thus,
\begin{align*}
\mathbb P_w(Z_n(u) \ge 2u^2) \le e^{-\frac{nu^2}{64}}.
\end{align*}
\myendproof

\begin{lemma} \label{lemma: 13.5}
Fix sample points $\{x_i\}_{i=1}^n$.
Suppose $\mathcal F^*$ is star-shaped, and let $\delta_n$ be any positive solution to $\frac{\mathcal G_n(\delta;\mathcal F^*)}{\delta} \le \delta$. Then for any $t\ge \delta_n$,
\begin{align*}
\mathbb P_w\big[||\hat{f}_{\mathcal F} - f^*||_n^2 \ge 16 t\delta_n\big]\le e^{-\frac{nt\delta_n}{64}}.
\end{align*}
\end{lemma}

\proof{Proof of \cref{lemma: 13.5}}
By definition,
\begin{align*}
  \frac{1}{2n}\sum_{i=1}^n ||Y_i - \hat{f}_{\mathcal F}(x_i)||^2 \le \frac{1}{2n}\sum_{i=1}^n ||Y_i - f^*(x_i)||^2.
\end{align*}
Recall that $Y_i = f^*(x_i) +  w_i$, so we have
\begin{align}
\frac{1}{2}||\hat{f}_{\mathcal F} - f^*||_n^2 \le \frac{1}{n} \sum_{i=1}^n  w_{i} \tr(\hat{f}(x_i) - f^*(x_i)), \label{eq: basic inequality}
\end{align}

Apply \cref{lemma: 13.12} with $\mathcal H = \mathcal F^*$ and $u=\sqrt{t\delta_n}$ for some $t\ge \delta_n$, we get
\begin{align*}
  \mathbb P_w(\mathcal A^c(\sqrt{t\delta_n}))\ge 1-e^{-\frac{nt\delta_n}{64}}.
\end{align*}

Let us now condition on $A^c(\sqrt{t\delta_n})$.
If $||\hat{f}_{\mathcal F}- f^*||_n < \sqrt{t\delta_n}$, it is obvious that $||\hat{f}_{\mathcal F}- f^*||^2_n < 16t\delta_n$. Otherwise, if $||\hat{f}_{\mathcal F}- f^*||_n \ge \sqrt{t\delta_n}$, \cref{eq: basic inequality} implies that
\begin{align*}
||\hat{f}_{\mathcal F} - f^*||_n^2 \le \frac{2}{n} \sum_{i=1}^n  w_{i}\tr(\hat{f}(x_i) - f^*(x_i)) < 4||\hat{f}_{\mathcal F} - f^*||_n \sqrt{t\delta_n},
\end{align*}
or equivalently $||\hat{f}_{\mathcal F}- f^*||^2_n < 16t\delta_n$.
Therefore,
\begin{align*}
\mathbb P_w\big[||\hat{f}_{\mathcal F} - f^*||_n^2 \ge 16 t\delta_n\big]\le \mathbb P_w(\mathcal A(\sqrt{t\delta_n}))\le e^{-\frac{nt\delta_n}{64}}.
\end{align*}
\myendproof

We next state a lemma that controls the deviations in the random variable $\abs{||h||_2^2 - ||h||_n^2}$, when measured in a uniform sense over a function class $\mathcal H$.
\begin{lemma}\label{lemma: 14.1}
Given a star-shaped function class $\mathcal H$ with $\sup_{h\in \mathcal H}\sup_x ||h(x)|| \le b$.
Let $\delta_n$ be any positive solution of the inequality
\begin{align*}
\bar{\mathcal R}_n(\delta; \mathcal H) \le \frac{\delta^2}{16b}.
\end{align*}
Then for any $t\ge \delta_n$, we have
\begin{align}
\abs{||h||_2^2 - ||h||_n^2} \le \frac{1}{2}||h||_2^2 + \frac{1}{2}t^2 \quad \text{for all } h \in \mathcal H \label{eq: event Ec}
\end{align}
with probability at least $1-2e^{-C\frac{nt^2}{b^2}}$, where $C$ is a universal constant.
\end{lemma}

\proof{Proof of \cref{lemma: 14.1}}
Define
\begin{align*}
Z'_n = \sup_{h\in \mathbb B_2(t;\mathcal H)} \abs{||h||_2^2 - ||h||_n^2}, \quad \text{where } B_2(t;\mathcal H) = \{h\in \mathcal H \mid ||h||_2\le t\}.
\end{align*}
Let $\mathcal E$ denote the event that \cref{eq: event Ec} is violated, and $\mathcal A_0 = \{Z'_n \ge t^2/2\}$.

We first prove that $\mathcal E \subseteq \mathcal A_0$.
We divide the analysis into two cases.
First, if there exists some function with $||h||_2 \le t$ that violates \cref{eq: event Ec}, then we must have $Z'_n \ge \abs{||h||_n^2 - ||h||_2^2} > \frac{1}{2}t^2$.
Otherwise, if \cref{eq: event Ec} is violated by some function with $||h||_2 > t$, we can define the rescaled function $\Tilde h = \frac{t}{||h||_2}h$ so that $||\Tilde h||_2 = t$.
By the star-shaped assumption, $\Tilde h\in \mathcal H$, so $Z'_n \ge \abs{||\Tilde h||_n^2 - ||\Tilde h||_2^2} \ge \frac{t^2}{||h||_2^2}\abs{||h||_n^2 - ||h||_2^2} >\frac{1}{2}t^2$.

We now control event $\mathcal A_0$, where we need to control the tail behavior of $Z'_n$.

We first control $\expect[Z'_n]$. Note that
\begin{align*}
\abs{||h(x)||^2 - ||h'(x)||^2} \le ||h(x)-h'(x)||(||h(x)|| + ||h'(x)||) \le 2b ||h(x)-h'(x)||.
\end{align*}
Therefore,
\begin{align*}
\expect[Z'_n] \le & 2\expect [\sup_{h\in \mathbb B_2(t;\mathcal H)}\abs{\frac{1}{n}\sum_{i=1}^n\sigma_i ||h(X_i)||^2}]\\
\le & 4\sqrt{2}b\expect [\sup_{h\in \mathbb B_2(t;\mathcal H)}\abs{\frac{1}{n}\sum_{i=1}^n\sum_{j=1}^d\sigma_{ij} h_j(X_i)}]\\
= & 4\sqrt{2}b \bar{\mathcal R}_n(t; \mathcal H),
\end{align*}
where the first inequality follows from a standard symmetrization argument (cf. Theorem 2.2 of \cite{pollard1990empirical}), and the second inequality follows from Corollary 1 of \cite{maurer2016vector}.
Since $\mathcal H$ is star-shaped and $t\ge \delta_n$, by \cref{lemma: 13.6},
\begin{align*}
  \frac{\bar{\mathcal R}_n(t; \mathcal H)}{t} \le \frac{\bar{\mathcal R}_n(\delta_n; \mathcal H)}{\delta_n} \le \frac{\delta_n}{16b},
\end{align*}
where the last inequality follows from our definition of $\delta_n$.
Thus, we conclude that $\expect[Z'_n] \le \frac{\sqrt{2}}{4}t^2$.

Next, we establish a tail bound of $Z'_n$ above $\expect[Z'_n]$. Let $g(x) = ||h(x)||^2 - \expect||h(X)||^2$. Since $\sup_x||h(x)||\le b$ for any $h\in \mathcal H$, we have $||g||_{\infty}\le b^2$, and moreover
\begin{align*}
  \expect g^2(X) \le \expect ||h(X)||^4 \le b^2\expect||h(X)||^2 \le b^2t^2,
\end{align*}
using the fact that $h\in B_2(t;\mathcal H)$.
By Talagrand's inequality \citep[Theorem 3.27]{wainwright2019high}, there exists a universal constant $C$ such that
\begin{align*}
\pr(Z'_n \ge \expect[Z'_n] + \frac{1}{7}t^2) \le 2\exp(-C\frac{nt^2}{b^2}).
\end{align*}
We thus conclude the proof by observing that $\expect[Z'_n] + \frac{1}{7}t^2 \le \frac{1}{2}t^2$.
\myendproof

\subsubsection{Proof of the generic convergence result.}

Equipped with \cref{lemma: 13.5,lemma: 14.1}, we are now prepared to prove \cref{lemma: 14.15}.

\proof{Proof of \cref{lemma: 14.15}}
First of all, note that $\sup_{f-f^*\in \mathcal F^*}\sup_x||f(x)-f^*(x)|| \le 2$.

When $\delta_n \ge \epsilon_n$, we have $\frac{\mathcal G_n(\delta_n; \mathcal F^*)}{\delta_n} \le \delta_n$, and by \cref{lemma: 13.5},
\begin{align*}
\mathbb P_w\big[||\hat{f}_{\mathcal F} - f^*||_n^2 \ge 16 \delta_n^2\big]\le e^{-\frac{n\delta_n^2}{64}}.
\end{align*}
On the other hand, \cref{lemma: 14.1} implies that
\begin{align*}
\pr(||\hat{f}_{\mathcal F} - f^*||_2^2 \ge 2||\hat{f}_{\mathcal F} - f^*||_n^2 + \delta_n^2) \le 2e^{-Cn\delta_n^2/4}.
\end{align*}
Therefore,
\begin{align*}
\pr(||\hat{f}_{\mathcal F} - f^*||_2^2 \ge 31\delta_n^2, \delta_n \ge \epsilon_n) \le 3 e^{-n\delta_n^2/(64+4/C)}.
\end{align*}

We now assume that $\mathcal A = \{\delta_n < \epsilon_n\}$ holds.
Define $\mathcal E = \{||\hat{f}_{\mathcal F} - f^*||_2^2 \ge 32 \epsilon_n^2 + \delta_n^2\}$, and $\mathcal B = \{||\hat{f}_{\mathcal F} - f^*||_n^2 \le 16 \epsilon_n^2\}$. It suffices to bound
\begin{align*}
  \mathbb P (\mathcal E \cap \mathcal A) \le\mathbb P (\mathcal E \cap \mathcal B) + \mathbb P (\mathcal A \cap \mathcal B^c).
\end{align*}
By \cref{lemma: 14.1},
\begin{align*}
  \mathbb P (\mathcal E \cap \mathcal B) \le \mathbb P (||\hat{f}_{\mathcal F} - f^*||_2^2 \ge 2 ||\hat{f}_{\mathcal F} - f^*||_n^2 + \delta_n^2)  \le 2e^{-Cn\delta_n^2/4}.
\end{align*}
By \cref{lemma: 13.5},
\begin{align*}
\mathbb P (\mathcal A \cap \mathcal B^c) \le \expect[e^{-\frac{n\epsilon_n^2}{64}}\ind\{\mathcal A\}] \le e^{-\frac{n\delta_n^2}{64}}.
\end{align*}
Putting together the pieces yields the claim.
\myendproof

\subsection{Application to VC-Linear-Subgraph Case}

To prove \cref{thm: MSE ERM},
we next apply \cref{lemma: 14.15} to the case of a VC-linear-subgraph class of functions. To do this, the key step is to compute the critical radii, $\epsilon_n,\delta_n$.

\subsubsection{Computing the Critical Radii.}

\begin{lemma} \label{empirical radius}
  Suppose \cref{asm:vc} holds and $\mathcal F^*$ is star-shaped.
  Let $\hat{\delta}_n^*$ and $\hat{\epsilon}_n^*$ be the smallest positive solutions to the inequalities $\hat{\mathcal R}_n(\delta;\mathcal F^*) \le \frac{\delta^2}{32}$ and $\mathcal G_n(\epsilon;\mathcal F^*) \le \epsilon^2$, respectively. Then there is a universal constant $C$ such that
  \begin{align*}
    \pr(\hat{\delta}_n^* \le C\sqrt{\frac{\nu\log (nd+1)}{n}})= 1, \quad
    \pr(\hat{\epsilon}_n^* \le C\sqrt{\frac{\nu\log (nd+1)}{n}})= 1.
  \end{align*}
\end{lemma}

\proof{Proof of \cref{empirical radius}}
Let $\bm g(f)= (f_1(X_1), f_2(X_1), \dots, f_d(X_n))=(e_1\tr f(X_1),e_2\tr f(X_1),\dots,e_d\tr f(X_n)) \in \mathbb R^{nd}$, where $e_j$ is the $j\thh$ canonical basis vector, and $\mathcal S = \{\bm g(f): f\in \mathcal F^*, ||f||_n \le \delta\}$. Note that $||\bm s|| \le \sqrt{n}\delta$ for all $\bm s\in \mathcal S$.
By \citet[Theorem 3.5]{pollard1990empirical},
\begin{align*}
   \expect_{\sigma}\Psi\prns{\frac{1}{J}\sup_{f\in\mathcal F^*, ||f||_n \le \delta} \abs{\sum_{i=1}^n\sum_{j=1}^d \sigma_{ij} f_j(X_i)}}\le 1, \quad \text{where } J = 9 \int_0^{\sqrt{n}\delta} \sqrt{\log D(t, \mathcal S)} d t,
\end{align*}
so by \cref{eq: l1 orlicz},
\begin{align*}
  \hat{\mathcal R}_n(\delta;\mathcal F^*) \le \frac{5}{n} J.
\end{align*}

Treat $(e_1,X_1),(e_2,X_1),\dots,(e_d,X_n)$ as $nd$ data points.
By (a vector version of) \citet[Lemma 2.6.18 (v)]{van1996weak}, $\F''=\{(\beta,x)\mapsto \beta\tr f(x):f\in\F^*, ||f||_n \le \delta\}$ has VC-subgraph dimension at most $\nu$ per \cref{asm:vc}.
Note that $\sqrt{n}\delta$ is the envelope of $\mathcal F''$ on $(e_1,X_1),(e_2,X_1),\dots,(e_d,X_n)$.
Applying Theorem 2.6.7 of \cite{van1996weak} gives
\begin{align*}
  D(\sqrt{n}\delta t,\mathcal S)
  \le & C(\nu+1)(16e)^{\nu+1}\prns{\frac{4nd}{t^2}}^{\nu}
\end{align*}
for a universal constant $C$.
We therefore obtain that for a (different) universal constant $C$
\begin{align*}
  \hat{\mathcal R}_n(\delta;\mathcal F^*)\leq \frac{C}{32}\sqrt{\frac{\nu\log (nd+1)} {n}}\delta.
\end{align*}
Thus, for any samples $\{X_i\}_{i=1}^n$, any $\delta_n \ge C\sqrt{\frac{\nu\log (nd+1)} {n}}$ is a valid solution to $\hat{\mathcal R}_n(\delta;\mathcal F^*) \le \frac{\delta^2}{32}$, which implies the first conclusion.

Now let us focus on $\hat{\epsilon}_n^*$.
Define $G_f = \sum_{i=1}^n w_i\tr f(X_i)$. Since $w_i\tr (f(X_i)-f'(X_i)) \le 2||f(X_i)-f'(X_i)||$, it is $2||f(X_i)-f'(X_i)||$-sub-Gaussian. Moreover, $w_i$ are independet, so we know that $G_f - G_{f'}$ is $2\sqrt{n}||f-f'||_n$-sub-Gaussian.
By Theorem 5.22 of \cite{wainwright2019high},
\begin{align*}
  \mathcal G_n(\epsilon; \mathcal F^*) \le \frac{64}{n}\int_0^{2\sqrt{n}\epsilon} \sqrt{\log N(t, \mathcal S)} dt.
\end{align*}
The rest of the proof is similar as before, and we omit the details here.
\myendproof

\begin{lemma} \label{population radius}
  Suppose \cref{asm:vc} holds and $\mathcal F^*$ is star-shaped.
  Let $\delta_n^*$ be the smallest positive solution to the inequality $\bar{\mathcal R}_n(\delta;\mathcal F^*) \le \frac{\delta^2}{32}$. For $nd\ge 2$, there is a universal constant $C$ such that
  \begin{align*}
    \delta_n^* \le C\sqrt{\frac{\nu\log (nd+1)}{n}}.
  \end{align*}
\end{lemma}

\proof{Proof of \cref{population radius}}
In what follows, we write $\bar{\mathcal R}_n(\delta;\mathcal F^*)$ as $\bar{\mathcal R}_n(\delta)$ and $\hat{\mathcal R}_n(\delta;\mathcal F^*)$ as $\hat{\mathcal R}_n(\delta)$.

Let $\hat{\delta}_n^*$ be the smallest positive solutions to the inequality $\hat{\mathcal R}_n(\delta;\mathcal F^*) \le \frac{\delta^2}{32}$.
We first show that there are universal constants $c_1, c_2$ such that
\begin{align} \label{eq: delta and hat delta}
\pr(\frac{\delta_n^*}{5} \le \hat{\delta}_n^* \le 3\delta_n^*) \ge 1-c_1e^{-\frac{c_2 n (\delta_n^*)^2}{\sqrt{\nu\log (d+1)}}}.
\end{align}

For each $t>0$, define the random variable
\begin{align*}
\bar{Z}_n(t) =  \expect_{\sigma}\big[\sup_{f\in\mathcal F^*, ||f||_2 \le t} \abs{\frac{1}{n}\sum_{i=1}^n\sum_{j=1}^d \sigma_{ij} f_j(X_i)}\big]
\end{align*}
so that $\bar{\mathcal R}_n(t) = \expect_X[\bar{Z}_n(t)]$ by construction.
Define the events
\begin{align*}
  \mathcal E_0(t) = \{\abs{\bar Z_n(t) - \bar R_n(t)} \le \frac{\delta^*_n t}{112}\} \quad \text{and}\quad \mathcal E_1 = \{\sup_{f\in \mathcal F^*} \frac{\abs{||f||^2_n - ||f||^2_2}}{||f||^2_2 + (\delta^*_n)^2} \le \frac{1}{2}\}.
\end{align*}
Conditioned on $\mathcal E_1$, we have for all $f\in \mathcal F^*$,
\begin{align*}
  ||f||_n \le \sqrt{\frac{3}{2}||f||^2_2 + \frac{1}{2}(\delta^*_n)^2} \le 2||f||_2 + \delta_n^* \quad \text{and} \quad ||f||_2 \le \sqrt{2||f||^2_n + (\delta^*_n)^2} \le 2||f||_n + \delta_n^*.
\end{align*}
As a result, conditioned on $\mathcal E_1$,
\begin{align} \label{eq: z le r}
  \bar Z_n(t) \le \expect_{\sigma}\big[\sup_{f\in\mathcal F^*, ||f||_n \le 2t + \delta_n^*} \abs{\frac{1}{n}\sum_{i=1}^n\sum_{j=1}^d \sigma_{ij} f_j(X_i)}\big] = \hat{\mathcal R}_n(2t+ \delta_n^*)
\end{align}
and
\begin{align} \label{eq: r le z}
  \hat{\mathcal R}_n(t) \le  \bar Z_n(2t+ \delta_n^*).
\end{align}

Let us consider the upper bound in \cref{eq: delta and hat delta} first.
Conditioned on $\mathcal E_0(7\delta_n^*)$ and $\mathcal E_1$, we have
\begin{align*}
  \hat{\mathcal R}(3\delta_n^*) \le \bar Z_n(7\delta_n^*) \le \bar{\mathcal R}_n(7\delta_n^*) + \frac{7}{112}(\delta^*_n)^2,
\end{align*}
where the first inequality follows from \cref{eq: r le z}, and the second follows from $\mathcal E_0(7\delta^*_n)$. By \cref{lemma: 13.6}, $\bar{\mathcal R}_n(7\delta_n^*) \le 7\bar{\mathcal R}_n(\delta_n^*) \le \frac{7}{32}(\delta_n^*)^2$. Thus, $\hat{\mathcal R}(3\delta_n^*)\le \frac{(3\delta_n^*)^2}{32}$, and we have $\hat \delta_n^* \le 3\delta_n^*$.

Now let us look at the lower bound in \cref{eq: delta and hat delta}.
Conditioned on $\mathcal E_0(\delta_n^*)$, $\mathcal E_0(7\delta_n^*)$ and $\mathcal E_1$, we have
\begin{align*}
  \frac{(\delta_n^*)^2}{32} = \bar{\mathcal R}_n(\delta_n^*)\le \bar Z_n(\delta_n^*) + \frac{1}{112}(\delta_n^*)^2 \le \hat{\mathcal R}_n(3\delta_n^*) + \frac{1}{112}(\delta_n^*)^2 \le \frac{3\delta_n^* \hat{\delta}_n^*}{32} +\frac{1}{112}(\delta_n^*)^2,
\end{align*}
where the first inequality follows from $\mathcal E_0(\delta_n^*)$, the second follows from \cref{eq: z le r}, and the third follows from the fact that $\hat \delta_n^* \le 3\delta_n^*$ and \cref{lemma: 13.6}.
Rearranging yields that $\frac{1}{5}\delta_n^*\le \hat \delta_n^*$.

Till now we have shown that
\begin{align*}
  \pr(\frac{\delta_n^*}{5} \le \hat{\delta}_n^* \le 3\delta_n^*) \ge \pr(\mathcal E_0(\delta_n^*)\cap \mathcal E_0(7\delta_n^*)\cap \mathcal E_1).
\end{align*}
\cref{lemma: 14.1} implies that $\pr(\mathcal E_1^c) \le c_1 e^{-c_2 n (\delta_n^*)^2}$.
Moreover, let
\begin{align*}
\bar Z^k_n(t) =  \expect_{\sigma}\big[\sup_{f\in\mathcal F^*, ||f||_2 \le t} \abs{\frac{1}{n}\sum_{i\in [n]-\{k\}}\sum_{j=1}^d \sigma_{ij} f_j(X_i)}\big],
\end{align*}
and we have
\begin{align*}
0 \le \bar Z_n(t) - \bar Z^k_n(t) \le \expect_{\sigma_k}\big[\sup_{f\in\mathcal F^*, ||f||_2 \le t} \abs{\frac{1}{n}\sum_{j=1}^d \sigma_{kj} f_j(X_k)}\big] \le \frac{\sqrt{\nu\log (d+1)}}{n},
\end{align*}
where the last inequality follows from a standard chaining argument.
Thus, by \citet[Theorem 15]{boucheron2003concentration} and noticing the fact that $\bar{\mathcal R}(\alpha \delta_n^*) \ge \bar{\mathcal R}( \delta_n^*) = \frac{(\delta_n^*)^2}{32}$ for any $\alpha\ge 1$, we have
\begin{align*}
  \pr(\mathcal E_0(7\delta_n^*)) \le c_1e^{-\frac{c_2 n (\delta_n^*)^2}{\sqrt{\nu \log(d+1)}}} \quad \text{and} \quad
\pr(\mathcal E_0(\delta_n^*)) \le c_1e^{-\frac{c_2 n (\delta_n^*)^2}{\sqrt{\nu\log (d+1)}}},
\end{align*}
so \cref{eq: delta and hat delta} follows.

By \cref{empirical radius}, $\pr(\hat{\delta}_n^* \le C_0\sqrt{\frac{\nu\log (nd+1)}{n}})= 1$ for some universal $C_0$. Let $C> 5C_0$ be a constant such that $c_1 \exp(-c_2  C^2 ) < 1$. If $\delta_n^* > C\sqrt{\frac{\nu\log (nd+1)}{n}}$, by \cref{eq: delta and hat delta} we have $\pr(\hat{\delta}_n^* > C_0\sqrt{\frac{\nu\log (nd+1)}{n}})>0$, which leads to contradiction. Thus, $\delta_n^* \le C\sqrt{\frac{\nu\log (nd+1)}{n}}$.

\subsubsection{Proof of \cref{thm: MSE ERM}.}

\proof{Proof of \cref{thm: MSE ERM}}
By \cref{lemma: 14.15,empirical radius,population radius}, there exists universal constant $(c_0, c_1, c_2)$ such that for any $\delta\ge c_0\sqrt{\frac{\nu\log (nd+1)}{n}}$,
\begin{align*}
  \pr(||\hat{f}_{\mathcal F} - f^*||_2 \ge \delta) \le c_1e^{-c_2 n \delta^2},
\end{align*}
and our conclusion follows.
\myendproof

\subsection{Vector-Valued Reproducing Kernel Hilbert Spaces}\label{appendix:rkhs}

In this section we develop an analogue of \cref{thm: MSE ERM} to the case of vector-valued RKHS instead of a VC-linear-subgraph class. We proceed by computing the critical radii and applying our \cref{lemma: 14.15}. Then, in \cref{sec:linearrkhs} we apply this to \cref{ex: linear predictors} in order to avoid the suboptimal logarithmic term one obtains by instead relying on its VC-linear-subgraph dimension and applying \cref{thm:slowrate,thm:fastrateeto}.

A (multivariate) positive semidefinite kernel is a function $\K:\R {p}\times \R {p}\to\R{d\times d}$ such that 
$\K(x,x')=\K(x',x)$ is symmetric and for all 
$m\in\mathbb N,x_1,\dots,x_m\in\R p,v_1,\dots,v_m\in\R d$ we have 
$\sum_{i=1}^m\sum_{j=1}^mv_i\tr\K(x_i,x_j)v_j\geq0$. We then consider the space 
$\op{span}(\{\K(x,\cdot)v:x\in\R p,v\in\R d\})\subseteq[\R p\to\R d]$ endowed with the inner product $\ip{\sum_{i=1}^m\K(x_i,\cdot)v_i}{\sum_{j=1}^{m'}\K(x'_{i},\cdot)v'_{i}}=\sum_{i=1}^m\sum_{j=1}^{m'}v_i\tr K(x_i,x'_j)v'_j$.
Using the norm $\|f\|_\K=\ip{f}{f}$,
this has a unique completion to a Hilbert space, which we call $\Hil_{\K}$. This Hilbert space has the property that for every 
$f\in\Hil_{\K},x\in\R p,v\in \R d$ we have $v\tr f(x)=\ip{\K(x,\cdot)v}{f}$, known as the representer property. That is, $(x'\mapsto\K(x,x')v)\in\Hil_\K$ is the Riesz representer of the linear operator $f\mapsto v\tr f(x)$, which is bounded.

Given these definitions, we now consider the hypothesis class given by an $R$-radius ball in $\Hil_\K$:
$$\F=\{f\in\Hil_\K:\magd f_\K\leq R\}.$$
We also allow $R=\infty$, in which case we set $\F=\Hil_\K$.
A prominent example of a multivariate kernel is a diagonal kernel: given a usual univariate positive semidefinite kernel, $\K':\R p\times\R p\to\Rl$, we let $\K(x,x')=\K'(x,x')I_{d\times d}$, \eg, $\K'(x,x')=\exp(-\magd{x-x'}^2/\sigma^2)$ or $\K'(x,x')=x\tr x'$. Then the above hypothesis class can simply be written as $\F=\{f\in\Hil_{\K'}^d:\sum_{i=1}^d\magd{f_i}_{\K'}^2\leq R^2\}$.
In particular, we can see that \cref{ex: linear predictors} is exactly given by $R=\infty$ and $\K'(x,x')=\phi(x)\tr\phi(x')$.

Let $\hat T_\K:\Hil\to\Hil$ be the operator $\hat T_\K f=\frac1n\sum_{i=1}^n\K(X_i,\cdot)f(X_i)$.
Let $L_2(\pr)$ denote the space of functions $\R p\to\R d$ with integrable square norm with respect to $X$.
Let $T_\K:L_2(\pr)\to L_2(\pr)$ be 
$T_\K f=\E[\K(X,\cdot)f(X)]$.
Assuming that $\E\|\K(X,X)\|^2<\infty$ using the operator norm, $T_\K$ is compact.
Let $\lambda_1\geq\lambda_2\geq\dots\geq0$ 
and $\hat\lambda_1\geq\hat\lambda_2\geq\dots\geq0$ 
denote the eigenvalues of $T_\K$ and $\hat T_\K$, respectively.

\begin{lemma}\label{lemma: rkhs rad}
$\G_n(\delta;\F^*)\leq \frac{4\sqrt{2}}{\sqrt{n}}
\prns{\sum_{j=1}^{nd}\min\{\delta^2,R^2\hat \lambda_j\}}^{1/2}$.
If moreover $\E\|\K(X,X)\|^2<\infty$, then
$\bar{\mathcal R}(\delta;\F^*)\leq \frac{2\sqrt{2}}{\sqrt{n}}
\prns{\sum_{j=1}^{\infty}\min\{\delta^2,R^2 \lambda_j\}}^{1/2}$.
Here, for $R=\infty$, we define $R^2z=\infty$ for $z\neq0$ and $R^2z=0$ for $z=0$.
\end{lemma}

\citet[Chapter 12--14]{wainwright2019high} and \citet[Chapter 4]{williams2006gaussian} compute the spectra of a variety of kernels.
Together with \cref{lemma: rkhs rad,lemma: 14.15}, we can then obtain convergence rates for \cref{eq:MSEERM} with RKHS $\F$.
We next consider the specific application to \cref{ex: linear predictors}.

\subsubsection{Application to \cref{ex: linear predictors}}\label{sec:linearrkhs}

Notice we can write $\F$ in \cref{ex: linear predictors} as $\F=\Hil_\K$ with $\K(x,x')=(\phi(x)\tr\phi(x'))I_{d\times d}$ (\ie, $R=\infty$). We now leverage \cref{lemma: rkhs rad,lemma: 14.15} to derive MSE convergence rates and corresponding slow and fast rates for ETO.

\begin{corollary} \label{lemma: linear class rates}
Assume $\F$ is the vector-valued linear function class defined in \cref{ex: linear predictors} and $f^* \in \F$. Then there exist universal constants $C_0, C_1, C_2>0$ such that, for any $\delta \le C_0$, with probability at least $1-C_1\delta^{dp'}$,
  $$\expect_X ||\hat f_\F (X) - f^*(X)|| \le C_2 \sqrt{\frac{dp'\log(1/\delta)}{n}}.$$
  Consequently, for a universal constant $C$,
  $$\op{Regret}(\hat \pi_\F^{ETO}) \le CB\sqrt{\frac{dp'}{n}}.$$
  And, if \cref{asm:margin,asm:compatibility} hold, then for a constant $C(\alpha, \gamma, B)$ depending only on $\alpha,\gamma,B$, $$\op{Regret}(\hat \pi_\F^{ETO}) \le C(\alpha, \gamma, B)\prns{\frac{dp'}{n}}^{\frac{1+\alpha}{2}}.$$
\end{corollary}
\proof{Proof}
The operators $\hat T_\K$ and $T_\K$ have each rank at most $dp'$ because their nonzero eigenvalues are respectively given by the duplicating $d$ times the nonzero eigenvalues of the matrices $\frac1n\sum_{i=1}^n\phi(X_i)\phi(X_i)\tr$ and $\E[\phi(X)\phi(X)\tr]$.
Therefore, \cref{lemma: rkhs rad} gives $\G_n(\delta;\F^*)\leq 4\sqrt{2}\delta \sqrt{\frac{dp'}n}$
and 
$\bar{\mathcal R}(\delta;\F^*)\leq {2\sqrt{2}}\delta \sqrt{\frac{dp'}n}$. Note that $\F$ is convex so it is also star-shaped. Then, applying \cref{lemma: 14.15} we obtain the first statement. The second statement is given by integrating the tail bound and applying \cref{thm: slowrate generic}. The third statement is given by applying \cref{thm:fastrate}.
\myendproof

\subsubsection{Proof of \cref{lemma: rkhs rad}}

The proof adapts arguments from \citet{mendelson2002geometric} to the vector-valued case.

\proof{Proof of \cref{lemma: rkhs rad}}
We first argue $\hat T_\K$ has at most $nd$ nonzero eigenvalues.
Define
the matrix $K\in\R{(n\times d)\times(n\times d)}$ given
by $K_{(i,k),(j,l)}=\frac1n(\K(X_i,X_j))_{kl}$.
If $f\in\Hil$ is an eigenfunction of $\hat T_K$ with eigenvalue $\lambda$, then for any $i$ we have $\lambda f(X_i)=(\hat T_Kf)(X_i)=\frac1n\sum_{j=1}^n\K(X_i,X_j)f(X_j)$. Letting $v\in\R{n\times d}$ be given by $v_{ik}=f_k(X_i)$, this means that $Kv=\lambda v$. So, either $v=0$, which means that $\hat T_Kf=0$ and hence $\lambda=0$, or $v$ is an eigenvector of $K$ with eigenvalue $\lambda$.
Of course, $K$ has at most $nd$ eigenvalues.

Let $\varphi_1,\varphi_2,\dots$ be an orthonormal basis of $\Hil$ such that $\hat T_\K\varphi_i=\hat\lambda_i\varphi_i$ for $i=1,\dots,nd$ and $\hat T_\K\varphi_i=0$ for $i>nd$. Fix $0\leq h\leq nd$. Consider $f\in\F$ (\ie, $\magd f_\K\leq R$) with $\magd f_n\leq\delta$.
Then we have, $\delta^2\geq\frac1n\sum_{i=1}^n\|f(X_i)\|^2=\ip{f}{\hat T_\K f}=\sum_{i=1}^{nd} \hat\lambda_i\ip{f}{\varphi_i}^2\geq \sum_{i=1}^{h} \hat\lambda_i\ip{f}{\varphi_i}^2$.
Therefore,
\begin{align}
\sum_{i=1}^nw_i\tr f(X_i)
&=\sum_{i=1}^n\ip{f}{\K(X_i,\cdot)w_i}\\
&=\ip{f}{\sum_{j=1}^\infty \ip{\sum_{i=1}^n\K(X_i,\cdot)w_i}{\varphi_j}\varphi_j}\\
&=\sum_{j=1}^h\ip{f}{\varphi_j}\ip{\sum_{i=1}^n\K(X_i,\cdot)w_i}{\varphi_j}\\
&\phantom{=}+\ip{f}{\sum_{j=h+1}^\infty \ip{\sum_{i=1}^n\K(X_i,\cdot)w_i}{\varphi_j}\varphi_j}\\
&\leq\delta\prns{\sum_{j=1}^h\frac1{\hat\lambda_j}\ip{\sum_{i=1}^n\K(X_i,\cdot)w_i}{\varphi_j}^2}^{1/2}\\
&\phantom{=}+R\prns{\sum_{j=h+1}^\infty \ip{\sum_{i=1}^n\K(X_i,\cdot)w_i}{\varphi_j}^2}^{1/2}.
\end{align}
Next note that, since $\E[w_iw_i\tr]\preceq 4 I_{d\times d}$,
\begin{align*}
\frac1n\E_w\ip{\sum_{i=1}^n\K(X_i,\cdot)w_i}{\varphi_k}^2
&=\frac1n\sum_{i=1}^n\sum_{j=1}^n\varphi_k(X_i)\tr\E_w\bracks{w_iw_j\tr}\varphi_k(X_j)\\
&\leq\frac4n\sum_{i=1}^n\|\varphi_k(X_i)\|^2=4\ip{\varphi_k}{\hat T_\K\varphi_k}=4\hat\lambda_k.
\end{align*}
We conclude from $\sqrt a+\sqrt b\leq \sqrt2\sqrt{a+b}$ that
$$\G_n(\delta;\F)\leq \frac{2\sqrt{2}}{\sqrt{n}}
\prns{\sum_{j=1}^{nd}\min\{\delta^2,R^2\hat\lambda_j\}}^{1/2}
.$$

Repeating the same argument using Mercer's theorem for $\T_\K$ in $L_2(\pr)$ (which may have more than $nd$ nonzero eigenvalues) and using Rademacher variables instead of $w_i$, noting that $\E[\sigma_i\sigma_j\tr]=I_{d\times d}$, we find that
$$\bar{\mathcal R}(\delta;\F)\leq \frac{\sqrt{2}}{\sqrt{n}}
\prns{\sum_{j=1}^{\infty}\min\{\delta^2,R^2\lambda_j\}}^{1/2}
.$$

Noting that $\G_n(\delta;\F^*)\leq 2\G_n(\delta;\F)$ and $\bar{\mathcal R}(\delta;\F^*)\leq 2\bar{\mathcal R}(\delta;\F)$ completes the proof.
\myendproof

\section{Convergence Rates for Vector-Valued Local Polynomial Regression}\label{appendix: local poly}

In this section we provide rates for vector-valued regression assuming H\"older smoothness and using local polynomial regression. Our arguments are largely based on those of \citet{stone1982optimal,audibert2007fast} but avoid the bad $d$-dependence one would get by na\"ively invoking their results for each response component. To do this we leverage a vector Bernstein concentration inequality \citep{minsker2017some}.

Fix $\beta>0$ and define $\lfloor \beta \rfloor=\sup\{j\in\mathbb Z:j<\beta\}$ as the largest integer \emph{strictly} smaller than $\beta$ (slightly differently than the usual floor function).
For any $x\in \mathbb{R}^p$ and any $\lfloor \beta \rfloor$-times continuously differentiable real-valued function function $g: \mathbb{R}^p \rightarrow \mathbb{R}$, define the Taylor expansion of $g$ at $x$ as
$$g_x(x') = \sum_{|s|\le \lfloor \beta \rfloor} \frac{(x'-x)^s}{s!}D^s g(x).$$
We say that $g: \mathbb{R}^p \rightarrow \mathbb{R}$ is $(\beta, L, \mathbb{R}^p)$-H\"{o}lder if it is $\lfloor \beta \rfloor$-times continuously differentiable and satisfies
$$\abs{g(x') - g_x(x')} \le L||x-x'||^{\beta}\qquad\forall x, x' \in \mathbb{R}^p.$$
We say that $g: \mathbb{R}^p \rightarrow \mathbb{R}^d$ is $(\beta, L, \mathbb{R}^p \rightarrow \mathbb{R}^d)$-H\"{o}lder smooth if each component, $g^{(i)}:x\mapsto (g(x))_i$, is $(\beta, L, \mathbb{R}^p)$-H\"{o}lder.
We also write $g_x = (g^{(1)}_x, \dots, g^{(d)}_x)$.

To estimate a vector-valued H\"{o}lder smooth function, we will use a local polynomial estimator.
Given a kernel $K(u)$ satisfying (examples include the uniform, Gaussian, and Epanechnikov kernels)
\begin{align}
  & \exists c>0: \quad K(x)\ge c\ind\{||x||\le c\} \quad \forall x \in \mathbb{R}^p, \label{eq: c constant}\\
  & \int_{\mathbb{R}^p} K(u)du = 1, \notag\\
  & \sup_{u\in \mathbb{R}^p}(1+||u||^{2\beta})K(u)<\infty, \notag\\
  & \int_{\mathbb{R}^p} (1+||u||^{2\beta})K(u)du <\infty, \notag\\
  & \int_{\mathbb{R}^p} (1+||u||^{4\beta})K^2(u)du <\infty, \notag
\end{align}
and a bandwidth $h>0$, the estimator at $x$ is defined as
$$
\hat{f}_n^{\text{LP}}(x)=\hat\vartheta_{(0,\dots,0)}\quad
\hat\vartheta\in\arg\min_{\vartheta\in\R{d\times M}}
      \sum_{i=1}^nK\prns{\frac{X_i-x}h} \left\|Y_i - \sum_{|s| \le \lfloor\beta\rfloor} {\vartheta}_s (X_i - x)^s\right\|^2,
$$
where $M$ is the cardinality of the set $\{s \in \mathbb{Z}^d_+: |s| \le  \lfloor\beta\rfloor\}$ and,
for each $|s| \le  \lfloor\beta\rfloor$, $\vartheta_s$ refers to the corresponding column of $\vartheta\in\R{d\times M}$.
In case of multiple minimizers $\hat\vartheta$ in the argmin, we just set $\hat{f}_n^{\text{LP}}(x)=0\in\R d$.
Finally, we define $\hat f^*_n(x)$ to be the projection of $\hat{f}_n^{\text{LP}}(x)$ onto the unit ball $\mathcal{B}_d(0,1)$.

We can then prove the following:
\begin{theorem} \label{thm: local poly}
Suppose $f^*$ is $(\beta, L, \mathbb{R}^p \rightarrow \mathbb{R}^d)$-H\"{o}lder.
Suppose the distribution of $X$ has a density with a compact support $\X$ on which it is bounded in $[\mu_{\min},\mu_{\max}]\subseteq(0,\infty)$.
Suppose moreover that for some $c_0,r_0>0$, $\text{Leb}[\mathcal{X} \cap \mathcal{B}(x,r)]\ge c_0 \text{Leb}[\mathcal{B}(x,r)]~\forall 0<r\le r_0,\, x\in \mathcal{X}$, where $\text{Leb}[\cdot]$ is the Lebesgue measure.
Set $h = n^{-1/(2\beta+p)}$.
Then, there exists $C_1,C_2>0$ depending only on $p,\beta,L,\mu_{\min},\mu_{\max},c_0,r_0$ 
such that for all $\delta>0$, $n\ge 1$, and almost all $x$,
  \begin{align}
    \pr\prns{\norm{\hat{f}^*_n(x) - f(x)}\ge \delta} \le   C_1 \exp\prns{-C_2 n^{2\beta/(2\beta+p)} \delta^2/d}.
  \end{align}
\end{theorem}

\subsection{Proof of \cref{thm: local poly}}

We first make some convenient definitions.
Define the vector $U(u) = (u^s)_{|s|\le \lfloor \beta \rfloor}\in \mathbb{R}^M$.
Let $h = n^{-1/(2\beta+p)}$ be a bandwidth, and define
the matrix $V \in \mathbb{R}^{d\times M}$, where for each $|s|\leq\lfloor\beta\rfloor$, the corresponding column of $V$ is
\begin{align}
  V_s = \sum_{i=1}^n (X_i-x)^s K \prns{\frac{X_i-x}{h}} Y_i \in \mathbb{R}^d,
\end{align}
and the matrix $Q = (Q_{s_1, s_2})_{|s_1|, |s_2|\le \lfloor \beta \rfloor}$, where
\begin{align}
  Q_{s_1, s_2}= \sum_{i=1}^n (X_i-x)^{s_1 + s_2} K \prns{\frac{X_i-x}{h}},
\end{align}
and the matrix $\bar{B} = (\bar{B}_{s_1, s_2})_{|s_1|, |s_2| \le \lfloor \beta \rfloor}$, where
\begin{align}
  \bar{B}_{s_1, s_2} = \frac{1}{nh^p}\sum_{i=1}^n \prns{\frac{X_i-x}{h}}^{s_1 + s_2} K \prns{\frac{X_i-x}{h}}.
\end{align}
It is easy to derive from \citet[Proposition 2.1]{audibert2007fast} that $\hat{f}_n^{\text{LP}}(x)$ can be expressed as
\begin{align*}
  \hat{f}_n^{\text{LP}}(x) = V Q^{-1} U(0)
\end{align*}
if $Q$ is positive definite, and $\hat{f}_n^{\text{LP}}(x) = 0$ otherwise.

\proof{Proof of \cref{thm: local poly}}
First of all, let $\mathcal{S}$ denote the class of all compact subsets of $\mathcal{B}(0,c)$ having Lebesgue measure $c_0 v_d c^p$, and we define
$$\mu_0 = \frac{1}{2}c \mu_{\min} \min_{ ||w|| = 1; S\in \mathcal{S}} \int_S \prns{\sum_{|s|\le \lfloor \beta \rfloor} w_s u^s}^2 du >0,$$
where $c$ is the constant in \cref{eq: c constant}.
By the same arguments as in \citet[Proof of Theorem 3.2]{audibert2007fast},
\begin{align} \label{eq: 6.3}
  \mathbb{P}(\lambda_{\min}(\bar{B})\le \mu_0) \le 2 M^2 \exp(-Cnh^p).
\end{align}
Since
\begin{align}\label{eq: 6.4}
  \mathbb{P}(||\hat{f}_n^{\text{LP}}(x)-f(x)||\ge \delta) \le \mathbb{P}(\lambda_{\min}(\bar{B})\le \mu_0)+ \mathbb{P}(||\hat{f}_n^{\text{LP}}(x)-f(x)||\ge \delta, \lambda_{\min}(\bar{B})> \mu_0),
\end{align}
we aim to control the second term in the rest of our proof.

Recall that we can write
\begin{align*}
  \hat{f}_n^{\text{LP}}(x) = V Q^{-1} U(0).
\end{align*}
Define the matrix $Z = (Z_{s,i})_{|s|\le \lfloor \beta \rfloor, 1\le i\le n}$ with elements
$$Z_{s,i} = (X_i-x)^s \sqrt{K \prns{\frac{X_i-x}{h}}}.$$
Denote the $s$th row of $Z$ as $Z_s$, and we introduce
$$Z^{(f)} = \sum_{|s|\le \lfloor \beta \rfloor} \frac{f^{(s)}(x)}{x!} Z_s \in \mathbb{R}^{d\times n}.$$
Since $Q = Z Z^T$, we get 
$$\forall |s|\le \lfloor \beta \rfloor, \quad Z_s Z^T Q^{-1} U(0) = \ind\{s=(0, \dots, 0)\},$$
hence $Z^{(f)} Z^T Q^{-1} U(0) = f(x)$. Thus, we can write
$$\hat{f}_n^{\text{LP}}(x) - f(x) = (V - Z^{(f)}Z^T)Q^{-1} U(0) = \textbf{a} \bar{B}^{-1} U(0),$$
where $\textbf{a} = \frac{1}{nh^p} (V - Z^{(f)}Z^T) H \in \mathbb{R}^{d\times M}$ and $H$ is the diagonal matrix $H = (H_{s_1, s_2})_{|s_1|, |s_2| \le \lfloor \beta \rfloor}$ with $H_{s_1, s_2} = h^{-s_1} \ind\{s_1 = s_2\}$. For $\lambda_{\min}(\bar{B})> \mu_0$,
\begin{align}\label{eq: 6.5}
  ||\hat{f}_n^{\text{LP}}(x) - f(x)|| \le ||\textbf{a}\bar{B}^{-1}||\le \mu_0^{-1}||\textbf{a}|| \le \mu_0^{-1} ||\textbf{a}||_{F} \le \mu_0^{-1} M \max_s||a_s||,
\end{align}
where $a_s$ is the $s$-th column of $\textbf{a}$ given by
$$a_s = \frac{1}{nh^p}\sum_{i=1}^n \prns{\frac{X_i -x}{h}}^s K\prns{\frac{X_i -x}{h}} \prns{Y_i - f_x(X_i)} .$$
Define
\begin{align*}
  & T_i^{(s,1)} = \frac{1}{h^p}\prns{\frac{X_i -x}{h}}^s K\prns{\frac{X_i -x}{h}} \prns{Y_i - f(X_i)}, \\
  & T_i^{(s,2)} = \frac{1}{h^p}\prns{\frac{X_i -x}{h}}^s K\prns{\frac{X_i -x}{h}} \prns{f(X_i) - f_x(X_i)}.
\end{align*}
We have
\begin{align*}
  \|a_s\| \le \norm{\frac{1}{n}\sum_{i=1}^n T_i^{(s,1)}} + \norm{\frac{1}{n}\sum_{i=1}^n (T_i^{(s,2)} - \expect T_i^{(s,2)})} + \norm{\expect T_i^{(s,2)}}.
\end{align*}
Define
\begin{align*}
  & \kappa_1 = \sup_{u\in \mathbb{R}^p} (1+||u||^{2\beta}) K(u),\\
  & \kappa_2= \mu_{\max} \int_{\mathbb{R}^p} (1+||u||^{4\beta}) K^2(u) du,\\
  & \kappa_3= \mu_{\max} \int_{\mathbb{R}^p} (1+||u||^{2\beta}) K(u) du.
\end{align*}
Note that $\expect T_i^{(s,1)} = 0$, $\norm{T_i^{(s,1)}}\le 2\kappa_1 h^{-p}$, and 
\begin{align*}
  & \expect \norm{T_i^{(s,1)}}^2 \le 4 h^{-p}   \mu_{\max}\int_{\mathbb{R}^p}  u^{2s} K^2\prns{u} du \le 4 \kappa_2 h^{-p},\\
  &  \norm{T_i^{(s,2)} - \expect T_i^{(s,2)}} \le \sqrt{d}L\kappa_1 h^{\beta-p} +\sqrt{d} L \kappa_3 h^{\beta} \le \sqrt{d}L(\kappa_1 + \kappa_3) h^{\beta-p} ,\\
  &  \expect \norm{T_i^{(s,2)} - \expect T_i^{(s,2)}}^2 \le \expect \norm{T_i^{(s,2)} }^2 \le dL^2 h^{2\beta-p} \mu_{\max}\int_{\mathbb{R}^p} \norm{u}^{2s+ 2\beta} K^2\prns{u}  du \le dL^2 \kappa_2 h^{2\beta-p} .
\end{align*}
Recall that $h = n^{-1/(2\beta + p)}$.
By \citet[Corollary 4.1]{minsker2017some}, for $\epsilon_1 \ge \frac{1}{3} (\kappa_1  + \sqrt{\kappa_1^2 + 36 \kappa_2 })h^{\beta} $,
\begin{align*}
  \pr\prns{\norm{\frac{1}{n}\sum_{i=1}^n T_i^{(s,1)}}\ge \epsilon_1} \le 28\exp\prns{-\frac{nh^p\epsilon_1^2}{8 \kappa_2  + 4  \kappa_1 \epsilon_1/3}},
\end{align*}
and for $\epsilon_2 \ge \frac{\sqrt{d}L}{6}(\kappa_1 + \kappa_3  + \sqrt{(\kappa_1 + \kappa_3)^2  + 36  \kappa_2 } ) h^{\beta}$,
\begin{align*}
  \pr\prns{ \norm{\frac{1}{n}\sum_{i=1}^n (T_i^{(s,2)} - \expect T_i^{(s,2)})} \ge \epsilon_2} \le  28\exp\prns{-\frac{nh^p\epsilon_2^2/2}{dL^2 \kappa_2  +  \sqrt{d}L(\kappa_1 + \kappa_3)\epsilon_2/3}}.
\end{align*}
Since also
$$\norm{\expect T_i^{(s,2)}} \le \sqrt{d}L h^{\beta} \mu_{\max}\int_{\mathbb{R}^p}   \norm{u}^{\beta + s} K\prns{u} du \le \sqrt{d}L \kappa_3 h^{\beta},$$
we get that when $2\ge \delta \ge M \mu_0^{-1} \prns{3\sqrt{d}L \kappa_3  \vee (\kappa_1  + \sqrt{\kappa_1^2 + 36 \kappa_2 }) \vee \frac{\sqrt{d}L}{2}(\kappa_1 + \kappa_3  + \sqrt{(\kappa_1 + \kappa_3)^2  + 36  \kappa_2 } ) } h^{\beta}$,
\begin{align*}
  \pr\prns{\norm{a_s}\ge \frac{\mu_0 \delta}{M}} \le & \pr\prns{\norm{\frac{1}{n}\sum_{i=1}^n T_i^{(s,1)}} \ge \frac{\mu_0 \delta}{3M}} + \pr\prns{\norm{\frac{1}{n}\sum_{i=1}^n (T_i^{(s,2)} - \expect T_i^{(s,2)})}\ge \frac{\mu_0 \delta}{3M}} \\
  \le & 56 \exp\prns{-Cnh^p \delta^2/d}.
\end{align*}
Recall that $\hat{f}^*_n(x)$ is the projection onto $\mathcal{B}_d(0,1)$.
Combined with \cref{eq: 6.3,eq: 6.4,eq: 6.5} we get when $\delta \ge M \mu_0^{-1} \prns{3\sqrt{d}L \kappa_3  \vee (\kappa_1  + \sqrt{\kappa_1^2 + 36 \kappa_2 }) \vee \frac{\sqrt{d}L}{2}(\kappa_1 + \kappa_3  + \sqrt{(\kappa_1 + \kappa_3)^2  + 36  \kappa_2 } ) } h^{\beta}$,
\begin{align*}
  \pr\prns{\norm{\hat{f}^*_n(x) - f(x)}\ge \delta} \le  C_1 \exp\prns{-C_2 nh^p \delta^2/d} = C_1 \exp\prns{-C_2 n^{2\beta/(2\beta+p)} \delta^2/d}.
\end{align*}
When $\delta < M \mu_0^{-1} \prns{3\sqrt{d}L \kappa_3  \vee (\kappa_1  + \sqrt{\kappa_1^2 + 36 \kappa_2 }) \vee \frac{\sqrt{d}L}{2}(\kappa_1 + \kappa_3  + \sqrt{(\kappa_1 + \kappa_3)^2  + 36  \kappa_2 } ) } h^{\beta}$, $\exp\prns{-C_2 n^{2\beta/(2\beta+p)} \delta^2/d}$ is lowered bounded by a constant independent of $n$ and $d$, so we know the inequality essentially holds for all $\delta>0$ (with possibly modified constants $C_1$).
\myendproof

\section{Omitted Details}

\subsection{Details for IERM under $\sigma^2=0$ in \cref{sec:simple}}\label{sec:simpleierm}

If $\sigma^2=0$ then $Y_i=X_i$ and therefore the set of IERM solutions is
$\argmin_{\theta\in[-1,1]}\frac1n\sum_{i=1}^nX_i(1-2\indic{X_i\leq\theta})=[\max(\{X_i:X_i<0\}),\,\min(\{X_i:X_i>0\})]$, where we define $\max(\varnothing)=-1$ and $\min(\varnothing)=1$.
Thus, we have that $\hat\pi_\F^\text{IERM}=\pi_{f_{\hat\theta_\text{IERM}}}$ where $\hat\theta_\text{IERM}=\max(\{X_i:X_i<0\})$.
Let $n_-=\abs{\{X_i:X_i<0\}}\sim\op{Bin}(n,\frac12)$.
Then, $\E[\hat\theta_\text{IERM}^2\mid n_-]=\int_0^1\fPrb{\hat\theta_\text{IERM}<-\sqrt{u}\mid n_-}du=\int_0^1(1-\sqrt u)^{n_-}du=2/(2+3n_-+n_-^2)$. Notice this works even when $n_-=0$.
We conclude that $\op{Regret}(\hat\pi_\F^\text{IERM})=\E[1/(2+3n_-+n_-^2)]=(4-(3+n)/2^n)/((n+1)(n+2))=\Theta(1/n^2)$.

\subsection{Details for the experiment in \cref{sec:exp}}\label{sec:expdetail}

\textbf{Data-Generating Process.}
Here we specify the distribution from which we draw $(X,Y)$.
Recall $Y$ has $d=40$ dimensions.
We consider covariates $X$ with $p=5$ dimensions, with the data generated as follows.
We let $X\sim\mathcal N(0,I_p)$ be drawn from the standard multivariate normal distribution. We then set $Y=\op{diag}(\epsilon)(W\phi(X)+3)$, where $\phi(x)\in\R{31}$ consists of all features and all products of any number of distinct features (\ie, $\phi(x)=(\prod_{j=1}^5x_j^{k_j}:k_j\in\{0,1\},1\leq\sum_{j=1}^5k_j\leq 5)=(x_1,\dots,x_5,x_1x_2,x_1x_3,\dots,x_1x_2x_3,\dots,x_1x_2x_3x_4x_5)$), $W\in\R{40\times 31}$ is a fixed coefficient matrix, and $\epsilon\sim\op{Unif}[3/4,5/4]^d$ is a multiplicative noise.
Note $f^*(x)=W\phi(x)+3$ is a degree-5 polynomial in $x$.
To fix some matrix $W$ we draw its entries independently at random from $\op{Unif}[0,1]$. We do this just once, with a fixed random seed of 10, so that that $W$ is a fixed matrix. For each replication of the experiment, we then draw a training dataset of size $n$ from this distribution of $(X,Y)$.

\textbf{Methods.}
As detailed in \cref{sec:exp}, we consider 6 methods: ETO using least-squares and SPO+, each using three different hypothesis classes $\F$.
We employ a ridge penalty with parameter $\lambda$ in each of these cases, \ie, $\lambda$ times the squared sum of linear coefficients for both linear settings and $\lambda$ times the RKHS norm of $f$ in the kernel setting.
In the kernel setting, there is an additional parameter, $\rho$, known as the length-scale of the Gaussian kernel.
We choose $\lambda$ (and also $\rho$ in the kernel setting) by validation. We use an independent validation dataset of size $n$. For ETO, we focus on least squares: we choose the parameters that result  in minimal squared error on the validation data. For SPO+, we focus on the decision-problem and we choose the parameters that result in minimal average decision costs on the validation data (\ie, the IERM cost function on the validation set). This validation scheme is in line with ETO doing the regression step as a completely separate procedure that disregards the optimization problem and IERM integrating the steps and directly targeting decision costs.
\edit{In the linear settings we search over $\lambda\in\{0, \frac1{10},\frac1{10^{2/3}},\frac1{10^{1/3}},1,\dots,100\}$.
In the RKHS settings, we search over $\lambda\in\{\frac1{10^3}, \frac1{10^2}, \frac1{10},\frac1{10^{2/3}},\frac1{10^{1/3}},1,\dots,100\}$ and $\rho\in\{0.01, 0.1,0.5,1,2\}$ (we drop $\lambda=0$ in the RKHS case as it leads to an ill-posed solution). 
We solve SPO+ using the formulation in Section 5.1 of \cite{elmachtoub2017smart} and Gurobi 9.1.1. 
Because this is extremely slow for the RKHS case, we can only do so up to $n = 500$, and for larger $n$, we use the stochastic gradient descent (SGD) approach in Appendix C of \cite{elmachtoub2017smart}. 
We follow the their accompanied implementation in \url{https://github.com/paulgrigas/SmartPredictThenOptimize}, setting the batch size to $10$, number of iterations to $1000$, and the step size to $\frac{1}{\sqrt{t+1}}$ for the $t^{\text{th}}$ SGD iteration.
In \cref{figa}, we show that the results of this SGD approach closely track the reformulation approach for $n \le 500$.}
We solve the ridge-penalized least squares using the python library scikit-learn.

\textbf{Results.}
For each of $n=50, 100, \dots, 1000$, we run $50$ replications of the experiment. Using a test data set of $10000$ draws of just $X$, we then compute the sample averages of $\E_X\bracks{f^*(X)\pi^*(X)}$ and of $\E_X\bracks{f^*(X)\tr\hat\pi(X)}$ for each policy $\hat \pi$ resulting from one of the 6 methods and for each replication. For any $x$ and $f$, we compute $\pi_f(x)$ using Gurobi 9.1.1. Recall $\pi^*(X)=\pi_{f^*}(X)$ so that this is also applied to computing $\pi^*(X)$. Finally, by computing averages over replications, we estimate the relative regret for $n$ and for each method $\hat\pi$, being $\E_\D\E_X\bracks{f^*(X)\tr(\hat\pi(X)-\pi^*(X))}/\E_\D\E_X\bracks{f^*(X)\pi^*(X)}$.

\end{APPENDICES}

\end{document}